\documentclass[10pt]{article} 
\usepackage[table]{xcolor}
\usepackage[utf8]{inputenc}
\usepackage[T1]{fontenc}
\usepackage{times}
\usepackage[round]{natbib}
\usepackage{latexsym}
\usepackage{amsmath,amssymb}
\usepackage{graphicx}
\usepackage{array}
\usepackage{placeins}
\usepackage{booktabs}
\usepackage{multirow}
\usepackage{longtable}
\usepackage{textcomp}
\usepackage{svg}
\usepackage{hyperref}
\usepackage[letterpaper,margin=1in]{geometry}
\newcommand{\iclrheader}{Under review as a conference paper at ICLR 2027}
\setcitestyle{round}
\makeatletter
\def\ps@plain{%
  \let\@mkboth\@gobbletwo
  \def\@oddhead{\footnotesize\hfil\iclrheader\hfil}%
  \def\@evenhead{\footnotesize\hfil\iclrheader\hfil}%
  \def\@oddfoot{\hfil\thepage\hfil}%
  \def\@evenfoot{\hfil\thepage\hfil}%
}
\makeatother

\definecolor{rowDeepSeek}{rgb}{0.91, 0.87, 0.98}
\definecolor{rowQwen}{rgb}{0.85, 0.96, 0.88}
\definecolor{rowBase}{rgb}{0.85, 0.92, 0.98}
\definecolor{rowSFT}{rgb}{0.99, 0.92, 0.80}
\definecolor{rowRL}{rgb}{0.99, 0.86, 0.86}

\title{Everything in Moderation: Per-Domain Coverage Optima and Alignment-Resistant Domain Gaps in Multi-Domain Mid-Training}

\author{\normalsize Yunpeng Xu \\
        \normalsize Kun Zheng}

\begin{document}
\maketitle

\begin{abstract}
Mid-training---the stage between pre-training and alignment---is where a model's per-domain data composition is usually decided by data availability rather than principled design. We ask what that decision buys and whether a later alignment pass can undo it. In a controlled logical-reasoning setting (Qwen3-8B-Base primary, with a 4B replication; five semantically rule-disjoint KOR-Bench domains) we train 30 allocations spanning the five-domain simplex---24 sweep configurations plus six withheld from the fit---at five seeds each. Three findings emerge. First, \textbf{every domain has an interior coverage optimum}: the moderate band ($10$--$40\%$) is best for all five domains (a descriptive band concordance; a calibrated permutation test for quadratic interiority gives $P\approx0.010$, Appendix~\ref{app:bands}), and the fitted mid-training-only curves, with 8B peaks between $9.9\%$ and $35.1\%$, reproduce out-of-sample for the curve shape (not the peak locations). Second, \textbf{the gaps survive a fixed-budget alignment pass}: compensatory SFT raises 116/120 cells (mean $+4.32$\,pp) yet bridges $0/240$ pairs at a $5$\,pp threshold and only $30/240$ at a $10\%$ ratio, an equal-budget uniform control behaves almost identically, and a permutation null reallocating the same gains at random would bridge $13.8\pm3.3$ and $77.9\pm8.5$ pairs ($P<0.001$). Third, zero coverage collapses mid-training-only accuracy, and the tested recipe reverses both signs---clearly for Operation, not resolvably for Counterfactual---though a FineWeb-Edu-only control shows the collapse is co-mingled with generic distributional drift. An exploratory $\theta^*$ allocation attains the largest full-pipeline gain of the three trained end-to-end ($+4.36$ vs.\ $+0.80$/$+0.64$\,pp) but is marginal under an uncorrected Welch test and is selected from the same sweep. Because coverage varies on a simplex, the five optima are mixture-level marginals, not the coordinates of one jointly optimal mixture.\end{abstract}

\section{Introduction}

The multi-stage training paradigm (pre-training $\rightarrow$ mid-training $\rightarrow$ Supervised Fine-Tuning (SFT) $\rightarrow$ Reinforcement Learning (RL)) assumes each stage provides a foundation the next can refine. But some early design choices may create constraints that later stages cannot undo. This paper tests that possibility for one specific decision: \textbf{per-domain data coverage at mid-training}. Mid-training---the stage between pre-training and alignment---is a general domain-adaptation technique, yet per-domain data composition is typically set by data availability rather than principled design. We study whether, in a fixed total-token mixture and a fixed downstream recipe, coverage-associated differences are subsequently reduced by SFT or RL; we do not treat this study as a test of whether alignment procedures in general can repair coverage gaps.

Prior work has partially explored this question but leaves the coverage-allocation dimension untested:~\citet{zhang2025interplay} showed pretraining exposure determines RL generalization in a small synthetic setting,~\citet{zhao2025echo} that RL amplifies pretrained behaviors, and the SFT/RL literature that SFT primarily teaches format while RL generalizes from existing capabilities~\citep{chu2025sft}; mid-training work treats continued pretraining as a bridge to downstream alignment~\citep{tu2025midtrain}, and data-mixture methods---DoReMi~\citep{xie2024doremi}, RegMix~\citep{liu2024regmix}---optimize domain proportions for aggregate perplexity. None of these varies \emph{per-domain mid-training coverage} while holding subsequent stages fixed (Appendix~\ref{app:related}).

\begin{figure}[!htbp]
\centering
\includegraphics[width=\textwidth]{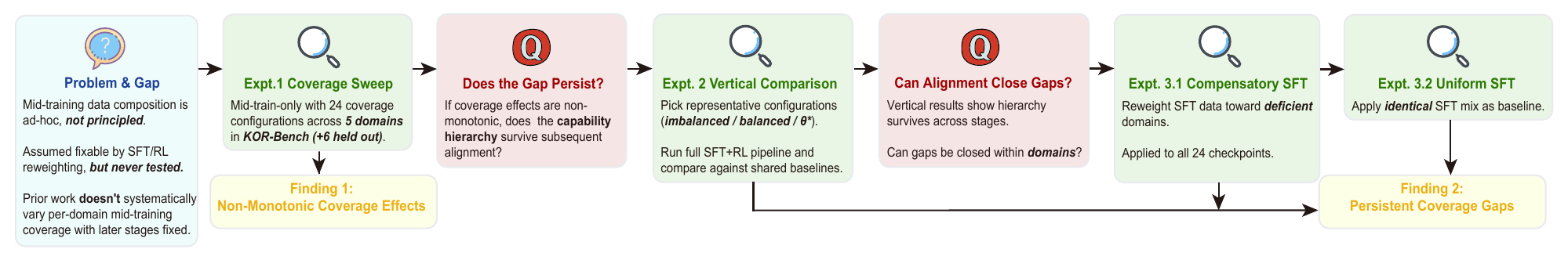}
\caption{Experimental flow. The downstream recipe (mid-training$\rightarrow$SFT$\rightarrow$RL) is held fixed while the five-domain coverage mixture varies: 24 sweep configurations (plus six held-out allocations) establish per-domain interior optima (Finding~1, \S\ref{sec:coverage_impact}); three allocations carry the vertical comparison through the full pipeline (\S\ref{sec:stages}); compensatory vs.\ uniform SFT on all 24 checkpoints tests gap closure (Finding~2, \S\ref{sec:gap_closure}).}
\label{fig:exp_flow}
\end{figure}

We address this gap through a controlled study on Qwen3-8B-Base using five logical-reasoning domains from KOR-Bench~\citep{ma2024korbench}; these domains are semantically rule-disjoint---each defines a self-contained symbolic system---which reduces semantic-transfer confounding, though they are not statistically or neurally independent (\S\ref{sec:bench}). We construct 24 coverage configurations spanning the five-domain simplex---from severely imbalanced to balanced to counter-skewed---and hold the planned downstream stages fixed. Coverage is the intended allocation variable, but changing one component necessarily changes the others and may also alter effective repetition and diversity. We evaluate on KOR-Bench and three external benchmarks (CounterBench~\citep{chen2026counterbench}, ProofWriter~\citep{tafjord2021proofwriter}, ZebraLogic~\citep{lin2025zebralogic}); three findings emerge (Figure~\ref{fig:exp_flow}; evidence in Sections~\ref{sec:coverage_impact}--\ref{sec:cross_domain}), each stated for the tested setting and marked by what the six withheld allocations support out-of-sample.

\textbf{Observation 1: The gaps survive the reweightings that keep the average gain; closing them forfeits that gain} (\S\ref{sec:stages}, \S\ref{sec:gap_closure}). Reweighting SFT data toward coverage-deficient domains raises absolute accuracy in 116 of 120 configuration--domain cells yet bridges $0/240$ pairwise gaps at the $5$\,pp threshold and only $30/240$ at the $10\%$ ratio, and an equal-budget uniform-SFT control behaves almost identically, so the non-closure is not an artifact of the compensation formula. A permutation null holding the observed gain magnitudes fixed would bridge $13.8\pm3.3$ pairs at $5$\,pp and $77.9\pm8.5$ at $10\%$ ($P<0.001$; Appendix~\ref{app:pairwise}): the gains are placed in a systematically gap-preserving way. Sharpening the policy within the same family closes gaps, but only by trading away the accuracy it was introduced to deliver: raising the concentration exponent closes $0/60\to12/60$ pairs at the $5$\,pp metric while the mean gain falls $+4.34\to+2.26$\,pp (Appendix~\ref{app:sharpen}), so closure and average accuracy are in tension under a fixed budget. The short GSPO stage is a secondary observation (\S\ref{sec:rl_limits}).

\textbf{Observation 2: Transient negative transfer under zero coverage} (\S\ref{sec:stages}). The strongest raw effect is negative transfer at the mid-training-only checkpoint: driving a domain to zero mid-training coverage collapses its accuracy even where the prior was high (Counterfactual 83.6\%$\rightarrow$45.6\%; Operation 60.4\%$\rightarrow$33.2\%). The tested recipe partly repairs this (Table~\ref{tab:survival}): Operation returns to $+5.2\pm0.3$\,pp by mid+SFT, clearly separated from zero, while Counterfactual only turns positive by the RL row ($+1.2\pm3.4$\,pp, not statistically resolved). The FineWeb-Edu-only control (\S\ref{sec:fineweb}) shows the collapse is co-mingled with generic distributional drift, so we do not attribute it specifically to coverage starvation.

\textbf{Observation 3: Every domain's own coverage marginal has an interior optimum---more is not better} (\S\ref{sec:coverage_impact}). No domain is best served by the largest share we gave it, nor by the smallest: holding the rest of the design fixed, accuracy rises with a domain's own share, peaks at a moderate value, and falls again. The pattern needs no curve fitting---pooling all trained allocations (the 24 sweep plus 6 held-out, and, once Appendix~\ref{app:expA} is included, 12 interior), the moderate band ($10$--$40\%$) is best for all five domains; this band concordance is descriptive, and a calibrated permutation test for quadratic interiority gives $P\approx0.010$ (Appendix~\ref{app:bands}). The fitted split-Gaussian curves at 8B place the peaks between $9.9\%$ and $35.1\%$. Six withheld allocations reproduce the \emph{curve shapes} out-of-sample (\S\ref{sec:coverage_impact}), closely for four domains and only approximately for Operation (the peak locations themselves are not validated on held-out data). A mid-training-only replication at 4B reproduces the pattern, though the effect is near-ceiling for Counterfactual ($1.3$\,pp band span) and the fitted peaks shift with scale ($-17.5$ to $+3.8$\,pp; Appendix~\ref{app:expB}). The claim is deliberately marginal rather than joint: the compositional response surface's stationary point reads as a saddle in every domain (Appendix~\ref{app:coda}), though this classification is provisional given the surface's parameter count.

\section{Methodology}
\label{sec:method}

We fix the planned subsequent training stages (SFT, RL) across configurations so mid-training coverage is the intended intervention. The reasoning benchmarks are described in \S\ref{sec:bench}.

\subsection{Training Stages}
\label{sec:pipeline}

We study a multi-stage pipeline consisting of mid-training followed by SFT and RL; our focus is not the recipe itself, but how the mid-training data distribution shapes later behavior.

\paragraph{Mid-training.}
Let $\theta_0$ be the base model parameters. Mid-training performs standard continued causal language modeling on a corpus $\mathcal{D}_{\mathrm{mid}}$, yielding checkpoint $\theta_{\mathrm{mid}}$, which initializes subsequent stages.

\paragraph{Supervised fine-tuning.}
SFT optimizes conditional next-token likelihood over problem-response pairs, following the standard instruction-tuning paradigm~\citep{wang2023selfinstruct}. For the Mid-training+SFT+RL configuration, SFT starts from $\theta_{\mathrm{mid}}$; for the SFT+RL baseline (no mid-training), it starts directly from $\theta_0$.

\paragraph{RL: GSPO.}
After SFT, we apply Group-based Sequence-level Policy Optimization (GSPO)~\citep{yang2025qwen3}, a sequence-level group-based RL with binary verifier rewards ($R_i=1$ if correct, $0$ otherwise); full equations are in Appendix~\ref{app:gspo}.

\subsection{Data Construction}
\label{sec:synth_method}

The pipeline (Figure~\ref{fig:data_pipeline}, Appendix~\ref{app:figures}) proceeds through seven numbered components---a \emph{rule library} (box 1), a \emph{domain dispatcher} (2), a \emph{rule synthesizer} (3), a \emph{difficulty sampler} (4), a \emph{deterministic solver} (5a), a \emph{verification} stage (6), and a \emph{data preprocessor} (7)---and in our runs all answers and traces arise from the deterministic solver; the optional \emph{teacher model} branch (box 5b) was not used here.

The training data are \emph{newly synthesised instances} derived from the benchmark rule definitions, not reused benchmark items: split isolation (by instance ID for KOR-Bench; by depth for ProofWriter) and format/information deduplication guarantee that no training instance coincides with an evaluation item (Appendix~\ref{app:decontam}).

Beyond the pipeline mechanics, two corpus-level design choices are notable. Mid-training controls token exposure rather than sample counts and uses a fixed total token budget ($\approx$1.5B tokens, two epochs; Table~\ref{tab:app-hparams}), a fixed-token mixture of the internal five-domain component and a fixed external component. The external component was generated, verified, and filtered through the same symbolic pipeline; it contains no original benchmark items, and roughly a third of its tokens (34.8\%) derive from the ProofWriter rule family, which is why we treat ProofWriter as a same-family exposure rather than a zero-exposure transfer test (\S\ref{sec:cross_domain}). The FineWeb-Edu-only baseline (\S\ref{sec:coverage_impact}) uses the same total budget with FineWeb-Edu in place of the internal component; SFT traces are generated by the same symbolic solvers---not by any LLM (token-share summary: Appendix~\ref{app:ledger}).

\subsection{Coverage Quantification}
\label{sec:coverage}

We measure per-domain mid-training exposure by token share rather than sample count, since token prediction determines the learning signal and longer-form traces would be understated by sample ratios. Domain coverage $\mathrm{Cov}(d)$ is the fraction of total mid-training tokens from domain $d$. Here coverage denotes token allocation only; it is not a direct measure of unique examples, data quality, task difficulty, or an independently manipulable dose. We track two deltas: $\Delta_{\mathrm{mid}}(d)=\mathrm{Acc}_{\mathrm{midtrain+sft+rl}}(d)-\mathrm{Acc}_{\mathrm{sft+rl}}(d)$ (net mid-training effect) and $\Delta_{\mathrm{rl}}(d)=\mathrm{Acc}_{\mathrm{midtrain+sft+rl}}(d)-\mathrm{Acc}_{\mathrm{midtrain+sft}}(d)$ (RL contribution); evaluation protocol details are in Appendix~\ref{app:training}.

\section{Experimental Setup}

\subsection{Reasoning Benchmarks}
\label{sec:bench}

\paragraph{Primary benchmark: KOR-Bench.}
We mainly use KOR-Bench, a structured benchmark for knowledge-orthogonal reasoning~\citep{ma2024korbench}. It spans five domains---ciphers, custom mathematical operations, formal logic, constraint puzzles, and counterfactual reasoning---each comprising 25 rule types with verifiable ground-truth answers across three difficulty levels.

A central property of KOR-Bench is \emph{semantic rule-disjointness}: each domain defines a self-contained symbolic system. This reduces semantic-transfer confounding but does not make the domains independent in the statistical or neural sense---they still share language, tokenization, output formats, model parameters, and potentially general reasoning procedures---so we treat per-domain coverage as a controlled mixture coordinate, not an approximately independent causal variable. Because our corpus is generated from KOR-Bench's own rule definitions (fresh instances, never the original items), KOR-Bench is our in-distribution primary metric; the external benchmarks are the only generalization probes (providing limited consistency checks).

\paragraph{External benchmarks.}
We additionally evaluate on three external benchmarks as limited consistency checks: \textit{ProofWriter}~\citep{tafjord2021proofwriter} (deductive reasoning; D5 is exposed in the fixed corpus, while D6--D9 are held out by depth), \textit{ZebraLogic}~\citep{lin2025zebralogic} Multiple Choice (MC) and Grid (constraint-satisfaction puzzles), and \textit{CounterBench}~\citep{chen2026counterbench} (zero direct KOR-Bench coverage). These benchmarks differ in exposure and structural similarity, so none is treated as a definitive independent transfer test.

\subsection{Model and Hyperparameters}
\label{sec:model_hparams}

All trainable configurations start from Qwen3-8B-Base without instruction tuning~\citep{yang2025qwen3} (scale rationale: Appendix~\ref{app:model_scale}). Mid-training uses $1\times10^{-5}$ learning rate (LR); SFT uses $5\times10^{-5}$ for 3 epochs. Both stages use full fine-tuning with LLaMA-Factory, DeepSpeed ZeRO-3~\citep{rajbhandari2020zero}, bf16, FlashAttention-2~\citep{dao2023flashattention2}, batch size 16, and sequence length 12,000 (Appendices~\ref{app:training},~\ref{app:compute}).

For the coverage sweep, each configuration is trained from five random seeds and evaluated on the held-out KOR-Bench test split; we report the mean $\pm$ 1 SD (Table~\ref{tab:mid_train}). Full-pipeline rows (Table~\ref{tab:main}) and external-benchmark measurements are likewise five-seed means with cross-seed SDs; 95\% bootstrap CIs over per-question resamples, where noted, are evaluation-sample estimates (protocol: Appendix~\ref{app:training}).

\subsection{Mid-Training Data Coverage Study}
\label{sec:exp_coverage}

We train mid-training-only checkpoints under 24 controlled domain-coverage configurations (Table~\ref{tab:mid_train}, Figure~\ref{fig:midtrain_domain_lines}), spanning the five-domain simplex, and measure evaluation accuracy before any SFT/RL intervention; a compensatory SFT pass is also applied to every configuration (\S\ref{sec:gap_closure}) to test how a fixed-budget alignment step changes the mid-training-induced differences.

Two design notes apply to Figure~\ref{fig:midtrain_domain_lines}: coverage percentages reflect \emph{internal} allocation only (the fixed external component contributes shared formal-deduction exposure, which is why Logic retains non-zero accuracy at 0\% internal coverage), and because each domain's proportion varies jointly with the other four, the curves describe mixture-level associations rather than isolated causal dose-response functions (Appendix~\ref{app:coverage_sweep}).

Each domain samples uniformly from all 25 rule types (equal token quota per rule type) with the same three-level difficulty stratification; this does not balance difficulty, answer length, or trace length across configurations.

\subsection{Pairwise Gap-Closure Analysis}
\label{sec:gap_closure}

We apply compensatory SFT to all 24 checkpoints: it reweights SFT data toward coverage-deficient domains under a fixed budget, $r_d \propto \max(0,\,60\%-M_d)$ (Appendix~\ref{app:comp_sft}). We measure the cross-configuration accuracy range before and after compensation and summarize pairwise gap closure over the $24\times\binom{5}{2}=240$ unordered domain pairs (Appendix~\ref{app:pairwise}); because the configurations are purposively selected design points, these counts are descriptive.

\subsection{Full Training Pipeline Comparison}

Table~\ref{tab:main} compares three shared no-mid-training baselines---Base, SFT only, SFT+RL---against the three mid-training stages for each of the three experimental allocations. Because the SFT data is identical (balanced by rule type) across all SFT-bearing rows, Mid-training+SFT vs.\ SFT only isolates the effect of inserting a particular mid-training checkpoint before the same SFT stage; the analogous comparison after RL is Mid-training+SFT+RL vs.\ SFT+RL (Table~\ref{tab:survival}). Qwen3-8B-Instruct~\citep{yang2025qwen3} is not trained in our pipeline and is reported only as a non-comparable external reference (Appendix~\ref{app:instruct}).

\section{Results}

\subsection{The Impact of Mid-Training Data Coverage}
\label{sec:coverage_impact}

\begin{figure}[!htbp]
\centering
\includegraphics[width=0.68\textwidth]{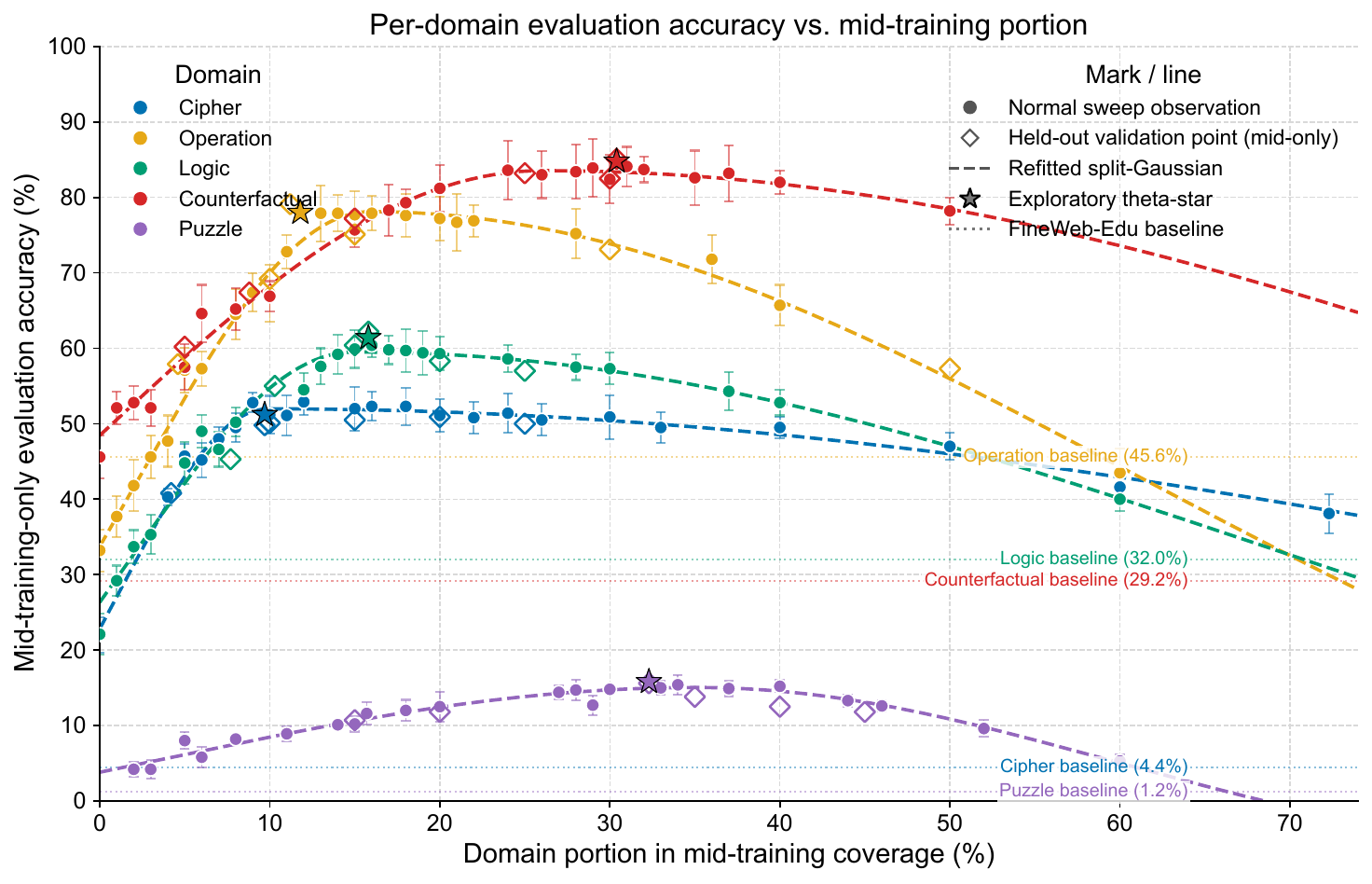}
\caption{Per-domain mid-training-only accuracy vs.\ the domain's own mid-training portion. Filled points: 23 of the 24 normal configurations (Table~\ref{tab:mid_train}; 19/7/19/24/31 omitted for legibility, retained in all fits). Hollow diamonds: six held-out allocations, withheld from the fit, roughly tracking the fitted curves for Cipher, Logic, Counterfactual, and Puzzle (Operation's low-coverage points sit $\approx6$\,pp above the fitted tail). Vertical bars: $\pm1$ SD across five seeds. Dashed lines: split-Gaussian fits; stars: exploratory $\theta^*$; dotted lines: FineWeb-Edu baselines.}
\label{fig:midtrain_domain_lines}
\end{figure}

Figure~\ref{fig:midtrain_domain_lines} plots per-domain evaluation accuracy against each domain's own mid-training proportion across the 24 normal configurations (\S\ref{sec:exp_coverage}); the FineWeb-Edu baseline and $\theta^*$ checkpoint are visual references, excluded from fitting.

We now ask where each domain's optimum actually sits. We fit the empirical coverage--accuracy points with parametric curves; the fits locate the optima, but nothing in the moderation claim itself depends on them.

\paragraph{Out-of-sample validation on held-out allocations.}
To probe whether the fitted curves and the compensatory-SFT result describe more than the 24 fitting points, we trained six additional allocations withheld from the fit and carried them through the full pipeline (Table~\ref{tab:mid_train}). The split-Gaussian fits generalize in all five domains (held-out residuals $1.0$--$1.8$\,pp for Counterfactual, Puzzle, Cipher, and Logic; Operation is the weakest match at $3.0$\,pp), and the compensatory-SFT result reproduces: gains in all 30 cells (mean $+4.5$\,pp) yet $0/60$ pairs closed at $5$\,pp and $5/60$ at $10\%$, and carrying these checkpoints through RL adds $+1.19$\,pp on average. The held-out allocations also test the premise underlying $\theta^*$: fitting on the 24 sweep configurations alone predicts their post-RL overall accuracy with $r=+0.967$ ($p=0.002$), whereas the relative-gain objective that actually selects $\theta^*$ ranks them only weakly ($r=+0.441$). The fitted curves are the trustworthy object; an allocation rule should optimise predicted accuracy directly (Appendix~\ref{app:oos}).

\paragraph{Exploratory marginal fits.}
Each domain's curve is fitted by an asymmetric split Gaussian (five parameters; definition and LOO-CV in Appendix~\ref{app:fitting}). With the refreshed 24-point sweep the fitted peaks are approximately Cipher 9.9\%, Operation 15.9\%, Logic 15.0\%, Counterfactual 26.0\%, and Puzzle 35.1\%, with the weakest quadratic signal for Cipher ($R^2_{\mathrm{quad}}\approx0.49$). Because the simplex constraint couples the proportions, these peaks are descriptive mixture-level summaries whose sum exceeds the 100\% budget; we therefore derive an illustrative allocation $\theta^*$ (Eq.~\ref{eq:opt}), whose fitted objective evaluates to $\approx+9.3\%$ relative gain over balanced, not an achieved gain: the re-trained $\theta^*$ checkpoint realizes $58.24$ vs.\ balanced $56.26$ at mid-training-only and $+4.36$\,pp over SFT+RL at the full-pipeline stage (Table~\ref{tab:main}).

\paragraph{Fitted 95\%-of-peak intervals.}
Table~\ref{tab:coverage_bands} (Appendix~\ref{app:fitting}) reports, for each domain, the set of $\theta_d$ where the fitted curve is within 95\% of its fitted peak (derivation in Appendix~\ref{app:fitting}). These are descriptive function-level thresholds for the present sweep; they are not confidence intervals and not validated allocation rules.

\subsection{Vertical Comparison: Mid-Training Contribution Across Stages}
\label{sec:stages}

We now test whether the observed mixture-associated differences change after SFT and RL; the selected full-pipeline comparisons are exploratory.

\begin{figure}[!htbp]
\centering
\includegraphics[width=\textwidth]{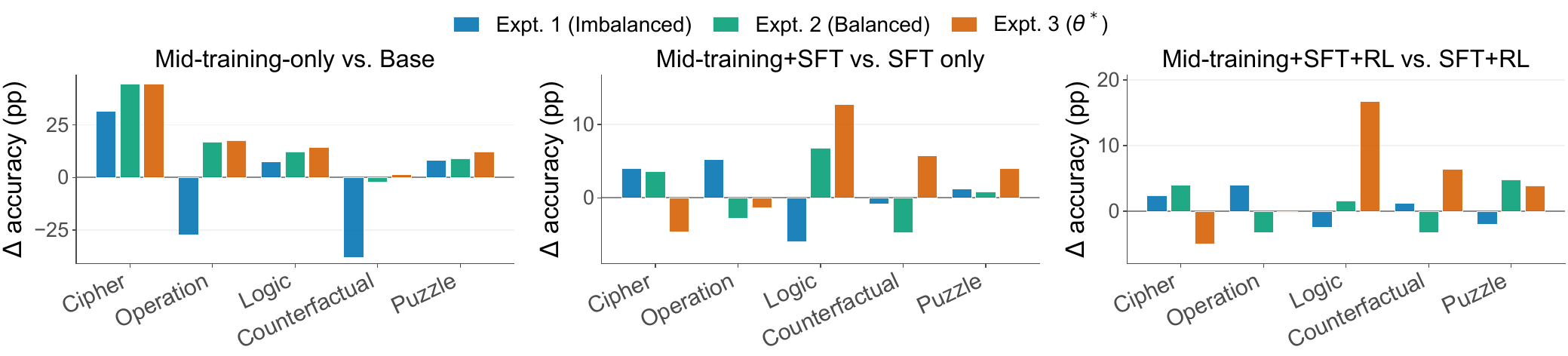}
\caption{Pipeline-stage breakdown: per-domain accuracy gain ($\Delta$, pp) across three stages (data from Table~\ref{tab:main}). \textbf{Left:} Mid-training-only vs.\ Base ($\Delta = \mathrm{Acc}_\mathrm{mid} - \mathrm{Acc}_\mathrm{base}$)---positive for domains where the base prior is low (Cipher, Puzzle); small or negative for Counterfactual, where the base model already achieves 83.6\%. \textbf{Middle:} Mid-training+SFT vs.\ SFT only. \textbf{Right:} Mid-training+SFT+RL vs.\ SFT+RL. Expt.\ 1 (imbalanced), Expt.\ 2 (balanced), Expt.\ 3 ($\theta^*$).}
\label{fig:pipeline_comparison}
\end{figure}

We select two representative sweep configurations and add one derived analytically from the fitted curves, carrying all three through the identical full SFT+RL pipeline: \textbf{Expt.\ 1} (imbalanced: Cipher 72.3\%, Operation 0\%, Logic 12.0\%, Counterfactual 0\%, Puzzle 15.7\%), \textbf{Expt.\ 2} (balanced: 20\% per domain), and \textbf{Expt.\ 3} (exploratory $\theta^*$; \S\ref{sec:coverage_impact})---a new allocation trained as a dedicated checkpoint. The full-pipeline comparison is restricted to three configurations for compute reasons; they span a coverage contrast but are not representative of the sweep (Table~\ref{tab:main}, Figure~\ref{fig:pipeline_comparison}).

\begin{table}[!htbp]
\centering
\footnotesize
\setlength{\tabcolsep}{4pt}
\begin{tabular*}{\textwidth}{@{\extracolsep{\fill}}lcccccc@{}}
\toprule
\textbf{Config} & \textbf{Overall} & \textbf{Cipher} & \textbf{Oper.} & \textbf{Logic} & \textbf{Counterf.} & \textbf{Puzzle} \\
\midrule
\multicolumn{7}{@{}p{0.85\textwidth}@{}}{\textit{No-mid-training baselines (shared across experiments)}} \\
\rowcolor{rowBase}\quad Base & $40.32\pm2.7$ & $6.8\pm2.5$ & $60.4\pm2.9$ & $47.2\pm1.9$ & $83.6\pm3.0$ & $3.6\pm2.0$ \\
\rowcolor{rowSFT}\quad SFT only & $64.08\pm1.8$ & $66.0\pm2.0$ & $88.8\pm2.7$ & $59.2\pm2.3$ & $86.4\pm3.0$ & $20.0\pm2.0$ \\
\rowcolor{rowRL}\quad SFT+RL & $65.28\pm1.9$ & $68.0\pm2.3$ & $88.8\pm3.0$ & $59.2\pm2.2$ & $87.6\pm3.0$ & $22.8\pm1.5$ \\
\midrule
\multicolumn{7}{@{}p{0.85\textwidth}@{}}{\textit{Expt.\ 1 --- Imbalanced: Cipher 72.3\%, Oper.\ 0\%, Logic 12.0\%, Counterf.\ 0\%, Puzzle 15.7\%}} \\
\rowcolor{rowBase}\quad Mid-training only & $36.60\pm1.8$ & $38.1\pm2.6$ & $33.2\pm2.8$ & $54.5\pm2.2$ & $45.6\pm2.8$ & $11.6\pm1.5$ \\
\rowcolor{rowSFT}\quad Mid-training+SFT & $64.80\pm2.5$ & $70.0\pm2.2$ & $94.0\pm2.7$ & $53.2\pm1.9$ & $85.6\pm2.8$ & $21.2\pm1.4$ \\
\rowcolor{rowRL}\quad Mid-training+SFT+RL & $65.92\pm1.5$ & $70.4\pm2.1$ & $92.8\pm3.0$ & $56.8\pm2.0$ & $88.8\pm3.5$ & $20.8\pm2.0$ \\
\midrule
\multicolumn{7}{@{}p{0.85\textwidth}@{}}{\textit{Expt.\ 2 --- Balanced: 20\% per domain (4/5 domains within fitted 95\%-of-peak intervals)}} \\
\rowcolor{rowBase}\quad Mid-training only & $56.26\pm1.5$ & $51.1\pm2.2$ & $77.2\pm2.9$ & $59.3\pm2.2$ & $81.2\pm3.1$ & $12.5\pm2.0$ \\
\rowcolor{rowSFT}\quad Mid-training+SFT & $64.80\pm1.7$ & $69.6\pm2.1$ & $86.0\pm2.9$ & $66.0\pm2.1$ & $81.6\pm2.9$ & $20.8\pm1.9$ \\
\rowcolor{rowRL}\quad Mid-training+SFT+RL & $66.08\pm2.3$ & $72.0\pm2.3$ & $85.6\pm3.2$ & $60.8\pm2.1$ & $84.4\pm3.3$ & $27.6\pm1.5$ \\
\midrule
\multicolumn{7}{@{}p{0.85\textwidth}@{}}{\textit{Expt.\ 3 --- Exploratory $\theta^*$: Cipher 9.7\%, Oper.\ 11.8\%, Logic 15.8\%, Counterf.\ 30.4\%, Puzzle 32.3\% (\textbf{four of five} within fitted 95\%-of-peak intervals; Operation's 11.8\% sits just below its re-fitted lower bound of 11.9\%). Re-trained full pipeline.}} \\
\rowcolor{rowBase}\quad Mid-training only & $58.24\pm1.5$ & $51.2\pm2.1$ & $78.0\pm2.8$ & $61.4\pm2.1$ & $84.8\pm3.1$ & $15.8\pm1.9$ \\
\rowcolor{rowSFT}\quad Mid-training+SFT & $67.34\pm2.1$ & $61.3\pm2.2$ & $87.4\pm2.9$ & $71.9\pm2.3$ & $92.1\pm3.5$ & $24.0\pm1.9$ \\
\rowcolor{rowRL}\quad Mid-training+SFT+RL & $69.64\pm2.6$ & $63.0\pm2.3$ & $88.6\pm2.5$ & $75.9\pm2.2$ & $94.0\pm3.3$ & $26.7\pm1.3$ \\
\bottomrule
\end{tabular*}%
\caption{KOR-Bench zero-shot accuracy (\%). Base, SFT only, and SFT+RL are shared no-mid-training baselines. Expt.\ 1--2 use selected sweep allocations and Expt.\ 3 uses the re-trained exploratory $\theta^*$ allocation; these comparisons probe, rather than establish, the effect of coverage. Values are five-seed means $\pm$ 1\,SD. The fitted 95\%-of-peak intervals are descriptive (Table~\ref{tab:coverage_bands}); the non-comparable Qwen3-8B-Instruct reference is in Appendix~\ref{app:instruct}. Overall-column differences among the three full-pipeline rows are at best nominally significant under uncorrected Welch $t$-tests (\S\ref{sec:stages}).}
\label{tab:main}
\end{table}

Table~\ref{tab:main} reports the results. The Base model shows strong domain-differentiated priors (83.6\% Counterfactual vs.\ 3.6\% Puzzle); mid-training is associated with the largest mid-training-only gains where the prior is low and small or negative differences where it is already high (Figure~\ref{fig:pipeline_comparison}, left). The incremental value of mid-training is descriptively largest for \textbf{Expt.\ 3} ($\theta^*$): $+4.36$\,pp overall vs.\ SFT+RL, vs.\ $+0.80$\,pp (balanced) and $+0.64$\,pp (imbalanced). Welch $t$-tests give $t\approx2.29$ ($p\approx0.052$) and $t\approx2.77$ ($p\approx0.030$) for $\theta^*$ vs.\ balanced/imbalanced; neither survives Bonferroni correction, so the ordering is descriptive throughout. The gain is domain-concentrated: Logic $+16.7$\,pp and Counterfactual $+6.4$\,pp, while Cipher is $-5.0$\,pp despite its 9.7\% allocation sitting near the fitted mid-training-only peak---$\theta^*$ trades Cipher exposure for Logic and Counterfactual coverage. The $+4.36$ vs.\ $+0.80$\,pp gap is of the same order as the rows' own overall SDs, so no per-domain $\Delta$ is claimed as individually significant. Expt.\ 3 shows the largest RL contribution ($+2.30$\,pp vs.\ $+1.12$ and $+1.28$), a secondary observation (Appendix~\ref{app:rl_budget}); its final overall score (69.64) exceeds the non-comparable Qwen3-8B-Instruct reference (63.80).

\paragraph{How much of the mid-training-only collapse survives alignment?}
The interference emphasized above is a \emph{mid-training-only} observation. Under Expt.\ 1 the two largest collapses are reversed in sign (Table~\ref{tab:survival}): Operation ($-27.2$\,pp vs.\ Base at mid-only) reaches $+5.2$\,pp over its SFT-only baseline at mid+SFT, a clear reversal; Counterfactual ($-38.0$\,pp) is still $+0.8$\,pp below at mid+SFT and only turns positive ($+1.2$\,pp) by the RL row, within its SD. What persists is a small residual deficit on Logic and Puzzle ($-2.4$ and $-2.0$\,pp). The interference effect (Observation~2) is thus a transient, checkpoint-level effect that the tested recipe partly repairs.

\subsection{Horizontal Comparison: Can Alignment Close Mid-Training Gaps?}

We test whether reweighting SFT data toward under-covered domains can correct the imbalance installed at mid-training, using two variants: compensatory SFT (weighted toward coverage-deficient domains under a fixed token budget) and uniform SFT (identical data mix, serving as a baseline). Both passes are described here and in Appendix~\ref{app:comp_sft}.

\begin{figure}[!htbp]
\centering
\includegraphics[width=0.58\textwidth]{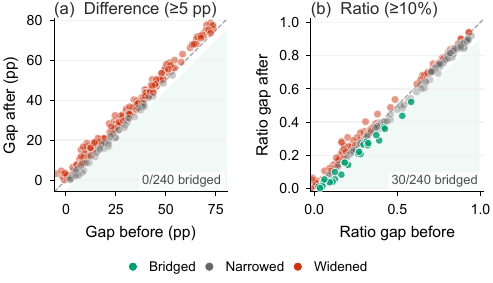}
\caption{Pairwise gap closure after compensatory SFT across the 24 coverage-sweep checkpoints. \textbf{(a)} Difference metric (pp), threshold $\ge 5$\,pp closure. \textbf{(b)} Ratio metric, threshold $\ge 10\%$ relative closure. Points below the diagonal indicate gap reduction: $0/240$ pairs bridged under the difference metric, $30/240$ under the ratio metric. Counts are descriptive because the pairs share the 24 configurations.}
\label{fig:comp_sft_pairwise_closure}
\end{figure}

\paragraph{Compensatory SFT.}
Applying the fixed formula-derived compensation policy (\S\ref{sec:gap_closure}) to all 24 checkpoints raises absolute scores in 116 of 120 configuration--domain cells (mean $+4.32$\,pp; the four cells without gain all receive zero compensatory allocation $r=0$). The cross-configuration accuracy range narrows for Cipher (30.9$\rightarrow$30.5\,pp), Operation (44.7$\rightarrow$42.5\,pp), Logic (38.3$\rightarrow$37.8\,pp), and Counterfactual (38.5$\rightarrow$36.2\,pp), while Puzzle widens only slightly. Pairwise closure over the 240 domain pairs is similarly limited: 30/240 under a $10\%$ relative-ratio threshold and 0/240 under a $5$\,pp difference threshold (Appendix~\ref{app:pairwise}, Figure~\ref{fig:comp_sft_pairwise_closure}); under the budget-proportional model the within-configuration gain differential is moreover arithmetically bounded (at most $\approx7$\,pp), so the tested family has limited dynamic range by construction. The same pattern holds on the six held-out allocations (\S\ref{sec:coverage_impact}).

\paragraph{Uniform SFT control.}
A uniform SFT pass (identical per-domain mix across configurations; three epochs) was also applied to all 24 checkpoints (Eval.\ (Uniform) rows in Table~\ref{tab:mid_train}; Appendix~\ref{app:comp_sft}). Its behaviour is close to that of the compensatory pass: gains in all 120 cells (mean $+4.20$\,pp vs.\ $+4.32$\,pp), comparable range narrowing, and pairwise closure 0/240 under $5$\,pp and 32/240 under $10\%$ (vs.\ 0/240 and 30/240). Per-domain mean gains show where reweighting actually differs (Table~\ref{tab:comp_uniform}): compensation is materially stronger only for Counterfactual ($+6.37$ vs.\ $+4.49$\,pp), roughly equal for Operation and Logic, and slightly weaker for Cipher and Puzzle. The failure to close pairwise gaps is a property of the tested SFT data and budgets.

\subsection{External Evaluation and Limited Consistency Checks}
\label{sec:cross_domain}

The gap pattern in \S\ref{sec:stages} may reflect benchmark-specific overlap or shared task structure. We therefore evaluate a fixed subset of the coverage configurations on three additional benchmarks---ProofWriter~\citep{tafjord2021proofwriter}, ZebraLogic~\citep{lin2025zebralogic}, and CounterBench~\citep{chen2026counterbench}---as limited consistency checks: the nine coverage configurations of Table~\ref{tab:ext_sweep} (Appendix~\ref{app:external}). These evaluations do not establish general transfer: the external corpus draws 34.8\% of its tokens from the ProofWriter rule family, the ZebraLogic family contributes 201 mid-training samples, and the available split controls are primarily instance/depth based.

\paragraph{ProofWriter.}
ProofWriter is stratified by proof depth (D5--D9); D5-family items appear in the external mid-training corpus while D6--D9 are held out by depth, so the D6 comparison is best interpreted as same-benchmark held-out-depth evaluation. The reported gain (Expt.\ 1 checkpoint; Appendix~\ref{app:external}) is $\Delta=+6.4$ with 95\% CI $[+3.3,+9.5]$; D7--D9 gains are positive but based on small strata ($N\le120$). Across the nine configurations of Table~\ref{tab:ext_sweep}, the rank correlation between the KOR average and ProofWriter accuracy is small and positive ($\rho=+0.34$, $n=9$; descriptive).

ZebraLogic Grid shows a positive ranking association with the KOR-Bench pattern ($\rho=+0.53$), whereas ZebraLogic MC does not ($\rho=-0.12$); CounterBench, with zero direct mid-training coverage and uniformly non-positive per-type deltas, correlates positively with the KOR average ($\rho=+0.67$). These Spearman correlations are descriptive ($n=9$), not evidence of a transfer mechanism. Overall, configuration differences are not entirely confined to one benchmark, while shared structure and exposure remain possible explanations.

\section{Discussion}
\label{sec:discussion}

The first finding is that the coverage-induced gaps are robust to the alignment budget in a specific, testable sense: a fixed-budget compensatory SFT pass leaves them essentially intact, and an equal-budget uniform-SFT control behaves almost identically, so the non-closure is not a property of the compensation formula. Three further results locate where it comes from: the permutation null (Appendix~\ref{app:pairwise}) shows the gains are placed in a gap-preserving way; the sharpening sweep (Appendix~\ref{app:sharpen}) shows closure is attainable, but only by trading roughly half the mean gain for it; and the six held-out allocations reproduce the whole pattern out-of-sample, including after RL. What this does not license is the general claim that alignment cannot repair coverage choices: we tested one policy family, one budget, and a short RL leg ($+1.1$ to $+2.3$\,pp; Appendix~\ref{app:rl_budget}). Two observations qualify the picture. First, checkpoint-level negative transfer: zero coverage collapses mid-training-only accuracy even where the base model already performs well; the tested recipe partly repairs this (Table~\ref{tab:survival}), and the FineWeb-Edu-only control (\S\ref{sec:fineweb}) shows the collapse is co-mingled with generic distributional drift, so we frame gradient dominance~\citep{yu2020pcgrad} and drift as competing hypotheses. Second, per-domain coverage is non-monotonically associated with accuracy, forming inverted-U curves with fitted peaks near moderate shares: a moderation-like mixture-level association, not an identified dose-response law.

These patterns are confined to the tested KOR-Bench/Qwen3-8B-Base setting, extending~\citet{zhang2025interplay}.

\paragraph{What would disconfirm the headline claim.} Two of the three observations are fragile by construction: the ``inverted-U'' and the ``interference'' rest on mid-training-only checkpoints under a simplex constraint (\S\ref{sec:simplex}--\ref{sec:fineweb}). The observation a single experiment could disconfirm is the \emph{structural negative result} (Observation~1): if a fixed-budget compensatory-SFT pass reweighting toward coverage-deficient domains fully closed the between-domain gaps (closure approaching 100\% under the $5$\,pp metric rather than 0/240), our claim would be refuted.

\subsection{Why the Tested Alignment Passes Did Not Close the Gaps}
\label{sec:rl_limits}

These observations are compatible with several explanations: budget competition, domain-specific diversity, optimization mismatch, limited post-training exploration, and the arithmetic cap of the compensation formula ($\approx7$\,pp by construction; Appendix~\ref{app:comp_sft}). A capability-envelope account remains a hypothesis: this study does not measure representations, gradient conflict, or parameter overlap.

\subsection{The Simplex Confound}
\label{sec:simplex}

A structural limit of this design deserves explicit treatment. Coverage is a zero-sum allocation: the five domain proportions always sum to $100\%$, so a domain's share cannot be varied independently of the other four, and an unvarying external corpus ($34.8\%$ ProofWriter) contributes shared exposure held constant across runs. Consequently, each per-domain curve is a \emph{mixture-level marginal} rather than a per-domain dose-response, and the inverted-U ``peak'' a domain shows is a property of where the simplex balances against the fixed external component, not an isolated property of that domain's coverage: ``coverage'' is not a cleanly manipulable dose. Separating them requires a simplex-aware joint response-surface model, which we ran (Appendix~\ref{app:coda},~\ref{app:expA}): mapping allocations to isometric log-ratio coordinates and fitting a per-domain quadratic surface fits well, but the stationary point reads as a \emph{saddle} in every domain, and remains one over the 42-allocation pool (30 base plus the 12 interior allocations). We stress this classification is provisional---with 15 quadratic terms estimated from 42 allocations the Hessian is too weakly determined to separate a saddle from a shallow flat region (Appendix~\ref{app:coda})---so the interior data neither establish a jointly optimal mixture nor exclude a shallow one. The five per-domain optima of Observation~3 are therefore marginal statements, not the coordinates of one jointly optimal mixture.

\subsection{A Control Confound: Mid-Training on Unrelated Data Also Degrades Out-of-Scope Reasoning}
\label{sec:fineweb}

The interference interpretation relies on the claim that ``starving a KOR-Bench domain'' is what hurts it. The FineWeb-Edu-only control (Table~\ref{tab:mid_train}) challenges this framing: mid-training on FineWeb-Edu alone---\emph{no} KOR-Bench data at all---drops Counterfactual from its Base prior of 83.6\% to 29.2\% ($-54.4$\,pp), a \emph{larger} drop than the zero-KOR-coverage Expt.\ 1 case ($-38.0$\,pp), and drops Operation from 60.4\% to 45.6\% ($-14.8$\,pp). The interference the sweep attributes to selectively starving a KOR-Bench domain is thus co-mingled with a broader effect: any continued pre-training, including on text wholly unrelated to the benchmark, degrades out-of-scope reasoning. We frame the interference finding as a within-setting observation.

\subsection{Implications for Design}
\label{sec:moderate}

Because the simplex confound is unresolved and no held-out allocation validates the fitted peaks, intervals, or $\theta^*$, these results support a hypothesis to test rather than a rule to follow: they do not establish that a domain should be kept ``in moderation,'' nor that later alignment generally cannot correct coverage choices.
\section{Conclusion}
\label{sec:conclusion}

Three findings hold in this setting. First, coverage-induced domain gaps survive the finite-budget post-training procedures we tested: the compensatory policy and an equal-budget uniform control both leave the gaps essentially intact (0/240 bridged pairs under the $5$\,pp metric), this generalizes to six held-out allocations, and a permutation null ($P<0.001$; Appendix~\ref{app:pairwise}) shows the gains are placed in a gap-preserving way: budget placement, not only its size, preserves the gaps. Second, zero mid-training coverage collapses mid-training-only accuracy even with a high prior, but the tested recipe reverses the sign of these collapses (Table~\ref{tab:survival}), leaving a small residual deficit on Logic and Puzzle, and the FineWeb-Edu-only control shows this is co-mingled with generic distributional drift. Third, every domain has an interior coverage optimum: the moderate band yields higher mean accuracy than the low or high band for all five domains, with no curve fitting involved (Appendix~\ref{app:bands}), and the fitted peaks lie between $9.9\%$ and $35.1\%$. These optima are per-domain marginals, not a jointly achievable mixture: the compositional response surface's stationary point reads as a saddle in every domain, even after 12 interior allocations are added (Appendix~\ref{app:coda},~\ref{app:expA}), a classification that is provisional rather than decisive---and causal attribution remains out of reach. The practical implication is conditional: coverage allocation may warrant explicit auditing, but the $\theta^*$ gain ($+4.36$\,pp) is marginal under an uncorrected Welch test and remains a model-selected candidate from the same sweep (scope and limits: Appendix~\ref{app:limitations}).


\appendix

\section{Related Work}
\label{app:related}

\subsection{Multi-Stage Training for LLM Reasoning}

Modern reasoning models are typically trained through a sequence of continued pretraining, SFT, and RL. The foundational Reinforcement Learning from Human Feedback (RLHF) framework~\citep{ouyang2022instructgpt} established the standard three-stage recipe of pretraining, supervised fine-tuning on demonstrations, and RLHF, which most subsequent systems adopt. \citet{guo2025deepseekr1} show that RL can elicit strong reasoning behavior when applied to a capable base model, while their full DeepSeek-R1 pipeline still relies on cold-start SFT and additional supervised data. \citet{shao2024deepseekmath} demonstrate that domain-specific continued pretraining can substantially improve mathematical reasoning before later post-training. Kimi k1.5 further scales reinforcement learning in a multi-stage setting~\citep{teamkimi2025kimi}. These works establish the effectiveness of stage-wise training, but they usually report aggregate benchmark scores and do not isolate how mid-training data imbalance shapes domain-level outcomes.

Broader model reports also reinforce the importance of training stages and data coverages. Llama~3 emphasizes the role of large-scale pretraining and post-training recipes~\citep{dubey2024llama3}. InternLM2~\citep{cai2024internlm2} explicitly incorporates a progressive training strategy with carefully curated mid-training data for reasoning, making domain allocation a first-class engineering concern. Phi-4 highlights the importance of high-quality synthetic and curated data for reasoning-oriented models~\citep{abdin2024phi4}. Our work differs from these system reports by treating mid-training data composition itself as the object of study.

\subsection{The Role of Mid-Training}

Mid-training is increasingly recognized as a stage with its own objectives and failure modes. \citet{tu2025midtrain} survey recent work and frame mid-training as a bridge between broad pretraining and downstream alignment. \citet{zhang2025interplay} provide controlled evidence that earlier training exposure determines how much later RL can generalize. Their finding motivates our central question, but their setting is deliberately controlled and synthetic. We instead examine a naturally imbalanced multi-domain corpus.

Several studies suggest that continued pretraining can inject useful reasoning structure. \citet{ruis2024procedural} argue that procedural knowledge in pretraining data drives reasoning performance. \citet{ishibashi2025mining} study continual pretraining with synthetic data for reasoning. \citet{yu2023metamath} show that bootstrapped mathematical data can improve math reasoning. In a related direction, \citet{lu2024mathcoder2} use model-translated mathematical code for continued pretraining. The clearest single-domain exemplar of this approach is Llemma~\citep{azerbayev2023llemma}, which continues pretraining Code Llama on the 55B-token Proof-Pile-2 corpus and achieves substantial gains on mathematical benchmarks; this work establishes that domain-specific continued pretraining on curated reasoning corpora is a reliable path to capability specialization. Our work extends this line to a multi-domain setting where allocation across domains is the primary planned coordinate. The resulting coverage-sensitive patterns are exploratory and cannot, under the present simplex design, be interpreted as an independently identified domain-specific dose-response effect.

The over-coverage half of the dose-response effect may be related to interference in multi-task optimisation~\citep{french1999catastrophic,kirkpatrick2017ewc,yu2020pcgrad}. In sequential continual learning, training intensively on a new task causes gradient updates to overwrite the parameter regions encoding previously learned tasks---a phenomenon termed catastrophic forgetting~\citep{french1999catastrophic,luo2023forgetting}. Scaling and data-curation strategies for continual pre-training have been studied as mitigation levers~\citep{ibrahim2024continual}, but they address sequential adaptation rather than allocation among simultaneous task domains. Our setting differs structurally: all domains are trained simultaneously rather than sequentially. However, imbalanced coverage may produce an analogous gradient-dominance effect: the high-coverage domain contributes the majority of the gradient signal, pushing shared representations toward its objectives and reducing the signal from low-coverage domains. We do not directly measure gradient conflict, parameter overlap, or Elastic Weight Consolidation (EWC)-style mitigation, so this explanation should be read as a hypothesis rather than a demonstrated mechanism.

The KOR-Bench domains have \emph{semantic rule-disjointness}: each domain defines a self-contained symbolic system whose rules do not rely on concepts or operations from other domains. This reduces one source of semantic-transfer confounding, but it does not make domains independent in the statistical or neural sense. Domains still share language, tokenization, output formats, model parameters, and potentially general reasoning procedures. We therefore treat the proportions as a controlled mixture coordinate, not as an independent causal dose. The present design does not separately identify coverage from unique-example diversity, repetition, sequence length, or all external-corpus exposure.

The quality and format of mid-training data matter as much as the quantity. \citet{wei2022cot} demonstrate that chain-of-thought reasoning traces—step-by-step explanations provided as training signal—substantially improve multi-step reasoning in large models; our pipeline embeds analogous structured traces in every generated instance. \citet{luo2023wizardmath} show that instruction evolution (Evol-Instruct) applied to mathematical seed problems produces progressively harder variants that boost fine-tuned model performance, demonstrating that data quality amplification can substitute for raw volume. \citet{toshniwal2024openmathinstruct} scale this approach with OpenMathInstruct-1, a 1.8M-problem corpus synthesized by Mixtral-8x7B targeting the MATH benchmark~\citep{hendrycks2021math}, which has become a canonical measure of mathematical reasoning capability. These studies collectively show that structured, verified, trace-annotated problems are the fundamental currency of mid-training effectiveness—a principle our multi-domain study extends to five heterogeneous reasoning types.

Data composition is another important thread. \citet{yang2024fineline} analyze how pretraining choices affect downstream capabilities. \citet{qin2026davinci} argue that pretraining establishes capability ceilings that later stages may not easily exceed. Xiaomi's MiMo report also emphasizes unlocking reasoning potential through carefully designed training data~\citep{xiaomi2025mimo}. Our study adds a domain-level view: the question is not only whether mid-training helps overall, but which domains benefit and which domains are left behind.

\subsection{Data Mixture and Coverage Optimization}

The question of how to allocate data across domains is well-studied in the pretraining literature. \citet{gururangan2020dapt} establish that domain-adaptive continued pretraining on even modest domain-specific data substantially improves downstream task performance, and that the benefit scales with the mismatch between the general-web pretraining distribution and the target domain. \citet{xie2024doremi} propose DoReMi, a Group Distributionally Robust Optimization (DRO)-based method that dynamically reweights domain contributions to equalize per-domain excess loss; their framework directly addresses the uniform-vs.-non-uniform weighting question that our sweep operationalizes empirically. \citet{fan2024doge} extend this line with DoGE, which reweights domains by their estimated contribution to a generalization objective via a small proxy model; DoGE and our $\theta^*$ both produce a single recommended allocation, but from opposite directions---DoGE optimizes a gradient-based generalization estimate without training at the target allocation, whereas $\theta^*$ is read off measured per-domain accuracy curves. A controlled comparison on the same five-domain corpus would be informative and is not attempted here.

At the corpus level, large-scale composition ablations in the Pile~\citep{gao2021pile} and Dolma~\citep{soldaini2024dolma} demonstrate that data-source mix is among the strongest predictors of benchmark performance. Complementary work addresses \emph{which} data to include rather than how to weight each domain. \citet{xie2023dsir} propose Data Selection via Importance Resampling (DSIR), selecting pretraining documents whose n-gram distributions match a target domain, effectively enabling soft domain filtering without hard corpus boundaries. \citet{shen2023slimpajama} construct SlimPajama-DC by rebalancing and deduplicating the RedPajama corpus, showing that domain rebalancing at the corpus level yields consistent downstream improvements. \citet{muennighoff2023scaling} examine the complementary question of data repetition: when a domain corpus is small, training for multiple epochs can partially substitute for additional data, though with diminishing returns; this bound is relevant when interpreting our low-coverage configurations. More fine-grained theoretical treatments have followed: \citet{ye2024datamixinglaws} establish quantitative scaling laws that predict language modeling loss as a function of domain mixture proportions, and \citet{liu2024regmix} propose RegMix, which trains 512 small proxy models across diverse domain combinations and fits a regression predictor to identify the optimal large-scale mixture---a principled complement to our empirical sweep. Both methods optimize for aggregate perplexity; our contribution is a domain-resolved diagnostic of how finite-budget mixture configurations are associated with downstream accuracy in this controlled setting, not a general dose-response law. More recent mixture-optimization methods such as MergeMix~\citep{wang2026mergemix} propose using model merging weights as a low-cost proxy for domain performance---a different optimization strategy. A controlled comparison of these methods in the present five-domain setup is future work; the current sweep does not validate their recommended allocations or the local moderate-band guidance.

\subsection{SFT and RL: Stage Combination and Refinement}

SFT and RL are often treated as mechanisms for alignment and refinement rather than as sources of entirely new capabilities. \citet{havrilla2024teaching} observe that RL can struggle to explore far beyond solutions already reachable by the supervised model. \citet{chen2024spin} show that self-play fine-tuning can improve weak models but also converges without continually expanding the data distribution. These findings are consistent with the idea that later stages inherit constraints from earlier stages.

Recent work further studies how SFT and RL should be combined. \citet{wang2025rlsr} compare reinforcement learning and supervised training in instruction following. \citet{chen2025sasr} propose step-wise adaptive integration of SFT and RL. \citet{deng2025supervisedrl} connect expert trajectories with step-wise reinforcement learning. \citet{ren2026rethinking} revisit generalization in reasoning SFT, while \citet{matsutani2025rl} contrasts the effects of SFT and RL on reasoning models. Two foundational contributions bound the effectiveness of post-pretraining stages. LIMA~\citep{zhou2023lima} demonstrates that as few as 1{,}000 carefully curated demonstrations can match much larger SFT corpora in instruction-following quality, implying that the capability headroom pre-established by earlier training stages is the primary bottleneck rather than fine-tuning volume. Direct Preference Optimization (DPO)~\citep{rafailov2023dpo} reformulates preference alignment as contrastive supervised learning, removing the reward model and substantially lowering the engineering complexity of the RL stage; the wider adoption of DPO has made controlled earlier-stage ablations such as ours easier to reproduce. Our experiments address both directions of this question. First, later stages do not reliably compensate for domain imbalance introduced at mid-training: even when SFT data is augmented toward coverage-deficient domains, difference-based pairwise gaps fail to close beyond the $5$\,pp repair threshold (0/240 pairs under a $5$\,pp difference metric), and ratio-based gaps show only limited average closure (30/240 pairs under a $10\%$ ratio). An equal-budget uniform-SFT control behaves almost identically (0/240 and 32/240 under the same metrics; Appendix~\ref{app:comp_sft}), so the non-closure is not an artifact of the compensation formula. Per-domain fits of the gain against the remedial allocation ($\Delta\approx a\log(1+b\,r)+c$) rise within each domain yet saturate at domain-specific ceilings, and pooling across domains explains little of the cell-level variance ($R^2\approx0.31$; Figure~\ref{fig:sft_amplification}, Table~\ref{tab:comp_logfit}): redirecting SFT data does not reset the mid-training starting point. Second, optimizing mid-training coverage makes later stages materially more effective when coverage places most domains within their fitted 95\%-of-peak intervals: mid-training adds $+4.36$\,pp over no-mid-training SFT+RL in Expt.\ 3 (four of five domains in-band, Operation just below the lower boundary), compared with $+0.8$\,pp under balanced coverage and $+0.6$\,pp under imbalanced coverage.

\subsection{Reasoning Evaluation and Domain-Level Analysis}

Reasoning ability is often reported as a single aggregate number, but aggregate scores can obscure where gains and losses occur. Two widely used benchmarks illustrate this masking problem: GSM8K~\citep{cobbe2021gsm8k} reduces grade-school arithmetic reasoning to a single pass@1 accuracy number, while BIG-Bench Hard~\citep{suzgun2022bbhard} aggregates 23 diverse tasks—spanning inductive, deductive, spatial, and algorithmic reasoning—into a single score that can conceal divergent per-domain trends of the kind our domain-level analysis reveals. \citet{ke2025survey} survey reasoning benchmarks broadly and emphasize the diversity of tasks used to evaluate LLM reasoning. Prior work on code and reasoning also decomposes broad competence into smaller behavioral units, for example in analyses of code-model functions~\citep{chen2024unlock} and parameter-efficient reasoning modules~\citep{huang2025lorapar}. \citet{huang2026remit} further use RL signals to guide mid-training data selection. We share the motivation of looking beyond a single global score, but our analysis is organized around domain-level mid-training coverage and downstream domain-level behavior.

\subsection{Knowledge-Orthogonal and Structured Reasoning Benchmarks}

Evaluating LLMs on structured reasoning tasks that require applying explicit rules rather than retrieving memorized world knowledge has gained increasing attention. \citet{han2022folio} introduce FOLIO, a first-order logic reasoning dataset in which models must derive entailment from formal logical statements without relying on surface-level heuristics; its difficulty for frontier models motivates the mid-training coverage study we conduct on the Logic domain. \citet{liu2020logiqa} present LogiQA, a multiple-choice logical reasoning benchmark drawn from Chinese civil-service examinations covering categorical, conditional, and disjunctive reasoning types; its structural diversity is representative of the rule-following challenge we study. \citet{sinha2019clutrr} construct CLUTRR to test systematic generalization in relational rule-following by withholding chain-length compositions from the training distribution—a compositional generalization challenge closely related to the formal-deduction structure of our Logic and Puzzle domains. More recently, \citet{zhou2024rulearena} propose RuleArena, evaluating models on complex real-world rule systems (airline policies, tax codes, NBA regulations), requiring the same kind of rule-adherence that KOR-Bench~\citep{ma2024korbench} formalizes as knowledge-orthogonal reasoning. COGS~\citep{kim2020cogs} probes compositional generalization in semantic parsing by constructing test cases whose structural complexity systematically exceeds the training distribution; the compositional gap it reveals—where models fail on combinations of primitives they handle individually—parallels KOR-Bench's rule-composition design and motivates the emphasis on generalization over memorization. The AI2 Reasoning Challenge (ARC)~\citep{clark2018arc} partitions elementary science questions into a retrieval-solvable Easy set and an inference-requiring Challenge set, providing an early demonstration that even modest reasoning demands stratify models in ways aggregate accuracy cannot capture.

Our work differs from these benchmark contributions in its focus on the \emph{training side}: rather than proposing new evaluation criteria, we ask how the proportion of domain-specific data during mid-training determines whether the corresponding structured reasoning capabilities are reliably acquired. The knowledge-orthogonal property of KOR-Bench—which ensures that training gains in one domain do not automatically transfer to others—makes it a particularly controlled instrument for isolating the effect of per-domain coverage.


\FloatBarrier
\section{Training and Evaluation Details}
\label{app:training}

\subsection{Compute}
\label{app:compute}

All training used full fine-tuning on Qwen3-8B-Base with DeepSpeed ZeRO-3 and bf16; evaluation used vLLM on 8 GPUs. We report the study's scale as run counts and token budgets, which are exactly determined by the design, rather than as GPU-hours, which we did not instrument per run.

Mid-training accounts for 160 runs at a fixed $\approx$1.5B-token budget over two epochs each: 24 sweep configurations, six held-out allocations, the $\theta^*$ allocation and the FineWeb-Edu baseline, at five seeds apiece. The supervised stage adds 155 compensatory passes and 155 matched uniform-control passes (both over the 24 sweep and six held-out checkpoints plus $\theta^*$, five seeds each), 20 standard-SFT passes for the three full-pipeline configurations and the shared SFT-only baseline, and a further 90 passes for the three additional sharpening branches of Appendix~\ref{app:sharpen} ($\gamma\in\{2,4,\infty\}$) over six configurations. The RL stage adds 50 GSPO runs at a fixed 200-step budget: the three full-pipeline configurations, the six held-out allocations, and the shared SFT+RL baseline, five seeds each. This totals 630 training runs (160 mid-training, 420 supervised, 50 RL) for the primary 8B study, of which mid-training dominates the cost because it is the only stage operating on a multi-billion-token corpus. Two validation sub-experiments are budgeted separately and on a smaller scale: Experiment A (Appendix~\ref{app:expA}) adds 12 interior allocations $\times$ 3 seeds $=$ 36 mid-training-only runs, and the 4B replication (Appendix~\ref{app:expB}) adds 8 allocations $\times$ 3 seeds $=$ 24 mid-training-only runs at the smaller model.

The design deliberately spends most of that budget on breadth at the mid-training stage rather than depth at the RL stage, which is why the RL leg is short (200 steps) and is reported as a secondary observation. A study prioritising the RL question differently would invert this allocation. For a group planning a replication, the cheapest informative subset is the coverage sweep plus the compensatory and uniform SFT passes---roughly the first two thirds of the runs above---since Observations~1 and~3 are both identified without any RL.

\subsection{Model Scale Selection}
\label{app:model_scale}

We chose Qwen3-8B-Base after measuring, rather than assuming, where a mid-training intervention has room to show itself. Table~\ref{tab:app-instruct} evaluates the Qwen3 family under our KOR-Bench protocol, and the relevant quantity is the headroom a base checkpoint still has before post-training: the base-to-instruct gap is $+21.2$\,pp at 4B ($41.88\rightarrow63.08$) and $+20.8$\,pp at 8B ($43.00\rightarrow63.80$), but only $+10.4$\,pp at 30B-A3B ($56.48\rightarrow66.88$). At the top of the range the base models have already closed most of that distance on their own---14B-Base reaches $60.56$ and 32B-Base $60.68$, against an instruct band of roughly $63$--$67$---so any mid-training reallocation competes for a shrinking remainder and its per-domain effects are compressed toward the ceiling. At the bottom, Qwen3-1.5B-Base scores $22.84$ overall with Cipher at $1.0$ and Puzzle at $4.2$, i.e.\ at the floor on two of the five domains, where allocation differences cannot register above seed noise. The 4B and 8B checkpoints are the two with the largest measured post-training headroom, carrying roughly twice that of 30B-A3B; 8B is the larger of the two and the one we could afford to run at full sweep scale (24 configurations $\times$ five seeds, plus six held-out allocations and the downstream passes). This is a statement about where the effect is measurable, not a claim that 8B is representative: whether the interior optima and the non-closure result reproduce at other scales is untested here.

\paragraph{What a 4B replication would have to show.}
Because 4B carries essentially the same post-training headroom as 8B ($+21.2$ vs.\ $+20.8$\,pp), it is the sharpest available test of whether our findings are scale-specific or headroom-specific: a failure to reproduce at 4B could not be explained away as a ceiling or floor artifact. A minimal replication needs three components. \emph{(i) Coverage sweep.} Train 8--12 allocations at 4B that span the low ($<10\%$), moderate ($10$--$40\%$) and high ($>40\%$) band for every domain---a subset of the allocations in Table~\ref{tab:mid_train} suffices---at three seeds, evaluated mid-training-only. This supports both the model-free band comparison of Appendix~\ref{app:bands} and the quadratic interiority permutation test on which our moderation claim actually rests. \emph{(ii) Alignment arm.} Apply the compensatory and uniform SFT passes to those same 4B checkpoints and recompute the pairwise closure counts together with the permutation null of Appendix~\ref{app:pairwise}; this separates a property of the recipe from a property of the 8B model. \emph{(iii) Peak comparison.} On allocations common to both scales, compare fitted peak locations at 4B and 8B. The three possible outcomes are diagnostic: peaks reproducing within seed noise would indicate the optimum is a property of the data mixture rather than the model; peaks shifting systematically with scale would make the fitted intervals scale-dependent and require the allocation guidance to be re-derived per scale; and an absence of interior optima at 4B, given its matched headroom, would localise the effect to 8B and falsify the generality we do not currently claim. A full-pipeline (SFT$+$RL) 4B arm is not required for any of these three tests, and would be needed only to check the $\theta^*$ ordering at a second scale.

\paragraph{Experiment A: an interior-simplex filler sub-sweep.}
\label{app:expA}
Our compositional reanalysis (Appendix~\ref{app:coda}) finds a saddle, not an interior maximum, because the 30 trained allocations cluster near the simplex's edges and mid-ridge, so the curvature of the interior cannot be resolved. To determine whether a jointly optimal mixture exists---which would let the per-domain moderation result of Observation~3 be promoted to a joint claim, or would establish that it cannot be---we therefore trained 12 new allocations confined to the interior (every domain share in $[12,34]\%$, each at least $5$\,pp from every existing allocation), drawn by Dirichlet($\alpha=8$) sampling and screened for feasibility. Table~\ref{tab:app-expA} reports them (mid-training-only, three seeds). \textbf{The result is consistent with the coda analysis and does not overturn it:} adding these 12 interior points to the 30-allocation pool and refitting the compositional surface leaves every domain's stationary point reading as a saddle ($R^2=0.78$--$0.96$ over the 42-pool), though with 15 quadratic terms from 42 allocations the Hessian classification is suggestive rather than decisive (Appendix~\ref{app:coda})---the interior data neither establish a joint optimum nor exclude a shallow flat region. The per-domain moderation result is correspondingly confirmed and delimited: over the full 42-allocation pool the moderate band remains the highest for all five domains (low/moderate/high means, e.g.\ Cipher $43.8/51.3/42.2$, Operation $51.9/75.8/50.4$), and the 12 interior points also supply an additional batch of out-of-sample curve-validation points. Because no joint optimum is identified, the moderation finding stays a per-domain marginal statement and is not promoted to a single recommended mixture.

\begin{table}[!ht]
\centering
\footnotesize
\setlength{\tabcolsep}{4pt}

\caption{Experiment A: 12 interior allocations (all components in 12--34\%, $\geq5$\,pp from every existing allocation), each trained mid-training-only over three seeds. Each row block reports the \textbf{Portion} (internal mid-training token share, \%), the \textbf{Evaluation} (mid-training-only accuracy, mean $\pm$\,SD across the three seeds), and the \textbf{Remedial SFT} allocation from Eq.~\ref{eq:comp_sft}. Adding these to the 30-allocation pool and refitting the compositional surface leaves a saddle in every domain, so no joint interior optimum is established (\textsection~\ref{sec:simplex}).}
\label{tab:app-expA}
\begin{tabular*}{\textwidth}{@{\extracolsep{\fill}}lccccc@{}}
\toprule
Domains & Ciphers & Operations & Logic & Counterfactual & Puzzles \\
\midrule
Config 1: Portion    &   12.9\% &   24.4\% &   29.0\% &   14.1\% &   19.5\% \\
Evaluation           & $52.9\pm1.6$\% & $76.4\pm3.0$\% & $57.2\pm1.8$\% & $74.2\pm2.4$\% & $12.0\pm1.2$\% \\
Remedial SFT         &   23.5\% &   17.8\% &   15.5\% &   22.9\% &   20.2\% \\
\midrule
Config 2: Portion    &   14.5\% &   18.3\% &   12.9\% &   21.2\% &   33.0\% \\
Evaluation           & $51.4\pm1.3$\% & $76.9\pm2.3$\% & $59.1\pm1.5$\% & $82.4\pm1.8$\% & $15.1\pm3.2$\% \\
Remedial SFT         &   22.7\% &   20.8\% &   23.5\% &   19.4\% &   13.5\% \\
\midrule
Config 3: Portion    &   12.5\% &   15.9\% &   27.5\% &   21.6\% &   22.5\% \\
Evaluation           & $51.5\pm2.1$\% & $77.2\pm2.9$\% & $57.6\pm0.6$\% & $82.6\pm3.0$\% & $13.6\pm1.3$\% \\
Remedial SFT         &   23.8\% &   22.0\% &   16.3\% &   19.2\% &   18.7\% \\
\midrule
Config 4: Portion    &   17.5\% &   22.0\% &   23.2\% &   12.0\% &   25.3\% \\
Evaluation           & $53.6\pm1.2$\% & $78.1\pm2.5$\% & $59.4\pm0.8$\% & $74.0\pm2.3$\% & $13.7\pm3.3$\% \\
Remedial SFT         &   21.3\% &   19.0\% &   18.4\% &   24.0\% &   17.3\% \\
\midrule
Config 5: Portion    &   30.4\% &   14.6\% &   12.9\% &   17.4\% &   24.6\% \\
Evaluation           & $50.7\pm0.7$\% & $77.4\pm0.7$\% & $58.6\pm2.8$\% & $78.0\pm3.4$\% & $14.4\pm1.8$\% \\
Remedial SFT         &   14.8\% &   22.7\% &   23.5\% &   21.3\% &   17.7\% \\
\midrule
Config 6: Portion    &   25.7\% &   24.5\% &   15.6\% &   16.6\% &   17.6\% \\
Evaluation           & $50.9\pm2.5$\% & $77.8\pm2.5$\% & $58.5\pm2.1$\% & $77.3\pm2.8$\% & $11.1\pm2.9$\% \\
Remedial SFT         &   17.2\% &   17.8\% &   22.2\% &   21.7\% &   21.2\% \\
\midrule
Config 7: Portion    &   24.9\% &   13.8\% &   18.6\% &   25.8\% &   16.9\% \\
Evaluation           & $50.4\pm1.1$\% & $77.9\pm3.1$\% & $59.5\pm1.1$\% & $81.9\pm2.8$\% & $11.5\pm2.9$\% \\
Remedial SFT         &   17.5\% &   23.1\% &   20.7\% &   17.1\% &   21.5\% \\
\midrule
Config 8: Portion    &   18.9\% &   24.7\% &   18.4\% &   15.9\% &   22.1\% \\
Evaluation           & $52.5\pm2.6$\% & $76.2\pm0.6$\% & $59.7\pm3.2$\% & $76.9\pm1.0$\% & $12.9\pm1.3$\% \\
Remedial SFT         &   20.5\% &   17.7\% &   20.8\% &   22.1\% &   18.9\% \\
\midrule
Config 9: Portion    &   24.3\% &   16.6\% &   16.4\% &   20.8\% &   21.9\% \\
Evaluation           & $51.5\pm2.2$\% & $77.2\pm2.6$\% & $60.4\pm2.5$\% & $80.9\pm1.2$\% & $12.8\pm1.5$\% \\
Remedial SFT         &   17.9\% &   21.7\% &   21.8\% &   19.6\% &   19.1\% \\
\midrule
Config 10: Portion   &   33.4\% &   17.4\% &   17.4\% &   12.7\% &   19.1\% \\
Evaluation           & $50.8\pm1.0$\% & $78.1\pm0.6$\% & $58.6\pm1.8$\% & $73.3\pm2.4$\% & $11.1\pm3.1$\% \\
Remedial SFT         &   13.3\% &   21.3\% &   21.3\% &   23.7\% &   20.5\% \\
\midrule
Config 11: Portion   &   13.2\% &   17.7\% &   23.8\% &   25.3\% &   20.1\% \\
Evaluation           & $51.8\pm3.4$\% & $78.9\pm1.0$\% & $58.8\pm2.8$\% & $83.5\pm3.4$\% & $13.0\pm3.2$\% \\
Remedial SFT         &   23.4\% &   21.2\% &   18.1\% &   17.4\% &   20.0\% \\
\midrule
Config 12: Portion   &   14.8\% &   19.2\% &   29.9\% &   15.4\% &   20.7\% \\
Evaluation           & $52.2\pm2.4$\% & $77.4\pm1.4$\% & $56.9\pm2.5$\% & $75.7\pm2.9$\% & $13.4\pm3.2$\% \\
Remedial SFT         &   22.6\% &   20.4\% &   15.1\% &   22.3\% &   19.6\% \\
\bottomrule
\end{tabular*}%
\end{table}

\section{Study II: A Cross-Scale Replication at 4B}
\label{app:expB}

The scale rationale of Appendix~\ref{app:model_scale} shows that the post-training headroom---the quantity that determines whether a mid-training reallocation can produce measurable per-domain differences---is largest at 4B ($+21.2$\,pp) and 8B ($+20.8$\,pp) and roughly half that at 30B-A3B ($+10.4$\,pp), while 1.5B sits on the floor (Cipher $1.0$, Puzzle $4.2$) and 14B/32B near the ceiling ($60.6$ vs.\ an instruct band of roughly $63$--$67$). Only the 4B and 8B checkpoints therefore lie in a regime where the effect can be observed. Because 4B carries almost the same headroom as 8B, a failure to reproduce at 4B cannot be written off as a ceiling or floor artifact, which makes it the sharpest available test of whether our findings are scale-specific or headroom-specific.

We therefore ran a mid-training-only replication at Qwen3-4B-Base over the eight allocations of Table~\ref{tab:app-expB}, chosen so that \emph{every domain} has low ($<10\%$), moderate ($10$--$40\%$) and high ($>40\%$) coverage samples across the set; the allocations reuse the corresponding 8B design points where they exist so the two scales can be compared on matched points. \textbf{The moderation pattern reproduces at 4B:} the moderate band again yields the highest mean accuracy in all five domains (e.g.\ Cipher $36.3/47.9/38.0$, Operation $49.3/90.8/61.1$), though the effect is far weaker for Counterfactual, whose three bands ($91.0/91.3/90.0$) span just $1.3$\,pp because its base prior is already near ceiling; the three-parameter quadratic is concave with an interior vertex in all five domains. The peak locations, however, shift with scale, and not in one direction: the fitted peaks at 4B (Cipher $8.8$, Operation $17.2$, Logic $18.8$, Counterfactual $8.5$, Puzzle $31.6$; the Counterfactual estimate is dominated by the near-flat response and its base prior of $90.5$, and Puzzle by a single high-coverage point) differ from the 8B peaks by $-1.1$, $+1.3$, and $+3.8$\,pp for Cipher, Operation, and Logic, $-3.5$\,pp for Puzzle, and $-17.5$\,pp for Counterfactual (deltas against the rounded peaks of Table~\ref{tab:coverage_bands}). This is the scale-dependence outcome flagged in Appendix~\ref{app:model_scale}: the interior optima are present at both scales, but their locations are not scale-invariant, so allocation guidance derived at one scale should not be transferred directly to another. We did not run the SFT or RL leg at 4B, so the alignment pass (compensatory vs.\ uniform closure, and the permutation null of Appendix~\ref{app:pairwise}) and the $\theta^*$ ordering were not re-tested at the second scale.

The band comparison and the quadratic interiority test (read off the mid-training-only evaluations) and the peak-location comparison are therefore the two of the three tests in Appendix~\ref{app:model_scale} that the current 4B data support; the alignment-pass test and the $\theta^*$ ordering remain to be run at the second scale.

\begin{table}[!ht]
\centering
\footnotesize
\setlength{\tabcolsep}{4pt}
\caption{Study II: eight cross-scale allocations for the Qwen3-4B-Base replication, each trained mid-training-only. Each row block reports the \textbf{Portion} (internal mid-training token share, \%), the \textbf{Bands} (each domain's low/moderate/high coverage), the \textbf{Evaluation} (mid-training-only accuracy, mean $\pm$\,SD), and the \textbf{Remedial SFT} allocation from Eq.~\ref{eq:comp_sft}. The 4B run was mid-training-only, so no post-SFT columns are reported; the moderation pattern reproduces at 4B (moderate band best in all five domains, concave quadratic with an interior vertex), but the fitted peak locations shift with scale.}
\label{tab:app-expB}
\begin{tabular*}{\textwidth}{@{\extracolsep{\fill}}lccccc@{}}
\toprule
Domains & Ciphers & Operations & Logic & Counterfactual & Puzzles \\
\midrule
Config 1: Portion    &   72.3\% &      0\% &     12\% &      0\% &   15.7\% \\
Bands (L/M/H)        &      H &      L &      M &      L &      M \\
Evaluation           & $38.0\pm0.8$\% & $44.4\pm1.8$\% & $45.2\pm3.0$\% & $90.5\pm1.0$\% & $7.4\pm2.5$\% \\
Remedial SFT         &      0\% &   28.3\% &   22.6\% &   28.3\% &   20.9\% \\
\midrule
Config 2: Portion    &     20\% &     20\% &     20\% &     20\% &     20\% \\
Bands (L/M/H)        &      M &      M &      M &      M &      M \\
Evaluation           & $45.1\pm0.8$\% & $96.9\pm1.3$\% & $49.9\pm1.5$\% & $91.0\pm0.8$\% & $9.1\pm0.7$\% \\
Remedial SFT         &     20\% &     20\% &     20\% &     20\% &     20\% \\
\midrule
Config 3: Portion    &      0\% &     10\% &     40\% &     10\% &     40\% \\
Bands (L/M/H)        &      L &      M &      M &      M &      M \\
Evaluation           & $25.2\pm1.8$\% & $75.3\pm3.1$\% & $42.2\pm0.7$\% & $91.9\pm3.0$\% & $10.7\pm1.9$\% \\
Remedial SFT         &     30\% &     25\% &     10\% &     25\% &     10\% \\
\midrule
Config 4: Portion    &      5\% &     15\% &      5\% &     15\% &     60\% \\
Bands (L/M/H)        &      L &      M &      L &      M &      H \\
Evaluation           & $43.8\pm0.6$\% & $96.3\pm1.7$\% & $37.1\pm3.1$\% & $91.7\pm1.4$\% & $2.3\pm2.0$\% \\
Remedial SFT         &   27.5\% &   22.5\% &   27.5\% &   22.5\% &      0\% \\
\midrule
Config 5: Portion    &     33\% &      5\% &      6\% &     50\% &      6\% \\
Bands (L/M/H)        &      M &      L &      L &      H &      L \\
Evaluation           & $48.7\pm1.1$\% & $55.9\pm1.6$\% & $35.9\pm0.7$\% & $90.0\pm2.4$\% & $2.3\pm3.0$\% \\
Remedial SFT         &   13.5\% &   27.5\% &     27\% &      5\% &     27\% \\
\midrule
Config 6: Portion    &      4\% &      3\% &     30\% &     29\% &     34\% \\
Bands (L/M/H)        &      L &      L &      M &      M &      M \\
Evaluation           & $39.8\pm0.8$\% & $47.6\pm0.9$\% & $49.4\pm2.2$\% & $90.6\pm0.8$\% & $12.6\pm3.0$\% \\
Remedial SFT         &     28\% &   28.5\% &     15\% &   15.5\% &     13\% \\
\midrule
Config 7: Portion    &     25\% &     45\% &     10\% &     10\% &     10\% \\
Bands (L/M/H)        &      M &      H &      M &      M &      M \\
Evaluation           & $48.3\pm2.1$\% & $61.1\pm2.5$\% & $45.7\pm1.9$\% & $91.5\pm1.7$\% & $4.6\pm3.1$\% \\
Remedial SFT         &   17.5\% &    7.5\% &     25\% &     25\% &     25\% \\
\midrule
Config 8: Portion    &     20\% &     25\% &     45\% &      5\% &      5\% \\
Bands (L/M/H)        &      M &      M &      H &      L &      L \\
Evaluation           & $49.3\pm0.9$\% & $94.9\pm0.9$\% & $40.5\pm0.7$\% & $91.4\pm1.8$\% & $3.8\pm1.4$\% \\
Remedial SFT         &     20\% &   17.5\% &    7.5\% &   27.5\% &   27.5\% \\
\bottomrule
\end{tabular*}%
\end{table}

\subsection{GSPO Objective}
\label{app:gspo}

Group-based Sequence-level Policy Optimization (GSPO)~\citep{yang2025qwen3} is a group-relative policy gradient method. For each prompt $q$, $G$ responses $\{o_1, \ldots, o_G\}$ are sampled from the current policy $\pi_\theta$. Each response receives a binary verifier reward $R_i \in \{0, 1\}$ (1 if correct, 0 otherwise). The advantage for response $o_i$ is computed via group-level normalisation:
\begin{equation}
A_i = \frac{R_i - \mathrm{mean}(\{R_1, \ldots, R_G\})}{\mathrm{std}(\{R_1, \ldots, R_G\})},
\label{eq:gspo-advantage}
\end{equation}
and the policy is updated by maximising the clipped surrogate objective:
\begin{equation}
\resizebox{\textwidth}{!}{$
\mathcal{J}_{\mathrm{GSPO}}(\theta) = \mathbb{E}_{q,\{o_i\}}\!\left[\frac{1}{G}\sum_{i=1}^G \min\!\left(\rho_i A_i,\; \mathrm{clip}(\rho_i, 1-\varepsilon, 1+\varepsilon)A_i\right)\right],
$}
\label{eq:gspo-obj}
\end{equation}
where $\rho_i = \pi_\theta(o_i \mid q) / \pi_{\mathrm{old}}(o_i \mid q)$ is the \emph{sequence-level} probability ratio. Under the standard autoregressive factorization this is the product of per-token ratios, $\rho_i = \prod_t \pi_\theta(o_{i,t}\mid q, o_{i,<t})/\pi_{\mathrm{old}}(o_{i,t}\mid q, o_{i,<t})$, equivalently the sum of per-token log-ratios in log space. $\varepsilon = 0.2$ is the clip ratio, and $\theta_{\mathrm{old}}$ denotes the policy parameters before the update. When all rewards within a group are equal, $\mathrm{std}(\{R_1,\ldots,R_G\})=0$ and the normalized advantage in Eq.~\ref{eq:gspo-advantage} is undefined. Because a group with all-equal binary rewards carries no preference signal, its treatment affects only that group's gradient contribution and not the relative ordering of updates across other groups; the exact handling in our runs is implementation-specific and is acknowledged here as an implementation detail rather than a design choice. Following \citet{yang2025qwen3}, we use group size $G = 8$. Training proceeds for 200 steps with a peak learning rate of $1 \times 10^{-6}$ under the cosine schedule with 5\% warmup (Table~\ref{tab:app-hparams}). RL rollouts are sampled with the same 16,384-token sequence cutoff as evaluation, so rollout truncation does not differentially bias verifier rewards for long reasoning traces (Appendix~\ref{app:rl_budget}).

\subsection{Training Hyperparameters}

Table~\ref{tab:app-hparams} lists the complete hyperparameter set for all three stages. The compensatory-SFT pass and its uniform-SFT control reuse the SFT column's three-epoch protocol at a matched total token budget (Appendix~\ref{app:comp_sft}); the RL stage uses a fixed 200-step GSPO schedule (Appendix~\ref{app:rl_budget}).

\begin{table}[!ht]
\centering
\footnotesize
\setlength{\tabcolsep}{4pt}
\begin{tabular*}{\textwidth}{@{\extracolsep{\fill}}llll@{}}
\toprule
Parameter & Mid-Training & SFT & RL (GSPO) \\
\midrule
Base model & Qwen3-8B-Base (base) & Mid-training ckpt. & Mid-training+SFT ckpt. \\
Stage type & pt (cont. pretraining) & sft & rl (GSPO) \\
Fine-tuning type & Full & Full & Full \\
Learning rate & 1e-5 & 5e-5 & 1e-6 \\
Per-device batch size & 4 & 4 & 1 \\
Gradient accumulation steps & 4 & 4 & 16 \\
Effective batch size & 16 & 16 & 16 \\
Sequence length (cutoff) & 12,000 & 12,000 & 16,384 \\
Training volume & $\sim$1.5B tokens (2 ep.) & 36,537 inst.\ (3 ep.) & 200 steps \\
Epochs & 2.0 & 3.0 & — (step-based) \\
Optimizer & AdamW (ZeRO-3) & AdamW (ZeRO-3) & AdamW (ZeRO-3) \\
LR schedule & warmup\_stable\_decay & cosine & cosine \\
Warmup ratio & 0.05 & 0.05 & 0.05 \\
Precision & bf16 & bf16 & bf16 \\
Gradient checkpointing & enabled & enabled & enabled \\
Flash attention & FA2 & FA2 & FA2 \\
Packing & enabled & enabled & disabled \\
Validation split & 1\% & 5\% & reward-based \\
Chat template & None (completion) & qwen (chat) & qwen (chat) \\
\midrule
\multicolumn{4}{l}{\textit{RL-specific hyperparameters}} \\
Group size $G$ & — & — & 8 \\
Clip ratio $\varepsilon$ & — & — & 0.2 \\
Reward type & — & — & binary verifier \\
\bottomrule
\end{tabular*}%
\caption{Training hyperparameters for all three stages. Mid-training and SFT use LLaMA-Factory with full fine-tuning and DeepSpeed ZeRO-3. RL uses GSPO~\citep{yang2025qwen3} with binary verifier rewards at a fixed 200-step budget. The compensatory-SFT pass and its uniform-SFT control use the same 3-epoch protocol and total token budget as the SFT column (Appendix~\ref{app:comp_sft}). FA2 = FlashAttention-2.}
\label{tab:app-hparams}
\end{table}

\subsection{Seed Selection and Model Selection Protocol}
\label{app:model_hparams}

For the mid-training coverage sweep (Table~\ref{tab:mid_train}), each configuration is trained from five random seeds and evaluated on a held-out KOR-Bench test split that was never used during training or model selection; we report the mean $\pm$ 1 SD across the five seeds, so the SDs capture cross-seed variance for the mid-training-only, compensatory-SFT, and uniform-SFT evaluations. The full-pipeline rows in Table~\ref{tab:main} (Mid-training+SFT, Mid-training+SFT+RL) are likewise five-seed means, and Table~\ref{tab:main} reports the corresponding cross-seed SDs in every row and cell; the external-benchmark measurements are five-seed means, and bootstrap 95\% CIs over per-question resamples, where reported, capture evaluation-sample variance and do not replace the cross-seed uncertainty. All values in this manuscript are five-seed means unless stated otherwise.

\subsection{Answer Extraction, Decoding, and Evaluation Protocol}
\label{app:eval_details}

We extract the final answer via regex \texttt{[[...]]} (last match). For Operation, we additionally apply SymPy-based equivalence matching. For Logic, whitespace and punctuation are stripped before comparison. All evaluations use greedy decoding (temperature = 0.0, max\_tokens = 16,384) with vLLM on 8 GPUs. Overall accuracy is the macro-average of domain-level accuracies $\mathrm{Acc}(d)=\frac{1}{N_d}\sum_{i=1}^{N_d}\mathbb{1}[\hat{y}_i=y_i^*]$. For difficulty-stratified external benchmarks, per-stratum accuracy is $\mathrm{Acc}(d,\ell)=\frac{1}{N_{d,\ell}}\sum_{i:\ell_i=\ell}\mathbb{1}[\hat{y}_i=y_i^*]$, where $\ell$ is proof depth, house count, or reasoning type. Confidence intervals are computed using bootstrap resampling for KOR-Bench domain results and asymptotic intervals for large external strata. External benchmarks are evaluated using their native protocols.

\subsection{External Reference Models (Qwen3 Family)}
\label{app:instruct}

Table~\ref{tab:app-instruct} evaluates the Qwen3 family~\citep{yang2025qwen3} under the same KOR-Bench protocol, for two purposes. None of these models is trained in our pipeline and their instruction-tuning and RL recipes differ from the recipe studied here, so no value in the table is comparable to our pipeline rows---in particular the Qwen3-8B-Instruct score of $63.80\%$ should not be read as evidence about any of our allocations. The table's role is instead (i)~to place our trained checkpoints against publicly available reference points, and (ii)~to supply the base-to-instruct headroom measurements that motivate the choice of scale in Appendix~\ref{app:model_scale}.

\begin{table}[!ht]
\centering
\footnotesize
\setlength{\tabcolsep}{4pt}
\caption{KOR-Bench zero-shot accuracy (\%) of the Qwen3 model family (1.5B/4B/8B/14B/30B-A3B/32B, base and instruct), evaluated with the same protocol as Table~\ref{tab:main}. None of these models is trained in our pipeline; their instruction-tuning and RL recipes differ from the recipe studied here, so no value is comparable to our pipeline rows. The base-to-instruct pairs supply the post-training headroom used in the model-scale rationale (Appendix~\ref{app:model_scale}).}
\label{tab:app-instruct}
\begin{tabular*}{\textwidth}{@{\extracolsep{\fill}}lcccccc@{}}
\toprule
Model & Overall & Cipher & Oper. & Logic & Counterf. & Puzzle \\
\midrule
Qwen3-1.5B-Base & 22.84 & 1.0 & 7.0 & 20.2 & 81.8 & 4.2 \\
\midrule
Qwen3-4B-Base & 41.88 & 18.2 & 60.0 & 39.2 & 90.0 & 2.0 \\
\midrule
Qwen3-4B-Instruct & 63.08 & 59.8 & 93.4 & 49.0 & 88.2 & 25.0 \\
\midrule
Qwen3-8B-Base & 43.00 & 13.2 & 60.8 & 49.2 & 86.0 & 5.8 \\
\midrule
Qwen3-8B-Instruct & 63.80 & 66.1 & 74.5 & 57.0 & 93.3 & 28.1 \\
\midrule
Qwen3-14B-Base & 60.56 & 50.4 & 88.8 & 56.0 & 93.2 & 14.4 \\
\midrule
Qwen3-30B-A3B-Base & 56.48 & 43.2 & 85.6 & 45.2 & 96.4 & 12.0 \\
\midrule
Qwen3-30B-A3B-Instruct & 66.88 & 65.6 & 90.0 & 58.4 & 91.6 & 28.8 \\
\midrule
Qwen3-32B-Base & 60.68 & 51.8 & 87.8 & 57.0 & 90.6 & 16.2 \\
\bottomrule
\end{tabular*}%
\end{table}

\subsection{RL Budget: Interpretive Limitations and Training Instability}
\label{app:rl_budget}

\begin{figure}[!htbp]
\centering
\includegraphics[width=\textwidth]{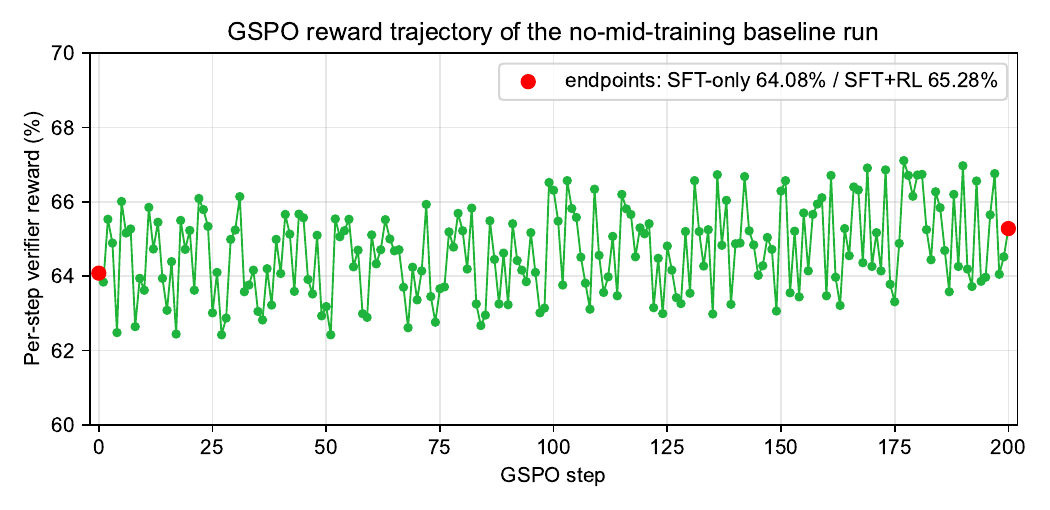}
\caption{Per-step group-mean verifier reward of the GSPO stage for the shared no-mid-training baseline run (SFT$\rightarrow$RL, 200-step schedule). Red endpoints mark the SFT-only evaluation accuracy at step 0 (64.08\%) and the final SFT+RL accuracy at step 200 (65.28\%). The reward wanders in a $\approx$62--67\% band without a consistent upward trend, matching the small evaluation gain of the RL stage ($+1.20$\,pp).}
\label{fig:reward_traj}
\end{figure}

The small RL gains observed across all three experiments ($+1.1$--$+2.3$\,pp from GSPO; the largest, $+2.30$\,pp, occurs under the exploratory $\theta^*$ allocation) raise the question of whether a larger RL budget would alter the compensatory conclusion.
Two interpretations are consistent with the data.
\textbf{Interpretation 1 --- Competence-limited RL.}
The 200-step GSPO stage may be genuinely limited by the model's current competence: verifier-based RL reinforces existing chains rather than discovering new symbolic procedures, so domains that mid-training left weak receive little RL gain regardless of budget.
\textbf{Interpretation 2 --- Insufficient RL budget.}
200 steps may simply be insufficient to manifest RL's corrective potential; a longer schedule could in principle reshape the domain gaps by giving the model more opportunities to explore and for the verifier to surface correct but low-probability chains in weak domains.
We cannot distinguish these interpretations from our experiments, and we treat the RL leg as a secondary observation rather than a headline claim for two reasons. First, the reward is a binary correctness verifier, so no dense per-token reward signal can be extracted from the chain-of-thought; RL can only reinforce whole-answer outcomes. Second, extending the GSPO budget is not a cleanly tunable knob in practice: longer schedules were unstable across seeds (reward divergence without a consistent directional trend) rather than simply maturing into a stable higher-performing policy. Two distinct RL settings appear in this paper and we separate them explicitly. (i)~The three full-pipeline configurations of Table~\ref{tab:main} use the standard pipeline (mid-training $\rightarrow$ rule-type-balanced SFT $\rightarrow$ RL). (ii)~The six held-out allocations are carried through mid-training $\rightarrow$ \emph{compensatory} SFT $\rightarrow$ RL, and it is these runs that produce the \textbf{RL Evaluation} rows of Table~\ref{tab:mid_train} and the out-of-sample results reported in \S\ref{sec:coverage_impact} (mean $+1.19$\,pp over the compensatory pass; $0/60$ pairs closed at the $5$\,pp threshold and $7/60$ at the $10\%$ ratio after the complete pipeline). Early exploratory attempts to add RL after compensatory SFT \emph{on the 24 sweep configurations} showed cross-seed instability and are excluded from the reported sweep results, under the same criterion used elsewhere (reward divergence or no consistent directional trend within the 200-step budget); that exclusion applies to those sweep runs only and not to the six held-out allocations, whose RL runs are reported in full. Whether a compensatory-SFT-plus-RL schedule would close gaps across the whole 24-configuration sweep therefore remains open, while on the six held-out allocations it demonstrably does not. We scope the RL stage to the tested 200-step schedule and do not read its small gains as evidence about post-training repair in general. Figure~\ref{fig:reward_traj} illustrates the pattern on the shared no-mid-training baseline run: the per-step group-mean verifier reward fluctuates between $\approx62\%$ and $67\%$ across the 200 steps without a consistent upward trend, and the evaluation gain over the whole stage is small ($+1.20$\,pp, SFT-only 64.08\% to SFT+RL 65.28\%).

\subsection{Code and Data Availability}
\label{app:code}

To support replication, we will release the training configurations, the data-generation pipeline (rule library, synthesizer, deterministic solvers, verification, and preprocessing), the evaluation harness (prompt templates, answer-extraction rules, and difficulty stratification), the analysis scripts that produce every reported table and figure, and per-seed results for all reported tables, together with the RL reward trajectories of all runs, including those excluded under the criteria of Appendix~\ref{app:rl_budget}.

\subsection{Pipeline and Study-Roadmap Figures}
\label{app:figures}

The synthetic data-generation pipeline is reproduced here.

\begin{figure}[!htbp]
\centering
\includegraphics[width=\textwidth]{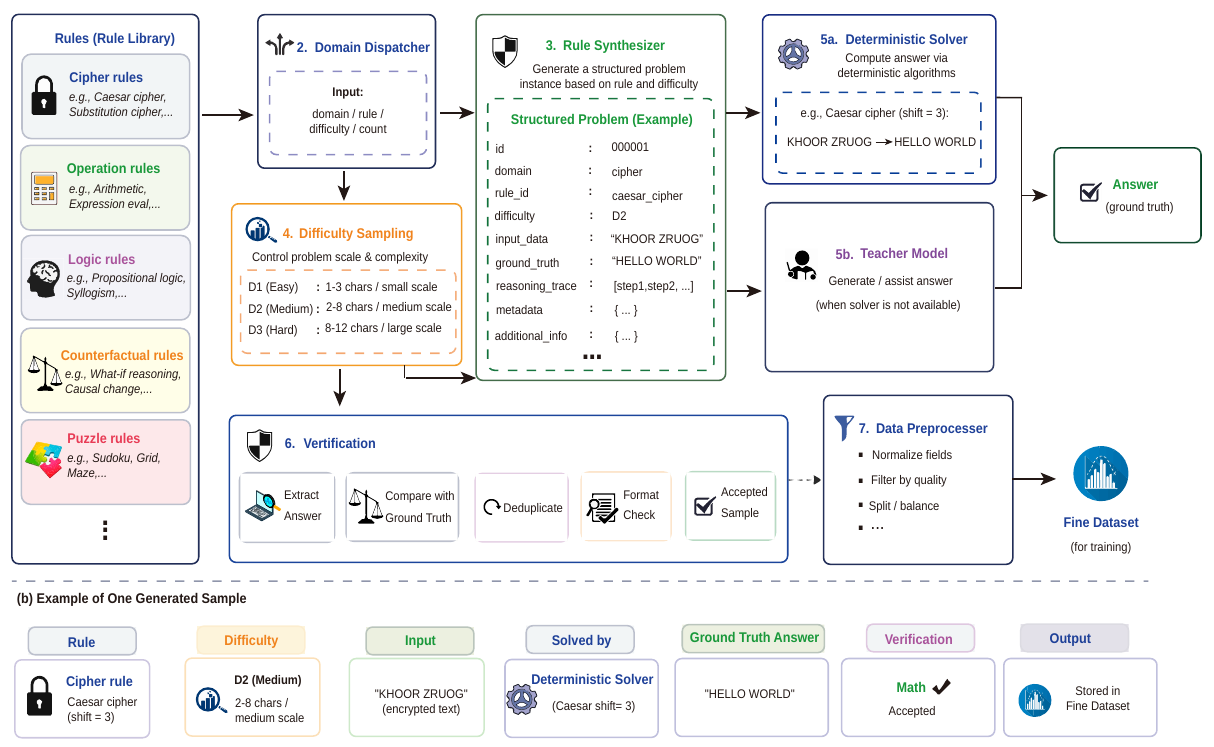}
\caption{Synthetic data generation pipeline: symbolic solver produces (problem, answer) pairs from KOR-Bench rule definitions, passing through verification, split isolation, and deduplication.}
\label{fig:data_pipeline}
\end{figure}

\begin{figure}[!htbp]
\centering
\includegraphics[width=\textwidth]{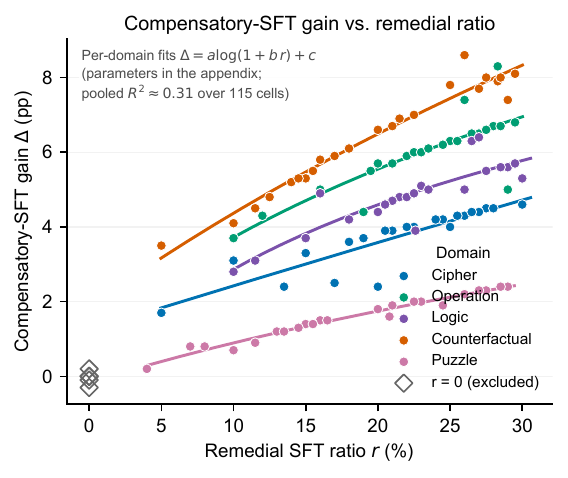}
\caption{Compensatory-SFT gain $\Delta=$ Eval.\,(Remedial)$-$Evaluation vs.\ the domain's remedial SFT ratio $r$, from Table~\ref{tab:mid_train} (115 cells with $r>0$, coloured by domain; the five $r=0$ cells are open diamonds and excluded from the fits). Solid curves: per-domain fits $\Delta \approx a\log(1+b\,r)+c$ (Table~\ref{tab:comp_logfit}). The gain rises with $r$ and saturates at a domain-specific ceiling; pooling across domains leaves most cell-level variance unexplained (pooled $R^2\approx0.31$). Descriptive, not confirmatory.}
\label{fig:sft_amplification}
\end{figure}

\begin{table}[!htbp]
\centering
\footnotesize
\setlength{\tabcolsep}{4pt}
\caption{Per-domain accuracy change ($\Delta$, pp) vs.\ the same-stage no-mid-training baseline at each pipeline stage, for Expt.\ 1 (imbalanced). Mid-only vs.\ Base; Mid+SFT vs.\ SFT only; Mid+SFT+RL vs.\ SFT+RL. All entries in pp. The mid-training-only collapse (Operation $-27.2$, Counterfactual $-38.0$) is reversed in sign by SFT/RL---strongly for Operation, and within seed noise for Counterfactual; a smaller residual deficit persists on Logic and Puzzle. Rows are five-seed means.}
\label{tab:survival}
\begin{tabular*}{\textwidth}{@{\extracolsep{\fill}}lccc@{}}
\toprule
Domain & Mid-only vs.\ Base & Mid+SFT vs.\ SFT only & Mid+SFT+RL vs.\ SFT+RL \\
\midrule
Cipher         & $+31.3\pm1.8$ & $+4.0\pm1.0$ & $+2.4\pm4.4$ \\
Operation      & $-27.2\pm1.1$ & $+5.2\pm0.3$ & $+4.0\pm2.8$ \\
Logic          & $+7.3 \pm0.9$ & $-6.0\pm0.8$ & $-2.4\pm2.5$ \\
Counterfactual & $-38.0\pm2.4$ & $-0.8\pm1.2$ & $+1.2\pm3.4$ \\
Puzzle         & $+8.0 \pm0.5$ & $+1.2\pm0.3$ & $-2.0\pm3.0$ \\
\midrule
Overall        & $-3.72$ & $+0.72$ & $+0.64$ \\
\bottomrule
\end{tabular*}%
\end{table}
\FloatBarrier
\section{Data Accounting}
\label{app:ledger}

Coverage as measured in this study is the \emph{token share} of each domain within the internal five-domain component (Table~\ref{tab:mid_train}, Portion rows). Because the downstream recipe and the total token budget are held fixed, the coverage percentages fully determine the per-domain token allocation of that component; Table~\ref{tab:ledger} records the corresponding allocation for the two configurations used in the full-pipeline comparison (Expt.\ 1 and the balanced allocation). The composition and provenance of the fixed external corpus component are described in \S\ref{sec:synth_method}; the internal/external token split and the per-component budgets are fixed across all configurations.

\begin{table}[!ht]
\centering
\footnotesize
\setlength{\tabcolsep}{4pt}
\caption{Per-configuration data accounting for the mid-training coverage sweep. Coverage percentages are the measured internal-component token shares (Table~\ref{tab:mid_train}); the internal/external token ratio and the per-component token budgets are fixed across all 24 configurations.}
\label{tab:ledger}
\begin{tabular*}{\textwidth}{@{\extracolsep{\fill}}lccccc@{}}
\toprule
Quantity & Cipher & Operation & Logic & Counterfactual & Puzzles \\
\midrule
Internal token share (e.g.\ balanced config) & 20\% & 20\% & 20\% & 20\% & 20\% \\
Internal token share (e.g.\ Expt.\ 1) & 72.3\% & 0\% & 12.0\% & 0\% & 15.7\% \\
\bottomrule
\end{tabular*}%
\end{table}

\FloatBarrier
\section{Decontamination and Split Isolation}
\label{app:decontam}

\begin{enumerate}
\item \textbf{ProofWriter-family depth split.} The external mid-training corpus contains only regenerated ProofWriter-family items of depth at most 5; evaluation uses depths 5 to 9. Depth 5 is explicitly labeled as in distribution.

\item \textbf{KOR-Bench instance split.} SFT and evaluation data use different random seeds. No identical instance appears in both; evaluation instances are held out by instance ID.

\item \textbf{External benchmark isolation.} No CounterBench-family items and no ProofWriter-family items of depths 6--9 appear in any training corpus. ZebraLogic-family items contribute 201 mid-training samples (0.3\% coverage); these are distinct instances from the evaluation set but same-family exposure, so ZebraLogic is not a zero-exposure benchmark (Appendix~\ref{app:external}).

\item \textbf{Base model contamination.} We do not control for Qwen3-8B-Base's original pretraining exposure to public benchmarks. This limitation applies to all work with pretrained models.

\item \textbf{Teacher trace isolation.} SFT chain-of-thought traces are generated by the pipeline's symbolic solvers, not by any LLM. No teacher model accesses evaluation instances.
\end{enumerate}

\FloatBarrier
\section{Mid-Training Coverage Sweep}
\label{app:coverage_sweep}

Table~\ref{tab:mid_train} reports the full per-configuration results of the 24 coverage sweep variants (§\ref{sec:coverage_impact}) together with the FineWeb-Edu baseline and the exploratory $\theta^*$ allocation. Each row block gives: Portion (domain allocation), Evaluation (mid-training-only accuracy, five-seed mean $\pm$ 1 SD), Remedial SFT (compensatory allocation from Eq.~\ref{eq:comp_sft}), Eval.\ (Remedial) (accuracy after the compensatory SFT pass), and Eval.\ (Uniform) (accuracy after the matched uniform-SFT control; Appendix~\ref{app:comp_sft}). All 24 configurations include both SFT-pass results. The external corpus derives 34.8\% of its tokens from the ProofWriter rule family, contributing shared formal-deduction exposure that explains why Logic's curve sits above zero at 0\% internal coverage, while Puzzle collapses to near-zero.

\paragraph{Per-domain fitted-region interpretation.}
Cipher, Operation, and Logic have broad 95\%-of-peak intervals (upper bounds near 30--36\%), consistent with their large fitted $\sigma_R$ values; Cipher's interval [7.4\%, 36.0\%] spans the wide plateau observed in Figure~\ref{fig:midtrain_domain_lines} rather than a sharp peak. Counterfactual's fitted 95\%-of-peak interval [17.9\%, 47.7\%] is consistent with a broad fitted peak near 26.0\%. Puzzle's interval [27.2\%, 41.2\%] requires relatively high coverage; its fitted peak accuracy (15.0\%) is low, and whether a larger model could raise it remains open.

\paragraph{Simplex design validity.}
Each domain's x-coordinate in Figure~\ref{fig:midtrain_domain_lines} is its own internal proportion while the other four vary jointly. The curves describe exploratory mixture-level associations rather than isolated causal functions. A compositional or simplex-aware joint response-surface analysis (in the sense of compositional data analysis~\citep{aitchison1986compositional}) would be needed for stronger allocation claims; we report a preliminary such analysis in Appendix~\ref{app:coda}, which finds a saddle rather than an interior maximum even after the 12 interior allocations of Appendix~\ref{app:expA} are added to the pool. A factorial design that independently varies each domain's coverage remains future work. Any monotonicity or concavity tests below are likewise descriptive diagnostics, and should not be read as population-level causal tests.

\paragraph{Concavity test (descriptive).}
We fit a quadratic $y_d = \beta_0 + \beta_1\theta + \beta_2\theta^2$ per domain ($n=24$) as an exploratory shape summary. The negative $\hat\beta_2$ values in Table~\ref{tab:app-quadratic} indicate a concave trend within this sweep, but they are reported strictly as descriptive shape descriptors and are not tests of curvature: accuracy is a bounded proportion, observations share training configurations, and the five domain fits are not independent. A binomial/beta-binomial or clustered compositional analysis with multiplicity control would be required before making inferential claims about curvature; such an analysis is left for future work.

\begin{table}[!ht]
\centering
\footnotesize
\setlength{\tabcolsep}{4pt}
\caption{Per-domain quadratic regression concavity check, descriptive only. OLS fit $y = \beta_0 + \beta_1\theta + \beta_2\theta^2$ on all 24 normal coverage configurations per domain. $\hat\beta_2$ and its standard error are reported as shape descriptors only; the standard error is the ordinary-least-squares dispersion of the coefficient and is \emph{not} used as a curvature test. The calibrated test of interiority is the permutation test of Appendix~\ref{app:bands} (see $\S$\ref{app:coverage_sweep}). $\hat\beta_2$ and SE are rounded independently from unrounded estimates.}
\label{tab:app-quadratic}
\begin{tabular*}{\textwidth}{@{\extracolsep{\fill}}lrrr@{}}
\toprule
Domain & $\hat\beta_2$ & SE$(\hat\beta_2)$ & $R^2_\text{quad}$ \\
\midrule
Cipher         & $-0.013$ & $0.003$ & $0.494$ \\
Operation      & $-0.048$ & $0.004$ & $0.876$ \\
Logic          & $-0.033$ & $0.004$ & $0.816$ \\
Counterfactual & $-0.035$ & $0.002$ & $0.987$ \\
Puzzle         & $-0.012$ & $0.001$ & $0.956$ \\
\bottomrule
\end{tabular*}%
\end{table}

\subsection{Model-Free Check: Coverage Bands}
\label{app:bands}

The split-Gaussian summaries below are five-parameter fits, so a reader may reasonably ask how much of the inverted-U shape is a property of the model rather than of the data. We therefore repeat the moderation claim without any curve fitting. Pooling all 30 trained allocations (the 24 sweep configurations plus the six held-out allocations), we bin each domain's own coverage into \emph{low} ($<10\%$), \emph{moderate} ($10$--$40\%$), and \emph{high} ($>40\%$) and compare the mean mid-training-only accuracy in each band (Table~\ref{tab:app-bands}). For every one of the five domains the moderate band is the best of the three, and accuracy is lower both below and above it. If the best-of-three band were exchangeable across domains, the probability that all five select the moderate band would be $(1/3)^5\approx0.004$; because the five domains share the same 30 allocations they are not independent trials, so we report this as a descriptive concordance rather than a calibrated test. The high band is also thinly populated for several domains ($n=1$--$5$), so the low-side contrast is the better-supported half of the pattern; the 12 interior allocations of Appendix~\ref{app:expA}, all in the moderate band, do not add high-coverage points but thicken the moderate band (to $n=29$--$32$), which is the side the moderation pattern rests on. Within these limits, the interior optimum is visible in the raw group means and does not depend on the parametric fit, and it survives the addition of the 12 interior allocations (moderate best for 5/5 domains over the 42-allocation pool). One bookkeeping point deserves stating plainly: the six held-out allocations serve two roles. They are withheld from the split-Gaussian curve fitting, which is what makes the out-of-sample test of \S\ref{sec:coverage_impact} valid, but they are included in the 30-allocation pool used for the band comparison above, for the interiority permutation test, and for the compositional surface of \S\ref{app:coda}, because those analyses are not fitted to the curves and benefit from the extra design points. No statistic is therefore both fitted on and validated against the same allocations, but the band and interiority results are in-sample with respect to the pool (30 or, once Appendix~\ref{app:expA} is included, 42) and should be read as such. A permutation test sharpens this into a calibrated statement. For each domain we fit a three-parameter quadratic OLS to the 42-allocation pool (30 base plus the 12 interior allocations of Appendix~\ref{app:expA}) and ask whether it is concave ($\hat\beta_2<0$) with a vertex strictly inside the tested coverage range; all five domains satisfy both conditions. Under a null that permutes whole accuracy rows across allocations---destroying the coverage--accuracy link while leaving the simplex design and the cross-domain correlation structure exactly intact---an average of $1.78$ of five domains do so, and all five do so in $0.98\%$ of $20{,}000$ draws ($P\approx0.010$). Shuffling each domain independently gives the same answer ($P\approx0.005$). We use the quadratic here only to test concavity and interiority, not to locate the optima: a symmetric parabola pulls its vertex toward the centre of the coverage range, so its vertex positions are not comparable to the split-Gaussian peaks reported in Table~\ref{tab:coverage_bands}. Two further permutation results bear on how much the split-Gaussian fit itself can be trusted, and we report both, including the one that cuts against us. First, the fitted curves capture real structure: under the same whole-row permutation the mean per-domain $R^2$ of the split-Gaussian is $0.174\pm0.055$, against $0.982$ for the observed data ($P<0.001$, 250 draws). Second, and less favourably, the \emph{count} of interior peaks is not a usable statistic for a five-parameter split-Gaussian: a permuted dataset still yields interior peaks in all five domains $64\%$ of the time, because a flexible asymmetric curve can place a maximum inside the range even on noise. This is precisely why the interiority test above is run on the three-parameter quadratic, which cannot manufacture an interior optimum so easily, and why we do not treat ``all five peaks are interior'' as evidence on its own. Finally, the five fitted peaks sum to $\approx102\%$, i.e.\ they are close to jointly realisable within the 100\% budget. A referee might read a sum near the budget as the fingerprint of a simplex artifact; the permutation null does not support that reading, since permuted data give a peak sum of $121.5\pm33.3\%$ that does not concentrate near 100\%, and only $2.8\%$ of draws land as close to the budget as the observed sum. We nonetheless treat the near-feasibility as a numerical coincidence worth reporting rather than as a design rule, because the sum of five marginal optima has no guaranteed relationship to the joint optimum under the constraint (\S\ref{sec:simplex}).

\begin{table}[!ht]
\centering
\footnotesize
\setlength{\tabcolsep}{4pt}
\caption{Mean mid-training-only accuracy (\%) by coverage band, pooling all 30 trained allocations (24 sweep $+$ 6 held-out); $n$ is the number of allocations falling in each band for that domain. The moderate band is the best of the three for all five domains. No curve fitting is involved.}
\label{tab:app-bands}
\begin{tabular*}{\textwidth}{@{\extracolsep{\fill}}lccc@{}}
\toprule
Domain & Low ($<10\%$) & Moderate ($10$--$40\%$) & High ($>40\%$) \\
\midrule
Cipher         & 43.8 ($n{=}9$)  & \textbf{51.0} ($n{=}18$) & 42.2 ($n{=}3$) \\
Operation      & 51.9 ($n{=}11$) & \textbf{74.7} ($n{=}17$) & 50.4 ($n{=}2$) \\
Logic          & 39.6 ($n{=}9$)  & \textbf{58.1} ($n{=}20$) & 40.0 ($n{=}1$) \\
Counterfactual & 57.5 ($n{=}9$)  & \textbf{81.2} ($n{=}20$) & 78.2 ($n{=}1$) \\
Puzzle         & \phantom{0}6.1 ($n{=}5$)  & \textbf{13.1} ($n{=}20$) & 10.5 ($n{=}5$) \\
\bottomrule
\end{tabular*}%
\end{table}

\subsection{Compositional (Aitchison) Reanalysis}
\label{app:coda}

Because coverage is a closed composition, an analysis that treats the five shares as free variables is not strictly appropriate, and we report here the compositional analysis the simplex discussion (\S\ref{sec:simplex}) calls for. We map each allocation to four isometric log-ratio (ilr) coordinates after multiplicative replacement of zero shares, and regress each domain's mid-training-only accuracy on a response surface in ilr space over the 42-allocation pool (the 30 sweep and held-out allocations plus the 12 interior allocations of Appendix~\ref{app:expA}).

The result is informative in both directions. A full quadratic ilr surface fits well ($R^2 = 0.78$--$0.96$ per domain over the 42-allocation pool), which confirms that allocation carries real signal when modelled compositionally rather than marginally. But its stationary point is a saddle in every domain, not an interior maximum, and with 15 quadratic terms estimated from 42 allocations the Hessian is too weakly determined for that classification to carry weight. A reduced surface without cross terms (9 parameters) is more stable and predominantly concave: 15 of the 20 ilr curvature coefficients are negative (Logic 4 of 4; Cipher, Operation and Counterfactual 3 of 4; Puzzle 2 of 4), so the surface curves downward in most directions but not all.

We therefore do not claim a joint interior optimum over the simplex. The moderation result of Appendix~\ref{app:bands} is a statement about per-domain marginals---each domain's own coverage band---and it stands on the model-free band comparison and the permutation test, neither of which requires a joint surface. Establishing a joint compositional optimum would require more allocations than the 42 trained here. The 12 interior allocations of Appendix~\ref{app:expA} were added precisely to test this and did not produce a joint maximum---every domain's stationary point remained a saddle---so we flag additional simplex coverage, not a different estimator, as the natural next design rather than a gap the present data can close.

\subsection{Fitting Procedure}
\label{app:fitting}

Each domain's dose--response curve is summarised by the following \emph{asymmetric linear-space Gaussian} (split Gaussian):
\begin{equation}
\begin{split}
  f(\theta) &= a\exp\!\left(-\tfrac{1}{2}\!\left(\tfrac{\theta-\mu}{\sigma(\theta)}\right)^{\!2}\right)+c,\\[2pt]
  \sigma(\theta) &= \begin{cases}\sigma_L, & \theta\le\mu,\\\sigma_R, & \theta>\mu,\end{cases}
\end{split}
\label{eq:asym_gauss}
\end{equation}
where the two widths $\sigma_L$ and $\sigma_R$ are free parameters with no ordering imposed, with five free parameters $(a,\,\mu,\,\sigma_L,\,\sigma_R,\,c)$ per domain. In the fitted curves, $\sigma_R{>}\sigma_L$ (steeper left rise than right decline) for four domains, while Puzzle's fitted curve has $\sigma_L{>}\sigma_R$. Parameters are estimated by weighted nonlinear least squares (\texttt{scipy.optimize.curve\_fit}). Configurations in which a domain's internal proportion is $\theta_d\le5\%$ or $\theta_d\ge55\%$ are assigned half-weight ($w=0.5$); all others receive $w=1.0$. This heuristic down-weighting reduces the leverage of extreme, near-degenerate points, but the exact thresholds are not claimed to be optimal.

Peak locations $\mu$ and peak accuracies per domain are listed in Table~\ref{tab:coverage_bands}. The 95\%-of-peak moderate region for each domain is the set $\{\theta_d \mid f_d(\theta_d) \ge 0.95{\cdot}f_d(\mu)\}$, solved numerically from the fitted split-Gaussian curve (Eq.~\ref{eq:asym_gauss}); the resulting intervals are listed in Table~\ref{tab:coverage_bands}.

\begin{table}[!htbp]
\centering
\footnotesize
\setlength{\tabcolsep}{4pt}
\begin{tabular*}{\textwidth}{@{\extracolsep{\fill}}lccc@{}}
\toprule
Domain & Peak (\%) & $\mu$ (\%) & 95\%-band \\
\midrule
Cipher         & 52.0 &  9.9 & [7.4,\;36.0] \\
Operation      & 78.1 & 15.9 & [11.9,\;29.4] \\
Logic          & 59.5 & 15.0 & [11.1,\;31.4] \\
Counterfactual & 83.5 & 26.0 & [17.9,\;47.7] \\
Puzzle         & 15.0 & 35.1 & [27.2,\;41.2] \\
\bottomrule
\end{tabular*}%
\caption{Fitted 95\%-of-peak intervals from split-Gaussian summaries. Each interval is the $\theta_d$ range satisfying $f_d(\theta_d)\ge 0.95\cdot f_d(\mu)$; the threshold is inclusive ($\ge$) and bounds are displayed rounded to one decimal place. These are descriptive function-level thresholds from the Qwen3-8B/KOR-Bench sweep under the simplex constraint (see Appendix~\ref{app:coverage_sweep}); they are not confidence intervals and not validated allocation rules. For Cipher, the fitted curve is dominated by a broad plateau ($\approx$51--53\% over $\approx$8--24\% coverage); the fitted peak at $\approx$9.9\% is not a distinct mode and should not be read as a sharp optimum.}
\label{tab:coverage_bands}
\end{table} \paragraph{Sensitivity to the down-weighting threshold.}
Because the $w=0.5$ threshold at $\theta_d\le5\%$ or $\theta_d\ge55\%$ is a heuristic, we refit the split-Gaussian across four alternative thresholds ($\le0/\ge100$, $\le3/\ge57$, $\le5/\ge55$, $\le8/\ge52$). The fitted peaks are essentially unchanged: Cipher $\mu\in[9.6,9.9]$, Operation $[15.9,15.9]$, Logic $[14.8,15.3]$, Counterfactual $[26.0,26.2]$, Puzzle $[34.9,35.4]$---a spread of at most $0.5$\,pp. The reported peaks are therefore not an artifact of this particular down-weighting choice. The $\theta_d\le5\%$ and $\theta_d\ge55\%$ thresholds are used throughout.

The illustrative allocation $\theta^*$ reported in §\ref{sec:coverage_impact} is obtained by solving:
\begin{equation}
\begin{split}
  \theta^* &= \operatorname*{arg\,max}_{\theta}\;\sum_d \frac{f_d(\theta_d)-b_d}{b_d}\\
  &\quad\text{s.t.}\quad \sum_d\theta_d=100\%,\;\theta_d\ge0,
\end{split}
\label{eq:opt}
\end{equation}
where $b_d$ is the FineWeb-Edu-only baseline accuracy, solved with Sequential Least Squares Programming (SLSQP; 2,000 random initialisations). Because this objective is defined on fitted marginal curves, the resulting allocation should be interpreted as a descriptive probe rather than a globally optimal mixture.

\paragraph{Goodness of fit.}
The split-Gaussian substantially outperforms a flat baseline on every domain. Weighted Root Mean Square Error (RMSE, in pp), split-Gaussian vs.\ weighted constant: Cipher 0.84 vs.\ 5.32; Operation 1.15 vs.\ 12.51; Logic 1.44 vs.\ 9.10; Counterfactual 1.18 vs.\ 10.60; Puzzle 0.63 vs.\ 3.28.

\paragraph{Model form comparison.}
Table~\ref{tab:model_comparison} compares the split-Gaussian with a quadratic OLS alternative ($y=\beta_0+\beta_1\theta+\beta_2\theta^2$, three parameters). The quadratic recovers concave shape with fewer assumptions and is the form we use for the calibrated interiority test of Appendix~\ref{app:bands}, precisely because its three parameters cannot manufacture an interior optimum from noise; the $\hat\beta_2$ values of Table~\ref{tab:app-quadratic} are shape descriptors accompanying that test, not a curvature test in themselves. For describing the curves, the split-Gaussian is preferred on evidence rather than flexibility: it attains both the lower weighted in-sample RMSE and the lower leave-one-out MAE in all five domains, and the LOO-MAE comparison does not reward its extra parameters. It additionally captures the left--right asymmetry the quadratic cannot represent. Both forms agree that the curves are non-monotonic within this sweep, with the strongest fit for Counterfactual.

\begin{table}[!ht]
\centering
\footnotesize
\setlength{\tabcolsep}{4pt}
\caption{Model fit comparison for the per-domain coverage--accuracy curves ($n=24$). All three forms are scored on the same footing: RMSE is the weighted in-sample RMSE under the identical weighting scheme (Appendix~\ref{app:fitting}), and LOO-MAE is leave-one-out mean absolute error, which is the out-of-sample criterion and does not reward extra parameters. The split-Gaussian attains the lower LOO-MAE in all five domains, so its advantage is not an artifact of its larger parameter count. RMSE and LOO-MAE in pp.}
\label{tab:model_comparison}
\begin{tabular*}{\textwidth}{@{\extracolsep{\fill}}lcccccc@{}}
\toprule
 & \multicolumn{2}{c}{Flat baseline} & \multicolumn{2}{c}{Quadratic OLS} & \multicolumn{2}{c}{Split-Gaussian} \\
\cmidrule(lr){2-3}\cmidrule(lr){4-5}\cmidrule(lr){6-7}
Domain & RMSE & LOO-MAE & RMSE & LOO-MAE & RMSE & LOO-MAE \\
\midrule
Cipher         &  5.32 &  4.28 & 3.95 & 4.22 & 0.84 & 1.10 \\
Operation      & 12.51 & 12.34 & 5.20 & 7.67 & 1.15 & 1.01 \\
Logic          &  9.10 &  8.65 & 4.28 & 6.36 & 1.44 & 1.16 \\
Counterfactual & 10.60 & 10.49 & 1.72 & 1.95 & 1.18 & 1.09 \\
Puzzle         &  3.28 &  3.16 & 0.71 & 0.66 & 0.63 & 0.55 \\
\bottomrule
\end{tabular*}%
\end{table}

\paragraph{Leave-one-out cross-validation.}
We apply LOO-CV over the 24 normal non-baseline configurations. For each fold we refit the split Gaussian on the other 23 configurations using the same deterministic multi-start procedure (four $\sigma$ seed pairs, two intercept seeds, and five $\mu$ seeds drawn from the sorted-top-3 targets plus a uniform grid) and record the prediction error on the held-out configuration.

Table~\ref{tab:loo_cv} reports LOO-MAE, the range of fitted $\mu$ across folds ($\Delta\mu$), and the fraction of held-out points within $\pm2{\times}\mathrm{RMSE_{full}}$ (in-band), all over the 24 normal non-baseline configurations. LOO-MAE is small and tight (0.55--1.16\,pp), and peak location is stable across folds for every domain ($\Delta\mu\le 1.78$\,pp). These diagnostics assess interpolation for the finite sweep only; they do not validate peak locations, moderate bands, or the jointly selected $\theta^*$ on new mixtures.

\begin{table}[!ht]
\centering
\footnotesize
\setlength{\tabcolsep}{4pt}
\caption{Leave-one-out cross-validation results for the split-Gaussian dose-response fits. LOO-MAE: mean absolute error on held-out; $\mu_\text{full}$: peak from full-data fit; $\Delta\mu$ LOO: range of $\mu$ across LOO folds; In-band: fraction within $\pm2{\times}\mathrm{RMSE_{full}}$.}
\label{tab:loo_cv}
\begin{tabular*}{\textwidth}{@{\extracolsep{\fill}}lcccc@{}}
\toprule
Domain & LOO-MAE (pp) & $\mu_\text{full}$ (\%) & $\Delta\mu$ LOO (pp) & In-band \\
\midrule
Cipher         &  1.10 &  9.9 & 1.04 & 21/24 (88\%) \\
Operation      &  1.01 & 15.9 & 0.94 & 21/24 (88\%) \\
Logic          &  1.16 & 15.0 & 1.78 & 20/24 (83\%) \\
Counterfactual &  1.09 & 26.0 & 0.58 & 22/24 (92\%) \\
Puzzle         &  0.55 & 35.1 & 1.11 & 22/24 (92\%) \\
\bottomrule
\end{tabular*}%
\end{table}

One sweep configuration (Cipher\,26\%, Operation\,16\%, Logic\,17\%, Counterfactual\,30\%, Puzzle\,11\%), added after an initial set of eight configurations to broaden simplex coverage, realizes 82.4\% Counterfactual accuracy at 30\% Counterfactual coverage---close to, but not above, the 83.9\%, 83.7\%, and 83.6\% attained near 29\%, 32\%, and 24\% Counterfactual coverage. Because several configurations across 24--32\% Counterfactual coverage cluster around this level, the fitted Counterfactual peak ($\approx$26\%; Table~\ref{tab:coverage_bands}) is supported by a group of points rather than by a single post-hoc observation, and no single configuration establishes a higher peak. The pattern remains directionally consistent with the relatively higher Counterfactual allocation in the illustrative $\theta^*$ (a separate, analytically derived allocation trained as an independent checkpoint outside the 24 sweep configurations; Table~\ref{tab:mid_train}), but a single configuration does not validate the fitted Counterfactual peak. To confirm that this design point does not drive the fitted curves, we refit all five domains with it removed. The peaks move by at most $0.20$\,pp (Cipher $9.95\rightarrow10.00$, Operation $15.90\rightarrow15.95$, Logic unchanged at $14.95$, Counterfactual $26.05\rightarrow26.25$, Puzzle unchanged at $35.10$) and their sum changes from $101.95\%$ to $102.25\%$, so every peak location and every 95\%-of-peak interval reported here is materially unchanged by excluding the post-hoc configuration.

\begingroup
\footnotesize
\setlength{\tabcolsep}{4pt}
\setlength{\LTcapwidth}{\textwidth}
\begin{longtable}{@{\extracolsep{\fill}}lccccc@{}}
\caption{Mid-training coverage sweep results with the compensatory-SFT pass and its uniform-SFT control. Each row block reports: \textbf{Portion} (domain coverage proportions), \textbf{Evaluation} (mid-training-only accuracy), \textbf{Remedial SFT} (compensatory allocation from Eq.~\ref{eq:comp_sft}), \textbf{Eval.\ (Remedial)} (accuracy after the compensatory pass), and \textbf{Eval.\ (Uniform)} (accuracy after the matched uniform control, whose data mix is identical across configurations; Appendix~\ref{app:comp_sft}); the six held-out blocks additionally carry an \textbf{RL Evaluation} row (accuracy after the full mid-training$+$SFT$+$RL pipeline, Appendix~\ref{app:comp_sft}). Values are reported as mean $\pm$ 1 SD across the five seeds. The \textbf{Exploratory $\theta^*$} block (Expt.~3 in Table~\ref{tab:main}) reports the analytically derived $\theta^*$ allocation (9.7/11.8/15.8/30.4/32.3), its target Remedial SFT allocation from Eq.~\ref{eq:comp_sft}, and the results of the dedicated re-run (five seeds); it does not coincide with any of the 24 sweep configurations. The \textbf{FineWeb-Edu} baseline block precedes six \textbf{held-out allocations} that are withheld from the curve fit and used as out-of-sample validation (mid-training-only, SFT-stage and RL-stage results; \S\ref{sec:coverage_impact}, Appendix~\ref{app:comp_sft}).}
\label{tab:mid_train}\\
\toprule
Domains & Ciphers & Operations & Logic & Counterfactual & Puzzles \\
\midrule
\endfirsthead
\multicolumn{6}{l}{\footnotesize Table~\ref{tab:mid_train} (continued)}\\
\toprule
Domains & Ciphers & Operations & Logic & Counterfactual & Puzzles \\
\midrule
\endhead
\midrule
\multicolumn{6}{r}{\footnotesize \textit{Continued on next page}}\\
\endfoot
\endlastfoot
Portion              & 72.3\% &   0.0\% &  12.0\% &   0.0\% &  15.7\% \\
Evaluation           & $38.1\pm2.6$\% & $33.2\pm2.8$\% & $54.5\pm2.2$\% & $45.6\pm2.8$\% & $11.6\pm1.5$\% \\
Remedial SFT         &   0.0\%$^\dagger$ &  28.3\% &  22.6\% &  28.3\% &  20.8\% \\
Eval.\ (Remedial)     & $38.3\pm0.5$\% & $41.5\pm0.4$\% & $58.4\pm0.8$\% & $53.5\pm0.8$\% & $13.2\pm0.3$\% \\
Eval.\ (Uniform)     & $41.5\pm0.7$\% & $41.6\pm0.3$\% & $59.3\pm0.5$\% & $50.8\pm0.2$\% & $13.8\pm0.7$\% \\
\midrule
Portion              & 26.0\% &  16.0\% &  17.0\% &  30.0\% &  11.0\% \\
Evaluation           & $50.5\pm2.1$\% & $77.9\pm2.3$\% & $59.8\pm1.9$\% & $82.4\pm3.2$\% & $8.9\pm1.0$\% \\
Remedial SFT         &  17.0\% &  22.0\% &  21.5\% &  15.0\% &  24.5\% \\
Eval.\ (Remedial)     & $53.0\pm0.6$\% & $83.8\pm0.4$\% & $64.6\pm1.0$\% & $87.7\pm0.8$\% & $10.8\pm0.3$\% \\
Eval.\ (Uniform)     & $54.2\pm0.7$\% & $83.4\pm0.3$\% & $64.4\pm1.0$\% & $86.5\pm0.8$\% & $11.2\pm0.9$\% \\
\midrule
Portion              & 30.0\% &  18.0\% &  16.0\% &  18.0\% &  18.0\% \\
Evaluation           & $50.9\pm2.9$\% & $77.6\pm2.8$\% & $60.4\pm1.6$\% & $79.3\pm1.8$\% & $12.0\pm1.4$\% \\
Remedial SFT         &  15.0\% &  21.0\% &  22.0\% &  21.0\% &  21.0\% \\
Eval.\ (Remedial)     & $54.2\pm0.6$\% & $83.3\pm0.3$\% & $65.2\pm0.6$\% & $86.0\pm1.0$\% & $13.9\pm0.5$\% \\
Eval.\ (Uniform)     & $54.6\pm0.4$\% & $83.0\pm0.3$\% & $65.1\pm1.1$\% & $83.5\pm0.8$\% & $14.1\pm0.6$\% \\
\midrule
Portion              & 20.0\% &  20.0\% &  20.0\% &  20.0\% &  20.0\% \\
Evaluation           & $51.1\pm2.2$\% & $77.2\pm2.9$\% & $59.3\pm2.2$\% & $81.2\pm3.1$\% & $12.5\pm2.0$\% \\
Remedial SFT         &  20.0\% &  20.0\% &  20.0\% &  20.0\% &  20.0\% \\
Eval.\ (Remedial)     & $53.5\pm0.9$\% & $82.9\pm0.3$\% & $63.7\pm0.5$\% & $87.8\pm0.9$\% & $14.3\pm0.2$\% \\
Eval.\ (Uniform)     & $54.9\pm0.7$\% & $82.6\pm0.5$\% & $63.9\pm0.7$\% & $85.4\pm0.6$\% & $14.6\pm0.8$\% \\
\midrule
Portion              &   0.0\% &  10.0\% &  40.0\% &  10.0\% &  40.0\% \\
Evaluation           & $22.0\pm2.3$\% & $67.3\pm3.8$\% & $52.8\pm1.7$\% & $66.9\pm2.0$\% & $15.2\pm0.8$\% \\
Remedial SFT         &  30.0\% &  25.0\% &  10.0\% &  25.0\% &  10.0\% \\
Eval.\ (Remedial)     & $26.6\pm0.7$\% & $73.6\pm0.3$\% & $55.6\pm1.0$\% & $74.7\pm0.5$\% & $15.9\pm0.4$\% \\
Eval.\ (Uniform)     & $26.8\pm0.8$\% & $72.9\pm0.2$\% & $57.2\pm0.4$\% & $71.3\pm0.8$\% & $17.1\pm1.0$\% \\
\midrule
Portion              & 10.0\% &   8.0\% &  37.0\% &   8.0\% &  37.0\% \\
Evaluation           & $51.2\pm2.5$\% & $64.5\pm3.3$\% & $54.3\pm2.5$\% & $65.2\pm2.8$\% & $14.9\pm1.0$\% \\
Remedial SFT         &  25.0\% &  26.0\% &  11.5\% &  26.0\% &  11.5\% \\
Eval.\ (Remedial)     & $55.2\pm0.8$\% & $71.9\pm0.5$\% & $57.4\pm0.5$\% & $73.8\pm0.6$\% & $15.8\pm0.1$\% \\
Eval.\ (Uniform)     & $55.2\pm0.8$\% & $70.2\pm0.8$\% & $58.7\pm0.4$\% & $69.7\pm0.9$\% & $16.8\pm0.7$\% \\
\midrule
Portion              &   5.0\% &  15.0\% &   5.0\% &  15.0\% &  60.0\% \\
Evaluation           & $45.7\pm1.6$\% & $77.7\pm3.1$\% & $44.8\pm2.8$\% & $75.7\pm2.3$\% & $5.3\pm0.9$\% \\
Remedial SFT         &  27.5\% &  22.5\% &  27.5\% &  22.5\% &   0.0\% \\
Eval.\ (Remedial)     & $50.2\pm0.5$\% & $83.7\pm0.4$\% & $50.3\pm0.7$\% & $82.7\pm0.4$\% & $5.0\pm0.5$\% \\
Eval.\ (Uniform)     & $49.9\pm0.5$\% & $83.2\pm0.6$\% & $49.8\pm0.8$\% & $80.0\pm0.7$\% & $7.0\pm0.8$\% \\
\midrule
Portion              & 50.0\% &  40.0\% &   0.0\% &   5.0\% &   5.0\% \\
Evaluation           & $47.0\pm1.8$\% & $65.7\pm2.7$\% & $22.1\pm2.7$\% & $57.5\pm3.0$\% & $8.0\pm1.1$\% \\
Remedial SFT         &   5.0\% &  10.0\% &  30.0\% &  27.5\% &  27.5\% \\
Eval.\ (Remedial)     & $48.7\pm0.7$\% & $69.4\pm0.5$\% & $27.4\pm0.8$\% & $65.5\pm0.4$\% & $10.3\pm0.1$\% \\
Eval.\ (Uniform)     & $50.5\pm0.5$\% & $70.9\pm0.6$\% & $27.7\pm1.0$\% & $62.6\pm0.9$\% & $10.5\pm0.9$\% \\
\midrule
Portion              & 40.0\% &   4.0\% &   8.0\% &  40.0\% &   8.0\% \\
Evaluation           & $49.5\pm1.4$\% & $47.7\pm3.4$\% & $50.2\pm1.9$\% & $82.0\pm1.6$\% & $8.2\pm0.7$\% \\
Remedial SFT         &  10.0\% &  28.0\% &  26.0\% &  10.0\% &  26.0\% \\
Eval.\ (Remedial)     & $52.6\pm0.8$\% & $54.4\pm1.0$\% & $55.2\pm0.6$\% & $86.1\pm1.1$\% & $10.4\pm0.1$\% \\
Eval.\ (Uniform)     & $53.1\pm0.4$\% & $55.6\pm0.4$\% & $55.1\pm0.5$\% & $86.0\pm0.9$\% & $10.6\pm0.9$\% \\
\midrule
Portion              &   8.0\% &  36.0\% &   2.0\% &   2.0\% &  52.0\% \\
Evaluation           & $49.5\pm1.9$\% & $71.8\pm3.2$\% & $33.7\pm2.2$\% & $52.8\pm2.2$\% & $9.6\pm1.1$\% \\
Remedial SFT         &  26.0\% &  12.0\% &  29.0\% &  29.0\% &   4.0\% \\
Eval.\ (Remedial)     & $53.8\pm0.7$\% & $76.1\pm0.4$\% & $39.3\pm0.5$\% & $60.2\pm0.8$\% & $9.8\pm0.4$\% \\
Eval.\ (Uniform)     & $53.6\pm0.6$\% & $77.0\pm0.7$\% & $38.9\pm0.9$\% & $60.6\pm0.4$\% & $11.4\pm0.6$\% \\
\midrule
Portion              &  9.0\% &  14.0\% &  14.0\% &  17.0\% &  46.0\% \\
Evaluation           & $52.8\pm1.3$\% & $77.9\pm2.4$\% & $59.2\pm2.6$\% & $78.3\pm3.4$\% & $12.6\pm0.7$\% \\
Remedial SFT         &  25.5\% &  23.0\% &  23.0\% &  21.5\% &   7.0\% \\
Eval.\ (Remedial)     & $57.1\pm0.9$\% & $83.9\pm0.9$\% & $64.3\pm0.5$\% & $85.2\pm1.1$\% & $13.4\pm0.4$\% \\
Eval.\ (Uniform)     & $56.8\pm0.4$\% & $83.4\pm0.6$\% & $63.9\pm0.7$\% & $82.5\pm0.4$\% & $14.4\pm0.6$\% \\
\midrule
Portion              & 11.0\% &   2.0\% &  15.0\% &  28.0\% &   44.0\% \\
Evaluation           & $51.1\pm2.7$\% & $41.8\pm3.4$\% & $59.9\pm2.5$\% & $83.4\pm3.6$\% & $13.3\pm0.8$\% \\
Remedial SFT         &  24.5\% &  29.0\% &  22.5\% &  16.0\% &   8.0\% \\
Eval.\ (Remedial)     & $55.3\pm0.9$\% & $46.8\pm0.8$\% & $64.8\pm1.2$\% & $89.2\pm0.9$\% & $14.1\pm0.4$\% \\
Eval.\ (Uniform)     & $55.1\pm0.5$\% & $47.8\pm0.3$\% & $64.6\pm0.6$\% & $87.5\pm0.3$\% & $15.1\pm0.7$\% \\
\midrule
Portion              &  4.0\% &   3.0\% &  30.0\% &  29.0\% &   34.0\% \\
Evaluation           & $40.3\pm1.1$\% & $45.6\pm2.8$\% & $57.3\pm2.1$\% & $83.9\pm3.8$\% & $15.4\pm1.3$\% \\
Remedial SFT         &  28.0\% &  28.5\% &  15.0\% &  15.5\% &  13.0\% \\
Eval.\ (Remedial)     & $44.8\pm0.6$\% & $52.3\pm0.7$\% & $61.0\pm1.0$\% & $89.4\pm0.6$\% & $16.6\pm0.4$\% \\
Eval.\ (Uniform)     & $44.6\pm0.3$\% & $51.5\pm0.1$\% & $61.8\pm0.3$\% & $88.0\pm0.5$\% & $17.3\pm0.5$\% \\
\midrule
Portion              & 15.0\% &   6.0\% &  19.0\% &  32.0\% &   28.0\% \\
Evaluation           & $52.0\pm2.9$\% & $57.3\pm2.3$\% & $59.4\pm2.9$\% & $83.7\pm1.7$\% & $14.7\pm1.4$\% \\
Remedial SFT         &  22.5\% &  27.0\% &  20.5\% &  14.0\% &  16.0\% \\
Eval.\ (Remedial)     & $56.0\pm0.5$\% & $63.8\pm0.7$\% & $64.0\pm1.2$\% & $88.9\pm1.1$\% & $16.2\pm0.2$\% \\
Eval.\ (Uniform)     & $55.9\pm0.8$\% & $63.1\pm0.8$\% & $64.0\pm0.5$\% & $87.7\pm0.6$\% & $16.7\pm0.9$\% \\
\midrule
Portion              & 60.0\% &  22.0\% &  13.0\% &   3.0\% &    2.0\% \\
Evaluation           & $41.6\pm1.4$\% & $76.9\pm2.1$\% & $57.6\pm2.4$\% & $52.1\pm2.4$\% & $4.2\pm1.0$\% \\
Remedial SFT         &   0.0\% &  19.0\% &  23.5\% &  28.5\% &  29.0\% \\
Eval.\ (Remedial)     & $41.6\pm0.8$\% & $81.3\pm0.8$\% & $62.6\pm0.8$\% & $60.1\pm1.0$\% & $6.6\pm0.1$\% \\
Eval.\ (Uniform)     & $45.0\pm0.3$\% & $82.3\pm0.6$\% & $62.3\pm0.8$\% & $56.8\pm0.6$\% & $6.9\pm0.7$\% \\
\midrule
Portion              & 12.0\% &  60.0\% &  24.0\% &   1.0\% &    3.0\% \\
Evaluation           & $52.9\pm1.7$\% & $43.5\pm3.2$\% & $58.6\pm1.8$\% & $52.1\pm2.2$\% & $4.2\pm1.2$\% \\
Remedial SFT         &  24.0\% &   0.0\% &  18.0\% &  29.5\% &  28.5\% \\
Eval.\ (Remedial)     & $57.1\pm0.6$\% & $43.5\pm0.7$\% & $62.8\pm1.2$\% & $60.2\pm0.7$\% & $6.6\pm0.2$\% \\
Eval.\ (Uniform)     & $56.9\pm0.5$\% & $48.5\pm0.4$\% & $63.1\pm0.6$\% & $57.1\pm0.5$\% & $6.8\pm0.8$\% \\
\midrule
Portion              &  6.0\% &  28.0\% &  28.0\% &  24.0\% &   14.0\% \\
Evaluation           & $45.2\pm2.3$\% & $75.2\pm3.3$\% & $57.5\pm1.7$\% & $83.6\pm3.9$\% & $10.1\pm0.6$\% \\
Remedial SFT         &  27.0\% &  16.0\% &  16.0\% &  18.0\% &  23.0\% \\
Eval.\ (Remedial)     & $49.6\pm0.5$\% & $80.2\pm0.5$\% & $62.4\pm1.1$\% & $89.7\pm0.5$\% & $12.1\pm0.5$\% \\
Eval.\ (Uniform)     & $49.4\pm0.4$\% & $80.5\pm0.4$\% & $62.0\pm0.9$\% & $87.7\pm1.0$\% & $12.3\pm0.5$\% \\
\midrule
Portion              &  7.0\% &  11.0\% &  18.0\% &  37.0\% &   27.0\% \\
Evaluation           & $48.0\pm1.6$\% & $72.8\pm2.2$\% & $59.7\pm2.8$\% & $83.2\pm3.7$\% & $14.4\pm0.9$\% \\
Remedial SFT         &  26.5\% &  24.5\% &  21.0\% &  11.5\% &  16.5\% \\
Eval.\ (Remedial)     & $52.4\pm0.8$\% & $79.0\pm0.5$\% & $64.4\pm0.9$\% & $87.7\pm0.6$\% & $15.9\pm0.5$\% \\
Eval.\ (Uniform)     & $52.1\pm0.6$\% & $78.4\pm0.5$\% & $64.3\pm0.7$\% & $87.2\pm0.7$\% & $16.4\pm1.0$\% \\
\midrule
Portion              & 18.0\% &   1.0\% &  60.0\% &   6.0\% &   15.0\% \\
Evaluation           & $52.3\pm2.5$\% & $37.7\pm2.7$\% & $40.0\pm1.6$\% & $64.6\pm3.8$\% & $10.2\pm1.0$\% \\
Remedial SFT         &  21.0\% &  29.5\% &   0.0\% &  27.0\% &  22.5\% \\
Eval.\ (Remedial)     & $56.2\pm0.8$\% & $44.5\pm0.5$\% & $39.9\pm0.9$\% & $72.3\pm0.5$\% & $12.2\pm0.2$\% \\
Eval.\ (Uniform)     & $56.1\pm0.7$\% & $43.9\pm0.8$\% & $44.2\pm0.8$\% & $70.2\pm0.3$\% & $12.4\pm0.9$\% \\
\midrule
Portion              & 24.0\% &  13.0\% &   7.0\% &  26.0\% &   30.0\% \\
Evaluation           & $51.4\pm2.6$\% & $77.9\pm3.7$\% & $46.6\pm2.3$\% & $83.0\pm3.1$\% & $14.8\pm0.6$\% \\
Remedial SFT         &  18.0\% &  23.5\% &  26.5\% &  17.0\% &  15.0\% \\
Eval.\ (Remedial)     & $55.0\pm0.8$\% & $84.0\pm0.6$\% & $52.9\pm0.9$\% & $88.9\pm0.7$\% & $16.2\pm0.2$\% \\
Eval.\ (Uniform)     & $55.1\pm0.6$\% & $83.4\pm0.7$\% & $51.5\pm0.6$\% & $87.1\pm0.3$\% & $16.8\pm0.8$\% \\
\midrule
Portion              & 19.0\% &   7.0\% &  19.0\% &  24.0\% &   31.0\% \\
Evaluation           & $51.3\pm1.2$\% & $60.2\pm3.6$\% & $59.7\pm2.4$\% & $82.5\pm1.7$\% & $15.1\pm0.8$\% \\
Remedial SFT         &  20.5\% &  26.5\% &  20.5\% &  18.0\% &  14.5\% \\
Eval.\ (Remedial)     & $55.2\pm0.9$\% & $66.7\pm1.0$\% & $64.3\pm0.7$\% & $88.6\pm0.5$\% & $16.4\pm0.3$\% \\
Eval.\ (Uniform)     & $55.1\pm0.6$\% & $65.9\pm0.7$\% & $64.3\pm1.0$\% & $86.6\pm0.5$\% & $17.1\pm0.6$\% \\
\midrule
Portion              & 22.0\% &   9.0\% &   1.0\% &  35.0\% &   33.0\% \\
Evaluation           & $50.8\pm2.1$\% & $67.4\pm2.5$\% & $29.2\pm2.0$\% & $82.6\pm3.6$\% & $15.0\pm0.9$\% \\
Remedial SFT         &  19.0\% &  25.5\% &  29.5\% &  12.5\% &  13.5\% \\
Eval.\ (Remedial)     & $54.5\pm0.6$\% & $73.7\pm0.6$\% & $34.9\pm0.8$\% & $87.4\pm0.9$\% & $16.2\pm0.5$\% \\
Eval.\ (Uniform)     & $54.6\pm0.7$\% & $73.0\pm0.2$\% & $34.6\pm0.3$\% & $86.6\pm0.5$\% & $16.9\pm0.8$\% \\
\midrule
Portion              & 16.0\% &  21.0\% &   3.0\% &  31.0\% &   29.0\% \\
Evaluation           & $52.3\pm1.9$\% & $76.7\pm3.8$\% & $35.3\pm2.6$\% & $84.1\pm2.6$\% & $12.7\pm1.3$\% \\
Remedial SFT         &  22.0\% &  19.5\% &  28.5\% &  14.5\% &  15.5\% \\
Eval.\ (Remedial)     & $56.3\pm0.5$\% & $82.2\pm0.3$\% & $40.9\pm1.0$\% & $89.4\pm1.1$\% & $14.1\pm0.1$\% \\
Eval.\ (Uniform)     & $56.2\pm0.6$\% & $82.1\pm0.2$\% & $40.4\pm0.5$\% & $88.2\pm0.7$\% & $14.7\pm1.0$\% \\
\midrule
Portion              & 33.0\% &   5.0\% &   6.0\% &  50.0\% &    6.0\% \\
Evaluation           & $49.5\pm2.0$\% & $57.1\pm3.0$\% & $49.0\pm2.2$\% & $78.2\pm1.8$\% & $5.8\pm1.4$\% \\
Remedial SFT         &  13.5\% &  27.5\% &  27.0\% &   5.0\% &  27.0\% \\
Eval.\ (Remedial)     & $51.9\pm0.5$\% & $63.7\pm0.6$\% & $55.4\pm0.6$\% & $81.7\pm0.5$\% & $8.1\pm0.4$\% \\
Eval.\ (Uniform)     & $53.1\pm0.5$\% & $62.9\pm0.5$\% & $54.0\pm1.0$\% & $82.1\pm0.8$\% & $8.3\pm0.5$\% \\
\midrule
\multicolumn{6}{l}{\textit{Exploratory $\theta^*$ (Expt.~3): Cipher 9.7\%, Oper.\ 11.8\%, Logic 15.8\%, Counterf.\ 30.4\%, Puzzle 32.3\%; dedicated re-run, five seeds.}}\\
Portion        &  9.7\% & 11.8\% & 15.8\% & 30.4\% & 32.3\% \\
Evaluation  & $51.2\pm2.1$\% & $78.0\pm2.8$\% & $61.4\pm2.1$\% & $84.8\pm3.1$\% & $15.8\pm1.9$\% \\
Remedial SFT   & 25.2\% & 24.1\% & 22.1\% & 14.8\% & 13.8\% \\
Eval.\ (Remedial) & $55.7\pm0.9$\% & $84.2\pm0.9$\% & $66.2\pm1.1$\% & $90.3\pm0.9$\% & $17.2\pm0.2$\% \\
Eval.\ (Uniform)     & $55.2\pm0.3$\% & $83.6\pm0.9$\% & $66.1\pm0.9$\% & $88.9\pm0.4$\% & $17.7\pm0.6$\% \\
\midrule
\multicolumn{6}{l}{\textit{FineWeb-Edu only (100\%); no reasoning-domain data.}}\\
Evaluation  & $4.4\pm2.2$\% & $45.6\pm2.9$\% & $32.0\pm2.8$\% & $29.2\pm2.3$\% & $1.2\pm1.2$\% \\
\midrule
\multicolumn{6}{l}{\textit{6 held-out allocations (withheld from the curve fit; used for out-of-sample validation).}}\\
Portion              & 10.0\% & 10.0\% & 15.0\% & 30.0\% & 35.0\% \\
Evaluation           & $50.1\pm1.7\%$ & $69.2\pm2.4\%$ & $60.4\pm1.2\%$ & $82.5\pm1.6\%$ & $13.8\pm1.7\%$ \\
Remedial SFT         & 25.0\% & 25.0\% & 22.5\% & 15.0\% & 12.5\% \\
Eval.\ (Remedial)     & $55.3\pm1.1\%$ & $74.8\pm2.7\%$ & $64.9\pm1.9\%$ & $87.7\pm0.9\%$ & $14.9\pm1.2\%$ \\
Eval.\ (Uniform)      & $53.1\pm1.3\%$ & $75.2\pm1.7\%$ & $65.6\pm1.2\%$ & $85.9\pm1.8\%$ & $16.1\pm3.4\%$ \\
RL Evaluation         & $55.9\pm2.1\%$ & $75.9\pm3.2\%$ & $66.1\pm2.8\%$ & $88.8\pm3.9\%$ & $15.2\pm4.3\%$ \\
\midrule
Portion              & 9.7\% & 50.0\% & 10.3\% & 15.0\% & 15.0\% \\
Evaluation           & $49.8\pm2.0\%$ & $57.3\pm1.7\%$ & $55.0\pm2.3\%$ & $77.2\pm1.9\%$ & $10.7\pm0.9\%$ \\
Remedial SFT         & 25.2\% & 5.0\% & 24.8\% & 22.5\% & 22.5\% \\
Eval.\ (Remedial)     & $54.0\pm1.5\%$ & $59.1\pm1.8\%$ & $60.7\pm1.4\%$ & $84.7\pm2.1\%$ & $12.6\pm1.3\%$ \\
Eval.\ (Uniform)      & $53.9\pm1.2\%$ & $61.8\pm1.6\%$ & $59.9\pm1.3\%$ & $81.9\pm1.5\%$ & $13.0\pm1.0\%$ \\
RL Evaluation         & $55.2\pm2.2\%$ & $61.0\pm2.9\%$ & $62.8\pm4.2\%$ & $85.2\pm3.1\%$ & $13.3\pm2.5\%$ \\
\midrule
Portion              & 15.0\% & 11.2\% & 20.0\% & 8.8\% & 45.0\% \\
Evaluation           & $50.5\pm1.3\%$ & $79.1\pm2.7\%$ & $58.3\pm1.9\%$ & $67.4\pm1.5\%$ & $11.8\pm1.2\%$ \\
Remedial SFT         & 22.5\% & 24.4\% & 20.0\% & 25.6\% & 7.5\% \\
Eval.\ (Remedial)     & $54.4\pm1.2\%$ & $85.0\pm4.2\%$ & $62.9\pm1.6\%$ & $75.0\pm2.8\%$ & $12.2\pm1.4\%$ \\
Eval.\ (Uniform)      & $54.5\pm1.9\%$ & $84.9\pm1.7\%$ & $63.2\pm1.7\%$ & $72.8\pm1.9\%$ & $13.7\pm2.5\%$ \\
RL Evaluation         & $55.2\pm3.1\%$ & $86.1\pm4.2\%$ & $65.7\pm4.5\%$ & $75.6\pm1.8\%$ & $13.2\pm3.5\%$ \\
\midrule
Portion              & 4.2\% & 15.0\% & 15.8\% & 25.0\% & 40.0\% \\
Evaluation           & $40.8\pm4.0\%$ & $75.1\pm2.5\%$ & $62.2\pm2.8\%$ & $83.2\pm2.1\%$ & $12.5\pm1.4\%$ \\
Remedial SFT         & 27.9\% & 22.5\% & 22.1\% & 17.5\% & 10.0\% \\
Eval.\ (Remedial)     & $45.3\pm1.7\%$ & $81.0\pm1.0\%$ & $67.1\pm3.5\%$ & $89.2\pm1.2\%$ & $12.9\pm1.4\%$ \\
Eval.\ (Uniform)      & $45.0\pm1.6\%$ & $80.8\pm3.2\%$ & $67.0\pm1.0\%$ & $87.5\pm2.9\%$ & $14.5\pm2.4\%$ \\
RL Evaluation         & $47.0\pm2.3\%$ & $81.8\pm2.6\%$ & $68.2\pm3.1\%$ & $90.1\pm1.7\%$ & $14.9\pm3.1\%$ \\
\midrule
Portion              & 20.0\% & 4.6\% & 25.0\% & 30.4\% & 20.0\% \\
Evaluation           & $50.9\pm1.2\%$ & $57.9\pm1.3\%$ & $57.0\pm2.5\%$ & $85.1\pm1.8\%$ & $11.8\pm0.3\%$ \\
Remedial SFT         & 20.0\% & 27.7\% & 17.5\% & 14.8\% & 20.0\% \\
Eval.\ (Remedial)     & $54.5\pm2.3\%$ & $64.4\pm2.0\%$ & $61.2\pm1.7\%$ & $90.5\pm1.1\%$ & $13.5\pm2.1\%$ \\
Eval.\ (Uniform)      & $54.8\pm1.2\%$ & $63.9\pm1.3\%$ & $61.5\pm1.0\%$ & $89.2\pm0.7\%$ & $14.0\pm3.0\%$ \\
RL Evaluation         & $55.2\pm2.7\%$ & $65.1\pm2.3\%$ & $62.6\pm2.8\%$ & $90.9\pm2.0\%$ & $14.1\pm2.4\%$ \\
\midrule
Portion              & 25.0\% & 30.0\% & 7.7\% & 5.0\% & 32.3\% \\
Evaluation           & $50.0\pm1.4\%$ & $73.1\pm1.9\%$ & $45.3\pm2.2\%$ & $60.2\pm2.5\%$ & $15.6\pm0.7\%$ \\
Remedial SFT         & 17.5\% & 15.0\% & 26.2\% & 27.5\% & 13.8\% \\
Eval.\ (Remedial)     & $56.3\pm2.2\%$ & $78.0\pm3.6\%$ & $50.9\pm1.7\%$ & $68.1\pm3.4\%$ & $16.8\pm2.3\%$ \\
Eval.\ (Uniform)      & $56.9\pm1.5\%$ & $77.3\pm1.3\%$ & $50.5\pm1.4\%$ & $65.9\pm1.3\%$ & $17.2\pm1.0\%$ \\
RL Evaluation         & $57.0\pm4.1\%$ & $79.3\pm2.9\%$ & $53.7\pm2.9\%$ & $71.0\pm2.9\%$ & $17.4\pm3.7\%$ \\
\bottomrule
\end{longtable}

\noindent{\footnotesize $^\dagger$~Domains whose mid-training coverage is at or above the 60\% threshold receive $r=0$: Configs 1 (Cipher 72.3\%), 7 (Puzzle 60\%), 15 (Cipher 60\%), 16 (Operation 60\%), and 19 (Logic 60\%). Config~1 Cipher's zero-allocation cell gains only $+0.2$\,pp (§\ref{app:comp_sft}). Values are mean $\pm$ 1 SD across the five seeds.\par}
\endgroup

\FloatBarrier
\subsection{Out-of-Sample Predictive Validity of the Fitted Curves}
\label{app:oos}

The six held-out allocations (Table~\ref{tab:mid_train}, held-out blocks) are trained at allocations withheld from the curve fitting and carried through the complete mid-training$+$SFT$+$RL pipeline. The split-Gaussian fits generalize to them across all five domains: mean absolute held-out residuals are $1.1$\,pp for Counterfactual (all six points within $\pm2$ fit-$RMSE$), $1.0$\,pp for Puzzle, $1.3$\,pp for Cipher, and $1.8$\,pp for Logic (five of six), while Operation is the weakest match ($3.0$\,pp; two low-coverage points realize $\approx+6$\,pp above the curve), so Operation's left tail rises somewhat faster than the split-Gaussian assumes. Applying the same fixed compensatory budget to the six held-out checkpoints gives gains in all 30 configuration--domain cells (mean $+4.5$\,pp) yet closes $0/60$ pairs at the $5$\,pp threshold and only $5/60$ at the $10\%$ ratio (uniform control: $0/60$ and $8/60$), and compensation's only material advantage over uniform remains Counterfactual ($+6.6$ vs.\ $+4.6$\,pp). The RL stage adds $+1.19$\,pp on average---in line with the $+1.1$ to $+2.3$\,pp seen in the main experiments---and leaves the picture unchanged: after the complete pipeline the held-out allocations still close $0/60$ pairs at the $5$\,pp threshold and $7/60$ at the $10\%$ ratio.

The held-out allocations also provide the only direct test of the premise underlying $\theta^*$: that curves fitted at the mid-training stage carry information about \emph{final} accuracy. Fitting on the 24 sweep configurations alone and predicting the six withheld allocations, the predicted overall accuracy tracks the realized mid-training-only value almost exactly (Pearson $r=+0.954$, Spearman $\rho=+1.000$, mean absolute error $0.64$\,pp) and---the point that matters for allocation choice---it also predicts the realized \emph{post-RL} overall accuracy after the complete mid-training$+$SFT$+$RL pipeline ($r=+0.967$, $p=0.002$; $\rho=+0.886$). The mid-training ranking is largely preserved through post-training ($r=+0.975$ between mid-only and post-RL overall), so a coverage choice made before SFT is still visible after RL. The predictive relationship is not carried by any single held-out point: dropping each of the six in turn moves the mid-training-only correlation only within $[+0.941,+0.973]$ and the post-RL correlation within $[+0.954,+0.987]$, and a bootstrap over the six allocations gives $95\%$ intervals of $[+0.923,+0.999]$ and $[+0.926,+0.998]$ respectively.

For completeness, $\theta^*$ reaches $62.72$ overall on the compensatory-SFT scale while the six held-out allocations span $54.02$--$59.52$, but this is \emph{not} an out-of-sample validation of $\theta^*$: $\theta^*$ was chosen as the argmax of the fitted objective and is evaluated here under a different downstream recipe (rule-type-balanced SFT in Table~\ref{tab:main} versus the compensatory pass applied to the held-out checkpoints), so the comparison records the internal consistency of an argmax across two SFT recipes rather than evidence that a fit-selected allocation generalises. One component does \emph{not} survive this test: the relative-gain objective of Eq.~\ref{eq:opt}, which is what actually selects $\theta^*$, ranks the six held-out allocations only weakly ($r=+0.441$, $p=0.38$; $\rho=+0.257$), and unlike the predicted-accuracy correlations it is fragile: leaving out one allocation moves it anywhere in $[+0.058,+0.750]$ and its bootstrap interval $[-0.875,+0.976]$ spans zero. The baseline normalisation $b_d$ appears to be what degrades it, so the fitted curves are the trustworthy object here while the $b_d$-normalised objective layered on top of them is not; an allocation rule should optimise predicted accuracy directly rather than the relative-gain objective. What remains outside this test is $\theta^*$'s own full-pipeline row on the matched protocol: its $+4.36$\,pp figure in Table~\ref{tab:main} follows the standard rule-type-balanced SFT, whereas the held-out allocations were carried through the compensatory pass, so the two are not run on identical downstream recipes.

\section{Compensatory SFT: Domain-Augmented Alignment Does Not Close Mid-Training Gaps}
\label{app:comp_sft}

\subsection{Experimental Design}

To test whether \emph{compensatory} SFT can close inter-domain gaps, we construct a domain-weighted SFT dataset:
\begin{equation}
  r_d = \frac{\max(0,\; 60 - M_d)}{\displaystyle\sum_{j=1}^{5}\max(0,\; 60 - M_j)} \times 100\%,
  \label{eq:comp_sft}
\end{equation}
where $M_d$ is the mid-training token percentage for domain $d$ (read directly from each coverage configuration) and $r_d$ is the target SFT proportion for domain $d$.
Domains whose mid-training coverage is at or above 60\% receive no compensatory SFT allocation ($r_d = 0$); all other domains receive SFT budget in proportion to their shortfall from the 60\% threshold, with the allocation vector normalised to sum to 100\%.
The threshold 60\% was chosen because any single domain exceeding 60\% of the five-domain token budget already constitutes extreme over-allocation; every domain below this threshold is treated as under-served and eligible for remediation.
Applying Eq.~\ref{eq:comp_sft} to all 24 configurations, five configurations have a domain at or above the 60\% threshold (Configs 1, 7, 15, 16, and 19; see Table~\ref{tab:mid_train}); in all other configurations, all five domains fall below 60\% and therefore receive positive compensatory SFT.
Total SFT token count is held equal to the uniform baseline so that only the cross-domain distribution changes, not the overall training volume. The uniform-SFT control uses the identical rule-type-balanced data mix for every configuration (the same mix as the shared SFT baselines in Table~\ref{tab:main}) with the same total token budget and the same 3-epoch schedule as the compensatory pass, so the two passes differ only in the cross-domain allocation of the SFT data.

\subsection{Capability Gap Metric: Pairwise Closure Framework}
\label{app:pairwise}

We evaluate whether compensatory SFT closes inter-domain gaps by comparing every unordered pair of domains within each of the 24 coverage-sweep checkpoints, giving $24 \times \binom{5}{2} = 240$ pairs. The pair members are ordered by their post-compensatory accuracies (see below), so the gap is always reported as the stronger-minus-weaker domain after compensation. Because pairs share both configurations and domains, all pairwise closure summaries are descriptive and require configuration-clustered inference for population-level claims.
For each pair $(i,j)$ with mid-training accuracies $a^{\mathrm{mid}}_i$, $a^{\mathrm{mid}}_j$ and post-compensatory-SFT accuracies $a^{\mathrm{comp}}_i$, $a^{\mathrm{comp}}_j$, sorted such that $a^{\mathrm{comp}}_i \le a^{\mathrm{comp}}_j$, we compute two gap measures:
\begin{align}
  G^{\mathrm{diff}}_{ij}(a) &= a_j - a_i, \label{eq:gap_diff}\\
  G^{\mathrm{ratio}}_{ij}(a) &= 1 - \frac{a_i}{a_j}. \label{eq:gap_ratio}
\end{align}
The closure rate for gap measure $G$ is
\begin{equation}
  C_{ij} = \frac{G_{ij}(a^{\mathrm{mid}}) - G_{ij}(a^{\mathrm{comp}})}{G_{ij}(a^{\mathrm{mid}})}.
\end{equation}
A pair is counted as \emph{bridged} (meaningfully repaired) if the gap narrows by at least a metric-specific margin: an absolute reduction of at least $5$\,pp in the difference gap, or a relative closure rate $C_{ij}$ of at least $10\%$ in the ratio gap. A stricter $18\%$ ratio threshold is examined only as a sensitivity check. These thresholds are calibrated to the scale of the intervention and of the seed noise in this sweep rather than to a statistical null (\S\ref{app:comp_sft}), and we interpret the resulting bridged-pair counts descriptively. Equal absolute gains naturally produce larger relative gains for weaker domains.
Under the difference metric ($5$\,pp), $0/240$ pairs are bridged; under the ratio metric ($10\%$), $30/240$ pairs are bridged. These counts are descriptive because the 240 pairs share all training configurations and are not independent Bernoulli trials; the thresholds were selected for exploratory sensitivity analysis rather than preregistered confirmatory testing. We therefore do not interpret the associated nominal binomial $p$-values as population-level evidence. The results indicate that this fixed compensatory policy produces absolute gains and limited, heterogeneous gap changes, but they do not establish that stronger alignment interventions would fail. The same pattern holds on the six held-out allocations withheld from the fit ($6\times\binom{5}{2}=60$ pairs): $0/60$ bridged at $5$\,pp and $5/60$ at $10\%$ (uniform $0/60$ and $8/60$), so the non-closure generalizes beyond the 24-configuration sample.

\begin{figure}[!htbp]
\centering
\includegraphics[width=\textwidth]{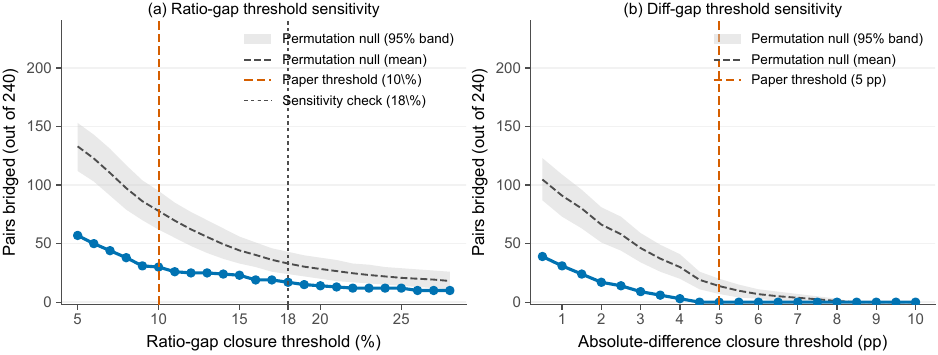}
\caption{Threshold sensitivity for pairwise gap-closure over the 240 domain pairs. Left: bridged pairs vs.\ ratio threshold ($5\%$--$28\%$). Right: same for difference threshold ($0.5$--$10$\,pp). The dashed line and shaded band are the mean and $95\%$ interval of a permutation null that reallocates the observed compensatory gains across domains within each configuration (2{,}000 draws), holding the gain magnitudes and the arithmetic budget cap fixed. The observed counts lie \emph{below} this null at both metrics ($P<0.001$), so the non-closure is not a mechanical consequence of the capped budget.}
\label{fig:threshold_sensitivity}
\end{figure}

\paragraph{Threshold sensitivity.}
Figure~\ref{fig:threshold_sensitivity} sweeps both thresholds over the 240 pairs: the bridged count falls from 57/240 at the most permissive ratio threshold ($5\%$) and 39/240 at the most permissive difference threshold ($0.5$\,pp) to $0/240$ from $4.5$\,pp upward. To calibrate these counts we use a permutation null rather than a ``half of all pairs'' reference line, which is not a null expectation: under a true no-effect null the expected number of \emph{bridged} pairs is near zero, not half. The null permutes, within each configuration, which domain receives which compensatory gain, holding the observed gain magnitudes---and hence the arithmetic budget cap---exactly fixed while destroying the alignment between a domain's gain and its position within each pair (2{,}000 reallocations). A random reallocation of the very same gains would bridge $13.8\pm3.3$ pairs under the $5$\,pp difference metric and $77.9\pm8.5$ under the $10\%$ ratio metric; the compensatory policy bridges $0$ and $30$ respectively ($P<0.001$ in both cases, one-sided). The observed non-closure is therefore not a mechanical consequence of the capped budget: the same gains, reallocated at random, would have closed substantially more gaps than the compensatory policy does.

\FloatBarrier
\subsection{A Stronger Reweighting Arm: Sharpening the Compensatory Policy}
\label{app:sharpen}

Observation~1 is stated for a policy family whose within-configuration gain differential is arithmetically bounded at $\approx7$\,pp, so a reader may reasonably ask whether the non-closure is an artifact of a policy too weak to succeed. Two analyses in this appendix already bear on that question. First, the feasible set is large: evaluated on the fitted gain curves, $192$ of the $240$ domain pairs ($80\%$) could in principle be narrowed by at least $5$\,pp by \emph{some} budget-feasible allocation of the same SFT tokens ($216/240$, $90\%$, under the $10\%$ ratio metric), against $0$ and $30$ actually closed. Second, the permutation null of Appendix~\ref{app:pairwise} shows the observed gains close fewer gaps than a random reallocation of the same magnitudes. Neither, however, answers the operational question: is there a \emph{stronger policy in the same family} that actually closes the gaps? This subsection reports the experiment that does.

\paragraph{Policy family.}
We generalise the compensatory rule of Eq.~\ref{eq:comp_sft} with a single sharpening exponent $\gamma$:
\begin{equation}
  r_d(\gamma) = \frac{[\max(0,\; T-M_d)]^{\gamma}}{\sum_{j=1}^{5}[\max(0,\; T-M_j)]^{\gamma}} \times 100\%,
  \qquad T=60\%,
  \label{eq:heavier_weight}
\end{equation}
where $M_d$ is domain $d$'s mid-training token percentage. The exponent controls how sharply the fixed SFT budget concentrates on coverage-deficient domains. At $\gamma=1$ the rule reduces exactly to Eq.~\ref{eq:comp_sft}, so the arm reported in the main text is the $\gamma=1$ member of this family and the comparison is nested rather than across unrelated policies; as $\gamma\to\infty$ it becomes winner-take-all, assigning the entire remedial budget to the single most deficient domain. Every branch holds the total SFT token count equal to the uniform baseline, exactly as at $\gamma=1$; only the cross-domain distribution changes. We ran $\gamma\in\{1,2,4,\infty\}$ on six sweep configurations that between them carry most of the feasible-but-unclosed pairs---each has $9$--$10$ of its $10$ domain pairs separated by $\ge 5$\,pp at the mid-training-only checkpoint---which required 90 additional SFT passes (three new branches $\times$ six configurations $\times$ five seeds; $\gamma=1$ was already trained) and no further mid-training or RL.

\paragraph{Result: closure is attainable, but it is bought with average accuracy.}
Table~\ref{tab:app-sharpen} reports the outcome over the $6\times\binom{5}{2}=60$ pairs these configurations contribute. Sharpening does close gaps that $\gamma=1$ leaves open---the count rises monotonically from $0/60$ at $\gamma=1$ to $12/60$ at $\gamma=\infty$ under the $5$\,pp metric---so the non-closure reported in the main text is \emph{not} an inescapable property of the budget. But closure is purchased directly out of average accuracy: the mean per-cell gain falls monotonically from $+4.34$\,pp to $+2.26$\,pp, so the winner-take-all branch that closes the most pairs also forfeits roughly half of the improvement the compensatory pass was introduced to deliver. The two objectives are in tension across the whole family we tested, and no branch achieves both: the branch that maximises mean gain closes nothing, and the branch that closes most gives up $2.1$\,pp of mean gain. This sharpens Observation~1 rather than overturning it---the gaps survive every setting that preserves the average gain---and it replaces an untested scope caveat with a measured trade-off. Even at $\gamma=\infty$, $48$ of the $60$ pairs remain unclosed at the $5$\,pp threshold.

\begin{table}[!ht]
\centering
\footnotesize
\setlength{\tabcolsep}{4pt}
\caption{Sharpened compensatory SFT over the six configurations carrying the most feasible-but-unclosed pairs (60 pairs). $\gamma=1$ is the policy reported in the main text. Mean gain is the per-cell accuracy change over the mid-training-only checkpoint, averaged over all 30 configuration--domain cells. Closure counts use the same two metrics as Appendix~\ref{app:pairwise}.}
\label{tab:app-sharpen}
\begin{tabular*}{\textwidth}{@{\extracolsep{\fill}}lccc@{}}
\toprule
Branch & Mean gain (pp) & Bridged, $5$\,pp & Bridged, $10\%$ ratio \\
\midrule
$\gamma=1$ (Eq.~\ref{eq:comp_sft}) & $+4.34$ & $0/60$  & $8/60$ \\
$\gamma=2$                         & $+4.17$ & $1/60$  & $11/60$ \\
$\gamma=4$                         & $+3.65$ & $5/60$  & $13/60$ \\
$\gamma=\infty$ (winner-take-all)  & $+2.26$ & $12/60$ & $12/60$ \\
\bottomrule
\end{tabular*}%
\end{table}

\subsection{Results}

Table~\ref{tab:mid_train} (Remedial SFT rows) reports compensatory allocations from Eq.~\ref{eq:comp_sft}; Eval.\ (Remedial) rows give post-compensatory-SFT accuracy. Compensatory SFT raises scores in 116 of 120 configuration--domain cells (mean $+4.32$\,pp; the four cells with no positive gain all receive zero compensatory allocation $r=0$). The re-run exploratory $\theta^*$ block, which is not one of the 24 sweep checkpoints, also gains in all five domains under compensation ($+1.4$ to $+6.2$\,pp; mean $+4.5$\,pp), with Puzzle's gain the smallest; its deviations from the fitted per-domain curves below are at most $\approx0.3$\,pp, so including or excluding this block does not change any conclusion.

\paragraph{Threshold calibration.}
The $10\%$ ratio and $5$\,pp difference cutoffs are calibrated to the scale of the intervention and of the seed noise in this sweep, not to a statistical null. The $5$\,pp difference threshold exceeds the largest cross-seed SD observed on any single configuration cell in Table~\ref{tab:mid_train} (mid-training-only SDs run 0.7--3.9\,pp over the 24 sweep configurations; the SFT-pass rows on those configurations have smaller SDs, though the six held-out blocks (which add the RL stage) carry a few larger values, up to $\approx$4.5\,pp), so a counted closure is larger than the seed-level dispersion of either endpoint. The $10\%$ ratio threshold is likewise beyond the seed-noise scale of the weaker-domain accuracies that dominate ratio-bridged pairs, and equal absolute gains naturally produce larger relative closures for weaker domains. The scan in Figure~\ref{fig:threshold_sensitivity} shows that the conclusion does not depend on the exact cutoffs: the bridged-pair count stays below the permutation null of \S\ref{app:pairwise} at every tested threshold (57/240 even at the most permissive $5\%$ ratio; 39/240 at $0.5$\,pp difference, and 0/240 from $4.5$\,pp upward). Under ratio ($10\%$), 30/240 pairs are bridged; under difference ($5$\,pp), 0/240 pairs are bridged. Both summaries indicate heterogeneous changes under this fixed policy, not general non-repairability.

These statistics underpin Figure~\ref{fig:comp_sft_pairwise_closure}. Redirecting SFT budget toward coverage-deficient domains raises accuracy but does not reliably close pairwise gaps created by mid-training.

\paragraph{Uniform-SFT control.}
As a control with the same data mix and budget, a uniform SFT pass (identical rule-type-balanced data mix across configurations; the same total token budget and 3-epoch schedule as the compensatory pass; Table~\ref{tab:app-hparams}) was applied to all 24 sweep checkpoints (Eval.\ (Uniform) rows, Table~\ref{tab:mid_train}). Table~\ref{tab:comp_uniform} compares the two passes. The uniform pass raises every one of the 120 cells (mean $+4.20$\,pp vs.\ $+4.32$\,pp compensatory), bridges the same 0/240 pairs under the $5$\,pp difference metric and 32/240 under the $10\%$ ratio metric (vs.\ 30/240), and produces per-domain mean gains close to the compensatory ones except for Counterfactual, where compensation is stronger by $+1.9$\,pp ($+6.37$ vs.\ $+4.49$\,pp), and Puzzle, where uniform SFT is slightly stronger ($+2.10$ vs.\ $+1.48$\,pp). Cross-configuration range narrowing is likewise comparable; uniform SFT narrows all five ranges (including Puzzle, $-0.7$\,pp), whereas compensation leaves Puzzle's range slightly wider ($+0.4$\,pp). The compensation formula therefore does not systematically outperform an equal-budget uniform re-training pass; its only material advantage is Counterfactual, the domain whose gain responds most steeply to the remedial ratio (below).

\begin{table}[!ht]
\centering
\footnotesize
\setlength{\tabcolsep}{4pt}
\caption{Per-domain mean accuracy gains (pp) of the compensatory and uniform SFT passes over the 24 sweep configurations (all 120 configuration--domain cells in the last row). Compensation exceeds the uniform control materially only for Counterfactual; it is slightly below the control for Cipher and Puzzle.}
\label{tab:comp_uniform}
\begin{tabular*}{\textwidth}{@{\extracolsep{\fill}}lccc@{}}
\toprule
Domain & Compensatory gain & Uniform gain & Difference (comp.$-$uniform) \\
\midrule
Cipher         & $+3.42$ & $+3.89$ & $-0.47$ \\
Operation      & $+5.73$ & $+5.77$ & $-0.04$ \\
Logic          & $+4.62$ & $+4.76$ & $-0.14$ \\
Counterfactual & $+6.37$ & $+4.49$ & $+1.88$ \\
Puzzle         & $+1.48$ & $+2.10$ & $-0.63$ \\
\midrule
All 120 cells  & $+4.32$ & $+4.20$ & $+0.12$ \\
\bottomrule
\end{tabular*}%
\end{table}

\paragraph{Gain vs.\ the remedial allocation: per-domain fits.}
Over the 115 configuration--domain cells with $r>0$, we fit, separately for each domain, the model
\begin{equation}
\Delta_d(r) \;\approx\; a_d\log(1+b_d\, r) + c_d ,
\label{eq:log1p_fit}
\end{equation}
where $r$ is the domain's remedial SFT proportion in percent (Figure~\ref{fig:sft_amplification}). The additive $+1$ inside the logarithm removes the absorption redundancy of the previous pooled form $\Delta\approx a\log(b\,r)+c$, in which $b$ and $c$ were not separately identifiable. Table~\ref{tab:comp_logfit} reports the fitted parameters. Within each domain the gain rises with $r$ and saturates at a domain-specific ceiling; the fits are tight (MAE 0.07--0.25\,pp; $R^2$ 0.71--0.97 over each domain's $r$-window), with the steepest rise across the observed window for Counterfactual (predicted $+3.2$\,pp at $r{=}5\%$ to $+8.2$\,pp at $r{=}29.5\%$) and the flattest for Puzzle ($+0.3$ to $+2.4$\,pp over $r\in[4,29]\%$). Domains differ more in overall gain level than in their response to $r$: a pooled fit of Eq.~\ref{eq:log1p_fit} over all 115 cells explains only $\approx31\%$ of the variance ($R^2{=}0.31$, MAE $1.36$\,pp; Table~\ref{tab:comp_logfit}), which is why the earlier pooled model appeared to describe a weak association. These fits are descriptive summaries; within each domain the observed $r$-window is narrow ($\approx$4--30\%), so $a_d$ and $b_d$ trade off against each other and, where the window is nearly linear (Cipher), the pair is only weakly identified individually even though the fitted curve is tight.

\begin{table}[!ht]
\centering
\footnotesize
\setlength{\tabcolsep}{4pt}
\caption{Per-domain fits of the compensatory gain $\Delta$ (pp) against the remedial ratio $r$ (\%) using $\Delta = a\log(1+b\,r)+c$ on the $r>0$ cells of Table~\ref{tab:mid_train} (Eq.~\ref{eq:log1p_fit}; Figure~\ref{fig:sft_amplification}). $R^2$ and MAE are computed on the fitted cells. The pooled row fits all 115 $r>0$ cells jointly. $^\dagger$Cipher's $a$ sits at its upper fit bound; over Cipher's observed window $\log(1+b r)\approx b r$, so the curve is effectively linear (slope $\approx+0.12$\,pp per percentage point of $r$) and $a,b$ are individually unstable.}
\label{tab:comp_logfit}
\begin{tabular*}{\textwidth}{@{\extracolsep{\fill}}lrrrrrr@{}}
\toprule
Domain & $n$ & $a$ & $b$ & $c$ & $R^2$ & MAE (pp) \\
\midrule
Cipher$^\dagger$         & 22 & 50.00 & 0.0024 & $+1.24$ & 0.78 & 0.25 \\
Operation      & 23 & 2.78 & 3.2420 & $-6.06$ & 0.71 & 0.25 \\
Logic          & 23 & 3.81 & 0.1351 & $-0.39$ & 0.77 & 0.25 \\
Counterfactual & 24 & 11.99 & 0.0237 & $+1.83$ & 0.95 & 0.19 \\
Puzzle         & 23 & 4.39 & 0.0273 & $-0.16$ & 0.97 & 0.07 \\
\midrule
Pooled         & 115 & 7.79 & 0.0360 & $+0.20$ & 0.31 & 1.36 \\
\bottomrule
\end{tabular*}%
\end{table}

\textbf{Boundary cases (zero-allocation domains).}
Five cells receive $r=0$ because their domain's mid-training coverage is at or above the 60\% threshold (Configs 1, 7, 15, 16, and 19; Table~\ref{tab:mid_train}). These cells show no meaningful gain under compensation (all five gains within $\pm0.3$\,pp, the largest being Config 1 Cipher at $+0.2$\,pp, which receives no remedial allocation at all), and the conclusions above hold when these configurations are excluded.

\paragraph{Budget sensitivity.}
Under our budget-proportional model, the within-configuration gain differential reaches at most $\approx\!7$\,pp; scaling total SFT volume would not increase this differential without changing the allocation formula. More aggressive allocations could produce larger differentials but remain untested. We scope our conclusion to this fixed-budget, fixed-formula setting.

\FloatBarrier
\section{External Benchmark Detailed Results}
\label{app:external}

\subsection{Coverage Sweep: External Benchmark Accuracy}
\label{app:ext_sweep}

Table~\ref{tab:ext_sweep} reports external benchmark accuracy for the coverage-sweep mid-training models (plus the FineWeb-Edu baseline), using the same model checkpoints as the KOR-Bench mid-training-only evaluation.
The external results are limited consistency checks rather than independent confirmation of a general transfer law. The external configuration set is fixed: the three external benchmarks evaluated on the nine coverage configurations (plus the FineWeb-Edu baseline) listed in Table~\ref{tab:ext_sweep}, which span the sweep from severely imbalanced to balanced allocations; Spearman rank associations of the KOR average with each external column over these configurations are reported in \S\ref{sec:cross_domain} ($n=9$). The external corpus draws 34.8\% of its tokens from the ProofWriter rule family, the ZebraLogic family contributes 201 mid-training samples, and the reported isolation does not establish template- or rule-family-level disjointness. The associations may therefore reflect shared structure or exposure as well as mixture allocation. The per-depth, per-house, and per-type comparisons below contrast the shared SFT+RL baseline with the Expt.\ 1 checkpoint (Cipher 72.3\%, Operation 0\%, Logic 12.0\%, Counterfactual 0\%, Puzzle 15.7\%; Table~\ref{tab:main}); the covered-depth ablation uses its own D3-/D5-heavy mixtures (below) and is not tied to any sweep configuration.

\begin{table}[!ht]
\centering
\footnotesize
\setlength{\tabcolsep}{4pt}
\caption{External benchmark accuracy (\%) for the coverage-sweep mid-training models (the FineWeb-Edu baseline is shown for reference). Coverage: Cipher\,/\,Operation\,/\,Logic\,/\,Counterfactual\,/\,Puzzle (internal); the ProofWriter rule family (34.8\% of external tokens) is fixed. KOR avg is the macro-average of the five KOR-Bench domain accuracies from Table~\ref{tab:mid_train} (\S\ref{app:coverage_sweep}); the external columns are five-seed means on the same checkpoints, matching the protocol of Table~\ref{tab:mid_train}. ``FineWeb-Edu'' = FineWeb-Edu-only baseline without KOR-Bench data. The nine coverage configurations listed here constitute the fixed external configuration set (\S\ref{sec:cross_domain}).}
\label{tab:ext_sweep}
\begin{tabular*}{\textwidth}{@{\extracolsep{\fill}}lccccc@{}}
\toprule
Coverage (Cipher/Oper./Logic/Counterf./Puzzle) & KOR avg & ProofWriter & ZebraLogic MC & ZebraLogic Grid & CounterBench \\
\midrule
FineWeb-Edu (100\%)            & 22.5 & 50.9 & 37.7 & 13.1 & 65.0 \\
40\,/\,4\,/\,8\,/\,40\,/\,8   & 47.5 & 80.7 & 45.8 & 21.0 & 70.7 \\
72.3\,/\,0\,/\,12\,/\,0\,/\,15.7 & 36.6 & 83.1 & 40.8 & 23.0 & 69.9 \\
0\,/\,10\,/\,40\,/\,10\,/\,40 & 44.8 & 85.1 & 31.6 & 37.8 & 76.4 \\
50\,/\,40\,/\,0\,/\,5\,/\,5   & 40.1 & 81.1 & 41.4 & 25.3 & 72.0 \\
5\,/\,15\,/\,5\,/\,15\,/\,60  & 49.8 & 81.2 & 37.4 & 26.2 & 72.5 \\
26\,/\,16\,/\,17\,/\,30\,/\,11 & 55.9 & 80.7 & \textbf{46.8} & 26.9 & 71.0 \\
30\,/\,18\,/\,16\,/\,18\,/\,18 & 56.0 & 84.4 & 40.6 & 37.2 & 77.7 \\
10\,/\,8\,/\,37\,/\,8\,/\,37  & 50.0 & 85.7 & 30.6 & 35.0 & 75.9 \\
\textbf{Balanced (20\%$\times$5)} & \textbf{56.3} & \textbf{86.3} & 38.2 & 35.2 & \textbf{79.5} \\
\bottomrule
\end{tabular*}%
\end{table}

\subsection{ProofWriter Per-Depth}

\begin{table}[!ht]
\centering
\footnotesize
\setlength{\tabcolsep}{4pt}
\caption{ProofWriter per-depth accuracy (\%). Mid-training trains only to depth~5; eval depths 5--9. \textsuperscript{\textdagger}$N<200$, not significant at 95\% level. $\Delta$ is computed from exact correct counts before rounding individual accuracies; apparent differences between $\Delta$ and (Mid-training+SFT$-$SFT+RL) on rounded values are rounding artifacts.}
\label{tab:app-ProofWriter-full}
\begin{tabular*}{\textwidth}{@{\extracolsep{\fill}}lccccc@{}}
\toprule
Depth & SFT+RL & Mid-training+SFT & $\Delta$ & 95\% CI ($\Delta$) & $N$ \\
\midrule
D5 & 80.1 & 84.4 & $+4.4$ & $[+3.6, +5.2]$  & 18,498 \\
D6 & 76.2 & 82.7 & $+6.4$ & $[+3.3, +9.5]$  & 1,292  \\
D7 & 75.8 & 82.5 & $+6.7$ & $[-3.5, +17.0]$\textsuperscript{\textdagger} & 120 \\
D8 & 77.1 & 83.3 & $+6.2$ & $[-5.0, +17.4]$\textsuperscript{\textdagger} & 96  \\
D9 & 75.0 & 91.7 & $+16.7$ & $[-3.8, +37.2]$\textsuperscript{\textdagger} & 24  \\
\bottomrule
\end{tabular*}%
\end{table}

\paragraph{Ablating the covered depth.}
Depth-5 ProofWriter exposure improves both the covered D5 and held-out D6 strata; D7--D9 strata are too small for strong claims. Tables~\ref{tab:app-ProofWriter-depth-dist} and~\ref{tab:app-ProofWriter-ablation} ablate whether transfer depends on covered training depth (reference: SFT-only baseline, Overall 76.90). The two mixtures swap D3 and D5 proportions while holding D0--D2 and Other fixed. The D5-heavy model after SFT is substantially stronger ($+5.60$\,pp overall; $+5.70$\,pp D5, $+4.64$\,pp D6), suggesting post-training benefits more when mid-training difficulty is closer to the evaluation depth.

\begin{table}[!ht]
\centering
\footnotesize
\setlength{\tabcolsep}{4pt}
\begin{tabular*}{\textwidth}{@{\extracolsep{\fill}}lcccc@{}}
\toprule
Mixture & D0--D2 & D3 & D5 & Other \\
\midrule
D3-heavy & 2.04 & 65.74 & 17.12 & 15.10 \\
D5-heavy & 2.04 & 17.12 & 65.74 & 15.10 \\
\bottomrule
\end{tabular*}%
\caption{ProofWriter training-depth distribution in the covered-depth ablation (\%). Other denotes examples without parsed depth.}
\label{tab:app-ProofWriter-depth-dist}
\end{table}

\begin{table}[!ht]
\centering
\footnotesize
\setlength{\tabcolsep}{4pt}
\begin{tabular*}{\textwidth}{@{\extracolsep{\fill}}lccc@{}}
\toprule
Setting & Overall & D5 & D6 \\
\midrule
D3-heavy         & $-2.69$ & $-2.51$ & $-4.18$ \\
D3-heavy + SFT   & $+0.89$ & $+0.97$ & $-0.24$ \\
D5-heavy + SFT   & $+5.60$ & $+5.70$ & $+4.64$ \\
\bottomrule
\end{tabular*}%
\caption{ProofWriter covered-depth ablation results. Values are accuracy deltas (pp) relative to the ablation suite's SFT-only baseline.}
\label{tab:app-ProofWriter-ablation}
\end{table}

\subsection{ZebraLogic MC Per-House}

Table~\ref{tab:app-zebra-full} reports ZebraLogic MC results (MC = multiple-choice), complementing the ProofWriter depth-extrapolation result in §\ref{sec:cross_domain}. With only 201 direct mid-training ZebraLogic samples (0.3\% coverage), Mid-training+SFT outperforms SFT+RL at every reported house count. This is consistent with shared formal-deduction and constraint structure, but ZebraLogic is a same-family held-in exposure rather than an out-of-distribution transfer test, and it is not evidence that a small exposure alone caused the gains. Template/family-level overlap remains uncontrolled in this study, and a zero-exposure replication is future work.

\begin{table}[!ht]
\centering
\footnotesize
\setlength{\tabcolsep}{4pt}
\caption{ZebraLogic MC per-house accuracy (\%). Mid-training coverage: 0.3\% (201 samples). 95\% asymptotic CIs are shown; these are exposure-uncontrolled comparisons and are reported descriptively.}
\label{tab:app-zebra-full}
\begin{tabular*}{\textwidth}{@{\extracolsep{\fill}}lccccc@{}}
\toprule
Houses & SFT+RL & Mid-training only & Mid-training+SFT & $\Delta$ & 95\% CI ($\Delta$) \\
\midrule
2 & 89.9 & 90.1 & 94.1 & $+4.2$ & $[+1.1, +7.3]$  \\
3 & 78.8 & 77.4 & 84.3 & $+5.5$ & $[+1.2, +9.8]$  \\
4 & 60.2 & 56.5 & 72.6 & $+12.4$ & $[+7.5, +17.3]$ \\
5 & 47.6 & 40.5 & 55.9 & $+8.3$ & $[+3.0, +13.6]$ \\
6 & 31.3 & 28.6 & 43.1 & $+11.8$ & $[+6.6, +17.0]$ \\
\bottomrule
\end{tabular*}%
\end{table}

\subsection{CounterBench Per-Type}

\begin{table}[!ht]
\centering
\footnotesize
\setlength{\tabcolsep}{4pt}
\caption{CounterBench per-type accuracy (\%). Zero direct mid-training coverage; non-positive $\Delta$ is consistent with a no-transfer interpretation, but is not a definitive negative control.}
\label{tab:app-counterbench-full}
\begin{tabular*}{\textwidth}{@{\extracolsep{\fill}}lcccc@{}}
\toprule
Type & SFT+RL & Mid-only & Mid+SFT & $\Delta$ \\
\midrule
Basic       & 90.8 & 80.8 & 86.8 & $-4.0$ \\
Conditional & 80.4 & 72.0 & 80.4 & $\ \ 0.0$ \\
Joint       & 81.2 & 68.8 & 76.8 & $-4.4$ \\
Nested      & 76.8 & 64.8 & 72.8 & $-4.0$ \\
\bottomrule
\end{tabular*}%
\end{table}

The per-type $\Delta$ values are all at or below zero---a contrast with the positive deltas reported for ProofWriter and ZebraLogic. This pattern is consistent with, but does not prove, a no-transfer interpretation: CounterBench may differ in required procedures, prompt format, difficulty, and base-model prior. It should therefore not be treated as a definitive negative control.

\FloatBarrier
\section{Limitations and Ethical Considerations}
\label{app:limitations}

\paragraph{Scale.} The primary experiments use Qwen3-8B-Base, with a mid-training-only replication at Qwen3-4B-Base (\S\ref{app:expB}); sensitivity to other scales and architectures remains open (see Appendix~\ref{app:model_scale} for the scale-selection rationale). Whether the qualitative patterns persist at other model scales is unknown; fitted 95\%-of-peak intervals may shift with model size.

\paragraph{Scope and inference.} Claims are scoped to KOR-Bench and three external benchmarks, the Qwen3-Base family (8B primary, with a mid-training-only 4B replication), the reported data mixture, and the tested finite-budget recipe. The central inferential limit is the absence of held-out validation for the fitted peaks/intervals, $\theta^*$, and the full-pipeline gain: the 24-point curve fits, the fitted 95\%-of-peak intervals, the compensation thresholds, and $\theta^*$ all reuse the same sweep for model selection. Six held-out allocations carried through the complete mid-training$+$SFT$+$RL pipeline (Table~\ref{tab:mid_train}) now probe the mid-training-only curves and the compensatory-SFT result, which they roughly track at seed-noise level for Cipher, Logic, Counterfactual, and Puzzle (Operation's held-out low-coverage points sit $\approx6$\,pp above the fitted tail; 0/60 pairs closed at $5$\,pp), and carrying them through the RL stage leaves that unchanged ($0/60$ at $5$\,pp after the complete pipeline), so the RL leg is covered out-of-sample as well. What remains unvalidated on held-out allocations is $\theta^*$ itself, together with the fitted peaks, the 95\%-of-peak intervals, and the compensation thresholds, all of which are read off the same sweep that produced them. The simplex design additionally makes the fitted curves mixture-level marginal associations rather than isolated per-domain causal effects, and we treat this confound explicitly rather than implicitly (\S\ref{sec:simplex}). We report the compositional analysis this calls for (\S\ref{app:coda}): the isometric log-ratio surface fits well but is a saddle in every domain, so a jointly optimal mixture is not identified and the moderation result stands as a per-domain marginal statement. We tested whether that negative finding was a sampling artifact by adding 12 interior allocations to the pool (Appendix~\ref{app:expA}): a joint maximum still does not emerge and every domain's stationary point remains a saddle, so the moderation result is robustly a per-domain marginal statement and is not promoted to a single recommended mixture. Reported sweep values are five-seed means with cross-seed $\pm$1\,SD (Table~\ref{tab:mid_train}); the full-pipeline rows are likewise five-seed means, and their cross-seed SDs are displayed in Table~\ref{tab:main}. External-benchmark rows are five-seed means, and bootstrap CIs where used capture evaluation-sample variance only and do not replace the cross-seed uncertainty. We therefore claim no per-domain comparison as individually significant and report all per-domain deltas descriptively. The only row-level inferential statistics are the Welch two-sample $t$-tests on the overall full-pipeline scores (\S\ref{sec:stages}), which are at best nominally significant and do not survive multiplicity control; whether seeds are matched across pipeline rows is not established in this manuscript, so paired seed-level tests are left to the released per-seed data (Appendix~\ref{app:code}). GSPO uses a fixed 200-step budget; because binary verifier rewards provide no dense per-token reward signal, longer schedules were unstable across seeds (Appendix~\ref{app:rl_budget}), and longer or differently designed RL may yield different outcomes. Compensatory SFT conclusions are scoped to the fixed-budget, fixed-formula setting (Appendix~\ref{app:comp_sft}); stronger token-level reweighting, curricula, adaptive sampling, and stable RL after compensatory SFT remain untested. The current report also does not fully separate coverage from unique-example diversity, repetition, sequence length, or external-corpus exposure; family/template-level decontamination and seed-level analyses of the full-pipeline rows are left for future work.

\paragraph{Ethical considerations.} This work studies data-mixture design for mid-training. It does not involve human subjects or personally identifiable information. All corpora are either generated by deterministic symbolic solvers or sourced from publicly released benchmarks. Trained models are research checkpoints not intended for deployment. Benchmark balance should not be equated with user benefit, safety, calibration, or fairness in a deployed system; in high-stakes domains, low coverage can create uneven risk and would require domain-specific assurance before deployment.

\FloatBarrier
\section*{Declaration of AI Assistance}

We utilized ChatGPT and Claude for grammatical checking and \LaTeX{} support of the content presented in this study but did not use them for the initial draft of this study. Cursor, Codex, and Claude Code were utilized for trivial and boilerplate code completion during data analysis. We declare that all content presented and code utilized in this study has been reviewed and edited by the authors.


\begin{thebibliography}{DeepSeek-AI2024}
\setlength{\itemsep}{0pt}
\setlength{\bibsep}{1pt}
\bibitem[Abdin et al.(2024)]{abdin2024phi4}
M.~Abdin et al.
\newblock Phi-4 technical report.
\newblock arXiv:2412.08905, 2024.
\bibitem[Aitchison(1986)]{aitchison1986compositional}
J.~Aitchison.
\newblock The statistical analysis of compositional data.
\newblock Chapman and Hall, London, 1986.
\bibitem[Azerbayev et al.(2024)]{azerbayev2023llemma}
Z.~Azerbayev et al.
\newblock Llemma: An open language model for mathematics.
\newblock In \textit{ICLR}, 2024.
\bibitem[Cai et al.(2024)]{cai2024internlm2}
Z.~Cai et al.
\newblock {InternLM2} technical report.
\newblock arXiv:2403.17297, 2024.
\bibitem[Chen et al.(2024a)]{chen2024unlock}
J.~Chen et al.
\newblock Unlock the correlation between {SFT} and {RL} in training code {LLMs}.
\newblock arXiv:2406.10305, 2024.
\bibitem[Chen et al.(2024b)]{chen2024spin}
Z.~Chen et al.
\newblock Self-play fine-tuning converts weak language models to strong language models.
\newblock In \textit{ICLR}, 2024.
\bibitem[Chen et al.(2025)]{chen2025sasr}
J.~Chen et al.
\newblock Step-wise adaptive integration of {SFT} and {RL} for task-specific {LLMs}.
\newblock arXiv:2505.13026, 2025.
\bibitem[Chen et al.(2026)]{chen2026counterbench}
Y.~Chen, V.~K.~Singh, J.~Ma, and R.~Tang.
\newblock {CounterBench}: Evaluating and improving counterfactual reasoning in large language models.
\newblock In \textit{AAAI}, 2026.
\bibitem[Chu et al.(2025)]{chu2025sft}
T.~Chu, Y.~Zhai, J.~Yang, S.~Tong, S.~Xie, D.~Schuurmans, Q.~V.~Le, S.~Levine, and Y.~Ma.
\newblock {SFT} memorizes, {RL} generalizes: A comparative study of foundation model post-training.
\newblock arXiv:2501.17161, 2025.
\bibitem[Clark et al.(2018)]{clark2018arc}
P.~Clark et al.
\newblock Think you have solved question answering? {Try ARC}, the {AI2} reasoning challenge.
\newblock arXiv:1803.05457, 2018.
\bibitem[Cobbe et al.(2021)]{cobbe2021gsm8k}
K.~Cobbe et al.
\newblock Training verifiers to solve math word problems.
\newblock arXiv:2110.14168, 2021.
\bibitem[Dao et al.(2024)]{dao2023flashattention2}
T.~Dao, D.~Y.~Fu, S.~Ermon, A.~Rudra, and C.~R\'{e}.
\newblock {FlashAttention-2}: Faster attention with better parallelism and work partitioning.
\newblock In \textit{ICLR}, 2024.
\bibitem[Deng et al.(2025)]{deng2025supervisedrl}
Y.~Deng et al.
\newblock Supervised {RL}: From expert trajectories to step-wise reasoning.
\newblock arXiv:2510.25992, 2025.
\bibitem[Dubey et al.(2024)]{dubey2024llama3}
A.~Dubey et al.
\newblock The {Llama} 3 herd of models.
\newblock arXiv:2407.21783, 2024.
\bibitem[Fan et al.(2024)]{fan2024doge}
S.~Fan, M.~Pagliardini, and M.~Jaggi.
\newblock {DoGE}: Domain reweighting with generalization estimation.
\newblock In \textit{ICML}, 2024.
\bibitem[French(1999)]{french1999catastrophic}
R.~M. French.
\newblock Catastrophic forgetting in connectionist networks.
\newblock \textit{Trends in Cognitive Sciences}, 3(4):128--135, 1999.
\bibitem[Gao et al.(2021)]{gao2021pile}
L.~Gao, S.~Biderman, S.~Black, L.~Golding, T.~Hoppe, C.~Foster, J.~Phang, H.~He, A.~Thite, N.~Nabeshima, S.~Presser, and C.~Leahy.
\newblock The {Pile}: An 800{GB} dataset of diverse text for language modeling.
\newblock arXiv:2101.00027, 2021.
\bibitem[Guo et al.(2025)]{guo2025deepseekr1}
D.~Guo et al.
\newblock {DeepSeek-R1}: Incentivizing reasoning capability in {LLMs} via RL.
\newblock arXiv:2501.12948, 2025.
\bibitem[Gururangan et al.(2020)]{gururangan2020dapt}
S.~Gururangan, A.~Marasović, S.~Swayamditta, K.~Lo, I.~Beltagy, D.~Downey, and N.~A. Smith.
\newblock Don't stop pretraining: Adapt language models to domains and tasks.
\newblock In \textit{ACL}, 2020.
\bibitem[Han et al.(2024)]{han2022folio}
S.~Han et al.
\newblock {FOLIO}: Natural language reasoning with first-order logic.
\newblock In \textit{EMNLP}, 2024.
\bibitem[Havrilla et al.(2024)]{havrilla2024teaching}
A.~Havrilla et al.
\newblock Teaching large language models to reason with RL.
\newblock arXiv:2403.04642, 2024.
\bibitem[Hendrycks et al.(2021)]{hendrycks2021math}
D.~Hendrycks et al.
\newblock Measuring mathematical problem solving with the {MATH} dataset.
\newblock In \textit{NeurIPS}, 2021.
\bibitem[Huang et al.(2025)]{huang2025lorapar}
Y.~Huang et al.
\newblock {LoRA-PAR}: A flexible dual-system {LoRA} partitioning approach.
\newblock In \textit{EMNLP}, 2025.
\bibitem[Huang et al.(2026)]{huang2026remit}
J.~Huang et al.
\newblock {ReMiT}: {RL}-guided mid-training for iterative {LLM} evolution.
\newblock arXiv:2602.03075, 2026.
\bibitem[Ibrahim et al.(2024)]{ibrahim2024continual}
A.~Ibrahim, B.~Th\'{e}rien, K.~Gupta, M.~L.~Richter, Q.~Anthony, T.~Lesort, E.~Belilovsky, and I.~Rish.
\newblock Simple and scalable strategies to continually pre-train large language models.
\newblock arXiv:2403.08763, 2024.
\bibitem[Ishibashi et al.(2025)]{ishibashi2025mining}
Y.~Ishibashi et al.
\newblock Mining hidden thoughts from texts: Evaluating continual pretraining with synthetic data for {LLM} reasoning.
\newblock arXiv:2505.10182, 2025.
\bibitem[Ke et al.(2025)]{ke2025survey}
Z.~Ke et al.
\newblock A survey of frontiers in {LLM} reasoning.
\newblock In \textit{TMLR}, 2025.
\bibitem[Kim and Linzen(2020)]{kim2020cogs}
N.~Kim and T.~Linzen.
\newblock {COGS}: A compositional generalization challenge based on semantic interpretation.
\newblock In \textit{EMNLP}, 2020.
\bibitem[Kirkpatrick et al.(2017)]{kirkpatrick2017ewc}
J.~Kirkpatrick, R.~Pascanu, N.~Rabinowitz, J.~Veness, G.~Desjardins, A.~A. Rusu, K.~Milan, J.~Quan, T.~Ramalho, A.~Grabska-Barwinska, D.~Hassabis, C.~Clopath, D.~Kumaran, and R.~Hadsell.
\newblock Overcoming catastrophic forgetting in neural networks.
\newblock \textit{Proceedings of the National Academy of Sciences}, 114(13):3521--3526, 2017.
\bibitem[Lin et al.(2025)]{lin2025zebralogic}
B.~Y.~Lin, R.~{Le Bras}, K.~Richardson, A.~Sabharwal, R.~Poovendran, P.~Clark, and Y.~Choi.
\newblock {ZebraLogic}: On the scaling limits of {LLMs} for logical reasoning.
\newblock In \textit{ICML}, 2025.
\bibitem[Liu et al.(2020)]{liu2020logiqa}
J.~Liu, L.~Cui, H.~Liu, D.~Huang, Y.~Wang, and Y.~Zhang.
\newblock {LogiQA}: A challenge dataset for machine reading comprehension with logical reasoning.
\newblock In \textit{IJCAI}, 2020.
\bibitem[Liu et al.(2025)]{liu2024regmix}
Q.~Liu et al.
\newblock {RegMix}: Data mixture as regression for language model pre-training.
\newblock In \textit{ICLR}, 2025.
\bibitem[Lu et al.(2024)]{lu2024mathcoder2}
Z.~Lu et al.
\newblock {MathCoder2}: Better math reasoning from continued pretraining on model-translated mathematical code.
\newblock arXiv:2410.08196, 2024.
\bibitem[Luo et al.(2023a)]{luo2023forgetting}
Y.~Luo, Z.~Yang, F.~Meng, Y.~Li, J.~Zhou, and Y.~Zhang.
\newblock An empirical study of catastrophic forgetting in large language models during continual fine-tuning.
\newblock arXiv:2308.08747, 2023.
\bibitem[Luo et al.(2023b)]{luo2023wizardmath}
H.~Luo et al.
\newblock {WizardMath}: Empowering mathematical reasoning for large language models via reinforced evol-instruct.
\newblock arXiv:2308.09583, 2023.
\bibitem[Ma et al.(2025)]{ma2024korbench}
K.~Ma, X.~Du, Y.~Wang, H.~Zhang, Z.~Wen, X.~Qu, J.~Yang, J.~Liu, M.~Liu, X.~Yue, W.~Huang, and G.~Zhang.
\newblock {KOR-Bench}: Benchmarking language models on knowledge-orthogonal reasoning tasks.
\newblock In \textit{ICLR}, 2025.
\bibitem[Matsutani(2025)]{matsutani2025rl}
K.~Matsutani.
\newblock {RL} squeezes, {SFT} expands: A comparative study of reasoning {LLMs}.
\newblock arXiv:2509.21128, 2025.
\bibitem[Muennighoff et al.(2023)]{muennighoff2023scaling}
N.~Muennighoff et al.
\newblock Scaling data-constrained language models.
\newblock In \textit{NeurIPS}, 2023.
\bibitem[Ouyang et al.(2022)]{ouyang2022instructgpt}
L.~Ouyang et al.
\newblock Training language models to follow instructions with human feedback.
\newblock In \textit{NeurIPS}, 2022.
\bibitem[Penedo et al.(2024)]{penedo2024fineweb}
G.~Penedo, H.~Kydlí\v{c}ek, L.~B.~allal, A.~Lozhkov, M.~Mitchell, C.~Raffel, L.~V.~Werra, and T.~Wolf.
\newblock The {FineWeb} datasets: Decanting the web for the finest text data at scale.
\newblock arXiv:2406.17557, 2024.
\bibitem[Qin et al.(2026)]{qin2026davinci}
Y.~Qin et al.
\newblock {daVinci-LLM}: Towards the science of pretraining.
\newblock arXiv:2603.27164, 2026.
\bibitem[Rafailov et al.(2023)]{rafailov2023dpo}
R.~Rafailov, A.~Sharma, E.~Mitchell, C.~D.~Manning, S.~Ermon, and C.~Finn.
\newblock Direct preference optimization: Your language model is secretly a reward model.
\newblock In \textit{NeurIPS}, 2023.
\bibitem[Rajbhandari et al.(2020)]{rajbhandari2020zero}
S.~Rajbhandari, J.~Rasley, O.~Ruwase, and Y.~He.
\newblock {ZeRO}: Memory optimizations toward training trillion parameter models.
\newblock In \textit{SC}, 2020.
\bibitem[Ren et al.(2026)]{ren2026rethinking}
Q.~Ren et al.
\newblock Rethinking generalization in reasoning {SFT}: A conditional analysis on optimization, data, and model capability.
\newblock arXiv:2604.06628, 2026.
\bibitem[Ruis et al.(2024)]{ruis2024procedural}
L.~Ruis et al.
\newblock Procedural knowledge in pretraining drives reasoning in large language models.
\newblock arXiv:2411.12580, 2024.
\bibitem[Shao et al.(2024)]{shao2024deepseekmath}
Z.~Shao et al.
\newblock {DeepSeekMath}: Pushing the limits of mathematical reasoning in open language models.
\newblock arXiv:2402.03300, 2024.
\bibitem[Shen et al.(2023)]{shen2023slimpajama}
Z.~Shen et al.
\newblock {SlimPajama-DC}: Understanding data combinations for {LLM} training.
\newblock arXiv:2309.10818, 2023.
\bibitem[Sinha et al.(2019)]{sinha2019clutrr}
K.~Sinha, S.~Sodhani, J.~Dong, J.~Pineau, and W.~L. Hamilton.
\newblock {CLUTRR}: A diagnostic benchmark for inductive reasoning from text.
\newblock In \textit{EMNLP-IJCNLP}, 2019.
\bibitem[Soldaini et al.(2024)]{soldaini2024dolma}
L.~Soldaini et al.
\newblock Dolma: An open corpus of three trillion tokens for language model pretraining research.
\newblock In \textit{ACL}, 2024.
\bibitem[Suzgun et al.(2023)]{suzgun2022bbhard}
M.~Suzgun et al.
\newblock Challenging {BIG-Bench} tasks and whether chain-of-thought can solve them.
\newblock In \textit{Findings of ACL}, 2023.
\bibitem[Tafjord et al.(2021)]{tafjord2021proofwriter}
O.~Tafjord, B.~D. Mishra, and P.~Clark.
\newblock {ProofWriter}: Generating implications, proofs, and abductive statements over natural language.
\newblock In \textit{ACL-IJCNLP}, 2021.
\bibitem[Team {KIMI}(2025)]{teamkimi2025kimi}
{Team KIMI}.
\newblock {Kimi} k1.5: Scaling reinforcement learning with {LLMs}.
\newblock arXiv:2501.12599, 2025.
\bibitem[Toshniwal et al.(2024)]{toshniwal2024openmathinstruct}
S.~Toshniwal et al.
\newblock {OpenMathInstruct-1}: A 1.8 million math instruction tuning dataset.
\newblock In \textit{NeurIPS}, 2024.
\bibitem[Tu et al.(2025)]{tu2025midtrain}
C.~Tu et al.
\newblock A survey on {LLM} mid-training.
\newblock arXiv:2510.23081, 2025.
\bibitem[Wang et al.(2023)]{wang2023selfinstruct}
Y.~Wang, Y.~Kordi, S.~Mishra, A.~Liu, N.~A. Smith, D.~Khashabi, and H.~Hajishirzi.
\newblock Self-Instruct: Aligning language models with self-generated instructions.
\newblock In \textit{ACL}, 2023.
\bibitem[Wang et al.(2025)]{wang2025rlsr}
Z.~Wang et al.
\newblock {RLSR}: Reinforcement learning with supervised reward outperforms {SFT} in instruction following.
\newblock arXiv:2510.14200, 2025.
\bibitem[Wang et al.(2026)]{wang2026mergemix}
J.~Wang, C.~Tian, K.~Chen, Z.~Liu, J.~Mao, W.~X.~Zhao, Z.~Zhang, and J.~Zhou.
\newblock {MergeMix}: Optimizing mid-training data mixtures via learnable model merging.
\newblock arXiv:2601.17858, 2026.
\bibitem[Wei et al.(2022)]{wei2022cot}
J.~Wei et al.
\newblock Chain-of-thought prompting elicits reasoning in large language models.
\newblock In \textit{NeurIPS}, 2022.
\bibitem[{Xiaomi LLM-Core}(2025)]{xiaomi2025mimo}
{Xiaomi LLM-Core}.
\newblock {MiMo}: Unlocking the reasoning potential of language model.
\newblock arXiv:2505.07608, 2025.
\bibitem[Xie et al.(2023a)]{xie2024doremi}
S.~M.~Xie, H.~Pham, X.~Dong, N.~Du, H.~Liu, Y.~Lu, P.~Liang, Q.~V. Le, T.~Ma, and A.~W. Yu.
\newblock {DoReMi}: Optimizing data mixtures speeds up language model pretraining.
\newblock In \textit{NeurIPS}, 2024.
\bibitem[Xie et al.(2023b)]{xie2023dsir}
S.~M.~Xie, S.~Santurkar, T.~Ma, and P.~Liang.
\newblock Data selection for language models via importance resampling.
\newblock In \textit{NeurIPS}, 2023.
\bibitem[Yang et al.(2024)]{yang2024fineline}
C.~Yang et al.
\newblock The fine line: Navigating {LLM} pretraining with down-streaming capability analysis.
\newblock arXiv:2404.01204, 2024.
\bibitem[Yang et al.(2025)]{yang2025qwen3}
A.~Yang et al.
\newblock {Qwen3} technical report.
\newblock arXiv:2505.09388, 2025.
\bibitem[Ye et al.(2025)]{ye2024datamixinglaws}
J.~Ye et al.
\newblock Data mixing laws: Optimizing data mixtures by predicting language modeling performance.
\newblock In \textit{ICLR}, 2025.
\bibitem[Yu et al.(2020)]{yu2020pcgrad}
T.~Yu, S.~Kumar, A.~Gupta, S.~Levine, K.~Hausman, and C.~Finn.
\newblock Gradient surgery for multi-task learning.
\newblock In \textit{NeurIPS}, 2020.
\bibitem[Yu et al.(2023)]{yu2023metamath}
L.~Yu et al.
\newblock {MetaMath}: Bootstrap your own mathematical questions for large language models.
\newblock arXiv:2309.12284, 2023.
\bibitem[Zhang et al.(2025)]{zhang2025interplay}
C.~Zhang, G.~Neubig, and X.~Yue.
\newblock On the interplay of pre-training, mid-training, and {RL} on reasoning language models.
\newblock arXiv:2512.07783, 2025.
\bibitem[Zhao et al.(2025)]{zhao2025echo}
R.~Zhao, A.~Meterez, S.~Kakade, C.~Pehlevan, S.~Jelassi, and E.~Malach.
\newblock Echo chamber: {RL} post-training amplifies behaviors learned in pretraining.
\newblock In \textit{COLM}, 2025.
\bibitem[Zhou et al.(2023)]{zhou2023lima}
C.~Zhou et al.
\newblock {LIMA}: Less is more for alignment.
\newblock In \textit{NeurIPS}, 2023.
\bibitem[Zhou et al.(2025)]{zhou2024rulearena}
R.~Zhou et al.
\newblock {RuleArena}: A benchmark for rule-guided reasoning with {LLMs} in real-world scenarios.
\newblock In \textit{ACL}, 2025.
\end{thebibliography}
\end{document}